\documentclass[11pt]{article}
\usepackage[margin=1in]{geometry}
\usepackage{fontspec}
\usepackage{amsmath,amssymb}
\newfontfamily\cjkfont{FandolSong-Regular.otf}[Path=, Extension=]
\newcommand{\cjk}[1]{{\cjkfont\XeTeXlinebreaklocale "zh"\XeTeXlinebreakskip=0pt plus 1pt #1}}
\usepackage{graphicx}
\usepackage{booktabs}
\usepackage{array}
\usepackage{adjustbox}
\usepackage[font=small,labelfont=bf,skip=4pt]{caption}
\usepackage{enumitem}
\usepackage[dvipsnames]{xcolor}
\usepackage{newunicodechar}
\usepackage{placeins}
\usepackage{microtype}
\usepackage[hidelinks]{hyperref}
\graphicspath{{figures/}}
\newunicodechar{§}{\S}
\newunicodechar{—}{\textemdash}
\newunicodechar{–}{\textendash}
\newunicodechar{±}{\ensuremath{\pm}}
\newunicodechar{×}{\ensuremath{\times}}
\newunicodechar{°}{\ensuremath{{}^\circ}}
\newunicodechar{→}{\ensuremath{\rightarrow}}
\newunicodechar{−}{\ensuremath{-}}
\newunicodechar{‑}{-}
\newunicodechar{≈}{\ensuremath{\approx}}
\newunicodechar{≤}{\ensuremath{\le}}
\newunicodechar{≥}{\ensuremath{\ge}}
\newunicodechar{≫}{\ensuremath{\gg}}
\newunicodechar{χ}{\ensuremath{\chi}}
\newunicodechar{²}{\ensuremath{{}^2}}
\newunicodechar{⁴}{\ensuremath{{}^4}}
\newunicodechar{↑}{\ensuremath{\uparrow}}
\newunicodechar{↓}{\ensuremath{\downarrow}}
\newunicodechar{‖}{\ensuremath{\|}}
\newunicodechar{Δ}{\ensuremath{\Delta}}
\newunicodechar{∼}{\ensuremath{\sim}}
\newunicodechar{’}{'}
\newunicodechar{ö}{\"o}
\newunicodechar{é}{\'e}
\title{Geometric and Behavioral Stratification in Transformer Residual Streams}
\author{Nelson Guda, PhD}
\date{}

\begin{document}
\maketitle

\begin{abstract}
Recent work has shown that trained transformer models develop privileged bases: specific coordinate axes whose statistical behavior differs from the rest of the residual stream. But what kind of direction does such a privileged basis select? We investigate the prediction direction, the unembedding direction of the token a model currently predicts, and find that it functions as a content-defined privileged anchor. When measured with respect to this anchor, residual-stream variation is geometrically and behaviorally stratified by proximity to the prediction.

The stratification appears in all eighteen models we test (dense and MoE, 7B–120B, base and instruction-tuned). A narrow, scale-invariant prediction interface concentrates readout-relevant structure while the vast remainder, a prediction-distal complement, expands with model scale. We show that variance-based analyses recover this organization only partly, because the prediction direction sits nearly orthogonal to the principal variance axes. Organization along the prediction axis becomes more pronounced when comparing among more linguistically heterogeneous prompts. So as prompts become more diverse, model separation among prompts becomes less visible to the principal variance axes and more visible in relation to the prediction axis.

Anchoring to the prediction direction reveals a steep geometric gradient. Prediction-proximal regions are highly structured and cluster semantically related prompts, while the complement is flatter and systematically anti-discriminates among prompt groupings. These prediction-proximal subspaces make up only a narrow slice of the residual stream, but prove functionally relevant. Disrupting the high variance directions that are most proximal to the prediction direction causes immediate divergence and frequent task-frame shifts. Disrupting directions in the next level of variance down produces later divergence and preserves the framing of the output. The rest of the residual stream geometry is low-variance and weakly aligned with the readout of the next token, but interventions show that it is causally and temporally load-bearing. Across the entire geometry we find that behavior is driven by directional structure, not magnitude, suggesting that the geometric stratification we identify helps models isolate readout structure from computation.

These results establish the prediction direction as a privileged anchor that is distinct from previously described coordinate-based axes and functionally aligned with the way models are trained. Our data also offers a geometric account of how high-dimensional computation can coexist with linear readout. We suggest this work has implications for interpretability and evaluation.
\end{abstract}

\begin{figure}[tbp]
\centering
\includegraphics[width=\linewidth]{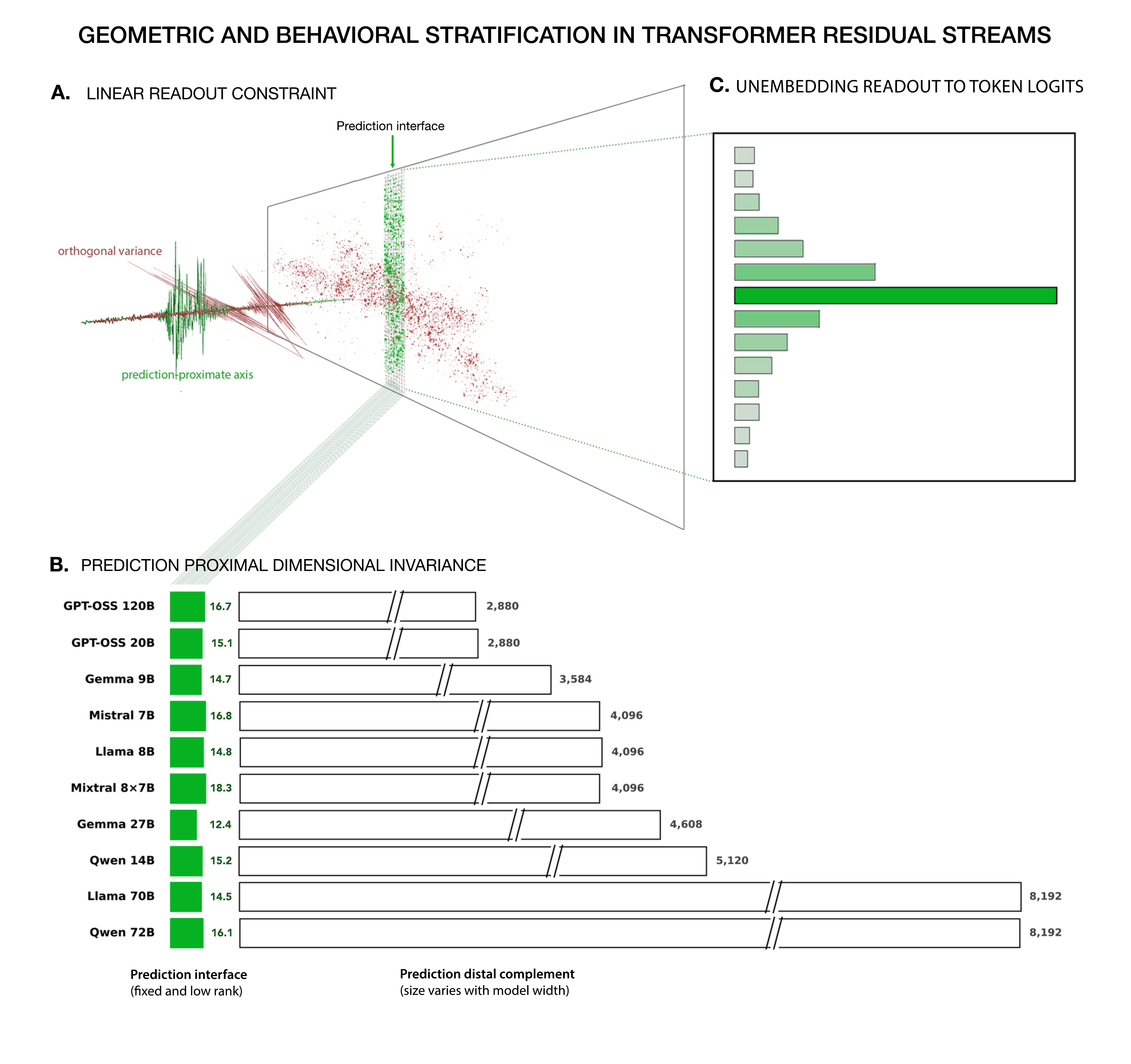}
\caption{\textbf{The linear readout uses only a narrow region of the residual stream to determine the next token, and this region stays the same size regardless of how large the model is — the architectural asymmetry the paper builds on.} \textbf{(A)} A conceptual visualization of how variation in the residual stream can be aligned with or orthogonal to the readout. The green variation represents the \emph{prediction interface}, the small region of the residual stream whose components directly determine which token the linear readout selects next. The rest of the residual stream — the red scatter, geometrically orthogonal to the prediction interface — represents the \emph{complement}, the vast majority of directions that carry the model's computational state but do not directly contribute to the readout. \textbf{(B)} The narrow dimensionality of the prediction interface is universal across models regardless of their size. The ten instruction-tuned models in the testbed, spanning a 2.8× range of hidden dimensions (full eighteen-model testbed in §4), all show the same pattern. The prediction interface in green occupies the same low effective dimensionality regardless of model size (adjacent number gives its effective dimensionality for $1 + D_{PR} + S_{PR}$ from Table 1). The complement (white outline) grows with the residual-stream width ($d_{\text{model}}$, right). The bars are drawn compressed (break marks). The complement would be far larger relative to the interface if rendered at true scale. In summary, wider models compute through a larger complement but read out through the same narrow band. \textbf{(C)} The linear readout produces a distribution over candidate tokens, read out from the prediction interface in (A), with the dominant bar (the argmax) selected as the prediction.}
\end{figure}

\FloatBarrier
\phantomsection\addcontentsline{toc}{section}{1. Introduction}
\section*{1. Introduction}

Transformer language models compute in a high-dimensional residual stream that has no rotational preference before training. Recent work has shown that training breaks native rotational symmetry and induces a \emph{privileged basis}: specific coordinate axes whose statistics differ systematically from the rest (van Nierop, 2024; Elhage et al., 2023). That basis has so far been described only as a set of anomalous coordinate axes, not by any functional role. In this paper we present evidence that the \emph{prediction direction}, the rank-1 unembedding direction for the token a model predicts at every step, is a content-defined privileged anchor. Across every model we test, residual-stream geometry organizes relative to it.

The prediction direction is a natural place for such structure to concentrate, because it is where the architecture makes computation legible. A transformer computes in a high-dimensional residual stream, the running per-layer state that accumulates updates across a sequence, but produces each next-token prediction through a \emph{linear readout}, a single linear projection of the residual stream state onto vocabulary logits (Vaswani et al., 2017; Elhage et al., 2021). This creates a basic asymmetry: computation occupies thousands of dimensions, but prediction becomes legible only through alignment with the unembedding. Because autoregressive training shapes the residual stream entirely through what this linear readout resolves at each step, the model's own prediction direction is a functionally defined axis against which residual-stream geometry can be measured. Prior research has used this direction as a diagnostic readout axis in logit-lens methods (nostalgebraist, 2020; Belrose et al., 2023). We use this work to show that the prediction direction is also a geometric anchor.

We examined the influence of the prediction direction on the geometric structure of the residual stream by looking at variance in residual stream geometry across structured prompt sets. By anchoring analyses to the prediction direction we identified a narrow \emph{prediction interface}, a low-dimensional, readout-proximal region, that is universal across our 18 model testbed. This narrow interface, through which residual-stream computation becomes legible to the unembedding, is an extremely thin slice of the geometry, fewer than 20 effective dimensions on constrained prompts among the many thousands that span the residual stream. The rest of the geometry, which we call the prediction-\emph{distal complement}, is weakly aligned with the readout per direction but remains causally and temporally essential for generation.

To explain this finding, we propose a geometric account of how high-dimensional computation coexists with a linear readout across a defined output. By demonstrating that variation in transformer model residual streams is primarily directional rather than magnitudinal, we propose that prediction-relevant structure must be kept clear of the much larger background of residual-stream variation in order to isolate the unembedding signal at readout. Our findings suggest that models achieve this by orienting the majority of variation away from the readout while preserving a narrow region where prediction-relevant structure can be expressed. Because the readout resolves the next token through a fixed, narrow interface, larger models should keep that interface narrow and expand the complement instead. Figure 1A shows a conceptual rendering of how variation on different axes may be oriented toward or away from the unembedding readout.

To assess this question we used a testbed of eighteen transformer models from six architecture families, Llama, Gemma, Mistral, Mixtral, Qwen, and GPT-OSS, spanning dense and mixture-of-experts designs, base and instruction-tuned variants, and a 17× parameter range (7B–120B). For each model, we measured prediction-proximal and prediction-distal structures in the residual stream geometry across three structured prompt sets that were designed to assess how models construct residual stream geometry in response to variation in prompts. We derived our primary unit of measurement with a prediction-anchored decomposition, \emph{PDSF}, that partitions each residual state into four orthogonal components ordered by variance proximity to the prediction direction. The PDSF decomposition first extracts P, the rank-1 projection onto the unembedding direction for the current argmax token, then takes successive variance-ordered cuts of the remainder: D, the top-k PCA basis of the P-residual; S, the top-k PCA basis of the (P+D)-residual; and F, the orthogonal complement of all three (detailed in §3). We call the combined prediction-proximal region P + D + S the prediction interface. The remaining portion of the geometry is by far the largest and has far less per-dimension variance among prompts. We call this prediction-distal region "F", and throughout the paper we also refer to it as the complement.

Using this decomposition method we find that the effective dimensionality of the prediction-proximal region is small and approximately invariant to model width. On constrained prompts (prompts that expect a simple one token answer), prediction-proximal D subspace has single-digit effective rank across both instruction-tuned models and base models. When we measure manifold complexity across the decomposed subspaces, we find a strong gradient along the axis from highly folded structure in D to flatter and less folded structure in F. Comparing instruction and base models shows that tuning sharpens the interface's internal organization but does not widen it. In every model we test, we find the same stratification.

Using geometric variance among prompts as a decomposition method gives an a priori expectation of a gradient in discrimination. A surprising finding was that the low variance, prediction-distal complement does not have less discrimination, but in fact anti-discriminates across structured prompt sets.

After finding geometric stratification along the prediction axis we tested the functional implications by assessing behavioral response of model output to targeted interventions on our measured subspaces. When we intervened on the prediction-distal component of the residual stream, we found that although the dimensions in this majority of the geometry are weakly aligned with the next token, the geometry remains causally critical to computation and is necessary at every step. When we intervened on the narrow prediction proximal subspaces D and S, we found that variance proximity to the next predicted token direction aligns with how far in the future the intervention disturbs the output. In other words, intervening on the dimensions that are the highest variance affect the output immediately, while intervening on the next highest variance group of dimensions affects the output only several tokens later. This result seems surprising and suggests that models partition directions that affect immediate versus future output.

None of these findings, the scale-invariance, the manifold-complexity gradient, F's sign inversion in group discrimination, or the behavioral hierarchy, are forced by how PDSF is constructed. The PDSF decomposition is an analytical coordinate system for measuring residual-stream geometry, and the only a priori expectation (the discrimination gradient) also revealed unexpected results.

We believe our results connect to work in linear-representation and probing studies that recover semantically meaningful information from low-dimensional directions in activation space (Marks \& Tegmark, 2023; Park et al., 2024) and sparse autoencoders that recover information as sparse directions (Bricken et al., 2023). All these studies show that readable structure is low-dimensional or sparse. Saurez et al. (2026) sharpened this into an architectural claim that features which read out through a linear interface must occupy context-invariant linear subspaces as a consequence of the readout geometry alone. These results also bear on work that characterizes residual-stream geometry through variance structure, intrinsic dimensionality, or depth-wise geometric phases (Ansuini et al., 2019; Valeriani et al., 2023; Cheng et al., 2025; Kirsanov et al., 2025). Our work shows that geometric structure in the residual stream is organized relative to the axis that next-token training shapes most directly - the model's own current prediction direction.

We believe this work has implications for scaling, interpretability, and evaluation. If model width mainly expands the substrate around a readout interface of roughly fixed size, then larger models likely improve by computing more precise projections into that interface rather than by widening it. Our findings of behavioral differences between subspaces with differing proximity to the prediction direction suggest that this may be a valuable tool for feature-level interpretability, because the two regions carry different kinds of information at different readout strengths. These results also contribute to the growing body of work which suggests that output-level evaluation under-measures computation. Our work adds to this by demonstrating that the majority of residual stream variance is weakly visible to the readout.

\textbf{Outline.} §2 develops the architectural constraint imposed by linear readout under norm concentration. §3 defines the PDSF decomposition and the geometric measurements used throughout. §4 reports the scale-invariant prediction interface, readout/variance separation, manifold-complexity gradient, group-discrimination inversion, and base/instruction comparison. §5 tests whether the complement's causal role is carried by direction rather than magnitude. §6 uses persistent rotational interventions to characterize behavioral stratification along the prediction-proximity axis. §7 identifies the static/dynamic partition of the complement during autoregressive generation. §8 discusses implications for scaling, interpretability, and evaluation blindness, and concludes.

\FloatBarrier
\phantomsection\addcontentsline{toc}{section}{2. Architectural Constraint of Linear Readout}
\section*{2. Architectural Constraint of Linear Readout}

A linear readout sees the residual stream only through projection onto vocabulary directions. At each step the projection creates a distribution of potential tokens (logits) computed as

\[
\ell_i = \mathbf{u}_i^T \mathbf{h},
\]

where $\mathbf{h} \in \mathbb{R}^d$ is the residual stream and $\mathbf{u}_i$ is the unembedding vector for token $i$. The predicted next token is the argmax over these dot products, so correct prediction requires that the residual state align more strongly with the correct token's unembedding vector than with competing tokens. The direction we use as the prediction direction is the unembedding vector of the token the model currently predicts — its argmax — which we denote $\hat{\mathbf{u}}$.

The architectural asymmetry we identify raises a question: how much of the total information carried by the residual stream is actually needed to produce the logit distribution? Likely far less than the stream carries in total. The choice of a next token is influenced by directions that influence future tokens, but models must also represent alternative continuations and intermediate computation that never directly drive the current token. This makes prediction a signal-to-noise problem in representation space. The signal is the margin by which the selected token leads the output distribution — set by $\mathbf{h}$'s projection onto that token's readout direction relative to its near-competitors. Noise is residual-stream variation that erodes this margin. A competitor direction is therefore not noise in itself; it becomes noise only insofar as variation along it narrows the selected token's lead.

To fully understand how this signal-to-noise problem applies to transformer models, it is helpful to view variation in the residual stream through a known aspect of high-dimensional geometry: when variables are approximately normally distributed, thin-shell theory predicts that the norm of a high-dimensional vector concentrates in a thin radial shell (Vershynin, 2018; Ledoux, 2001). Combined with the linear-readout constraint, norm concentration implies that functional variation must be expressed primarily through direction rather than magnitude. Despite evidence of anisotropy in residual-stream geometry (Ethayarajh, 2019), the normalizing routines applied at every layer (Ba et al., 2016; Zhang \& Sennrich, 2019) constrain the residual norm and suggest this geometric principle should apply. Related work has converged on evidence of the direction-over-magnitude picture from other angles: angular perturbations to hidden states damage language modeling far more than magnitude-matched ones (Vardhan \& Sai Teja, 2026), and architectures that remove the radial degree of freedom outright (Loshchilov et al., 2024) or approximate it from the same norm-concentration argument (Franke et al., 2025) train more efficiently. We first confirmed that the thin-shell hypothesis holds geometrically across our testbed (Appendix B.1), and then tested whether it carries causal force inside the prediction-distal complement (§5).

If transformer models operate under an implicit directional representational budget then all functional roles including prediction, discrimination, contextual state, and coordination across layers, must be expressed through directions in a shared high-dimensional space. They cannot be separated by magnitude. Under this budget, the signal-to-noise constraint favors a low-dimensional, well-isolated prediction interface that is variance-separated from a complement that maintains variation at low overlap with the readout. We elaborate on this in Appendix B.2.

In short, the complement protects the readout through orientation, not magnitude. It can hold large variance as long as that variance interferes little with the directions that decide the output.

By using three prompt sets of varying semantic diversity, we also show that effective dimensionality of the interface is determined not just by the structure of the readout but also by the demands of the prediction problem — how much the prompt distribution asks the model to distinguish — neither of which depends directly on the ambient hidden dimension $d$. This asymmetry also carries a consequence for evaluation: if most directions in the residual stream are only weakly aligned with the readout, output-based metrics will under-measure computation occurring in the complement (see §8.9 for discussion).

\FloatBarrier
\phantomsection\addcontentsline{toc}{section}{3. The PDSF Decomposition}
\section*{3. The PDSF Decomposition}

To assess the implications of a constrained interface, we used a coordinate system that partitions the residual stream relative to the model's own prediction direction. PDSF decomposes the hidden state at every layer into four strictly orthogonal components ordered by prediction proximity. Full implementation details including hidden state extraction, basis construction, and intervention machinery, are in Appendix A.

\[
\mathbf{h} = \mathbf{h}_P + \mathbf{h}_D + \mathbf{h}_S + \mathbf{h}_F
\]

The decomposition proceeds by sequential orthogonal projection and removal:

\textbf{P (Predictive)}: the rank-1 projection of $\mathbf{h}$ onto the normalized unembedding vector $\hat{\mathbf{u}}$ for the model's argmax predicted token, $\mathbf{h}_P = \hat{\mathbf{u}}(\hat{\mathbf{u}}^T \mathbf{h})$. We use the logit-lens methodology (nostalgebraist, 2020; Belrose et al., 2023) but treat the prediction-aligned direction as a geometric anchor rather than a readout tool. The anchor is the argmax direction of the same forward pass and is held fixed across depth, so intermediate-layer geometry is measured in a frame set by the token that pass resolves to rather than by any commitment the model has made at that depth. Holding it fixed is deliberate: depth-to-depth comparisons then reflect changes in the geometry rather than changes in the frame, whereas an anchor re-derived from each layer's own argmax would confound the two.

\textbf{D (Discriminative)}: After removing P, the per-prompt residuals $\mathbf{h}_{\perp} = \mathbf{h} - \mathbf{h}_P$ are collected across all prompts at the target layer into a matrix $H_{\perp}$, and the D basis $B_D$ is the top-$k_D$ PCA basis of $H_{\perp}$ after its cross-prompt mean is subtracted, so $B_D$ spans the directions along which prompts differ most from one another. Rank is set adaptively by the participation ratio of that same centered matrix, $k_D = \text{round}(\mathrm{PR})$, clamped to $[2, 12]$ as in A.3. Each prompt's D component is then the projection of its own uncentered residual onto that basis, $\mathbf{h}_D = B_D(B_D^T \mathbf{h}_{\perp})$, so the cross-prompt mean of $\mathbf{h}_{\perp}$ is not itself assigned to D.

\textbf{S (Situational)}: After removing both P and D, the double residuals $\mathbf{h}_{\perp\perp} = \mathbf{h}_{\perp} - \mathbf{h}_D$ are collected into a matrix $H_{\perp\perp}$, and the S basis $B_S$ is the top-$k_S$ PCA basis of $H_{\perp\perp}$, under the same cross-prompt centering and the same PR-adaptive rank procedure. Each prompt's S component is the projection of its own uncentered double residual onto that basis, $\mathbf{h}_S = B_S(B_S^T \mathbf{h}_{\perp\perp})$.

\textbf{F (Framework)}: F is the remainder, $\mathbf{h}_F = \mathbf{h}_{\perp\perp} - \mathbf{h}_S$. It is the largest of the found subspaces, spanning thousands of dimensions.

These four subspaces sample the variance continuum at decreasing levels of variance per dimension, where P > D > S > F. The construction methodology guarantees strict orthogonality and lossless reconstruction. We validated reconstruction and found a max reconstruction error = 0.0 across all models. This orthogonality ensures that interventions on one subspace do not mechanically contaminate another, a property Miller et al. (2026) show directly improves causal intervention success.

\phantomsection\addcontentsline{toc}{subsection}{Stable terminology}
\subsection*{Stable terminology}

Throughout, we use:

\begin{itemize}[leftmargin=1.4em,itemsep=1pt,topsep=2pt]
\item \textbf{prediction direction:} the rank-1 unembedding direction $\hat{\mathbf{u}}$ for the model's current argmax token (i.e., the P basis);
\item \textbf{prediction-proximity axis:} the analytical ordering of residual-stream structure by proximity to the prediction direction, from the rank-1 anchor outward through successive variance cuts to the complement. It is an ordering over subspaces, not a one-dimensional spatial direction in the residual stream, and the measurements that populate it are the binned quantities described below;
\item \textbf{prediction interface:} the combined region of P + D + S;
\item \textbf{F / prediction-distal complement} / \textbf{complement:} the remaining subspace;
\item \textbf{prediction-proximal:} for properties closer to the prediction direction along the prediction-proximity axis;
\item \textbf{prediction-distal:} properties or directions in the complement;
\item \textbf{F-topK:} the top-$k$ cross-prompt variance basis inside F — the next nested PCA cut after P, D, and S have been removed. Rank is selected adaptively by participation ratio per prompt set (on SpecA, 28–64 across models; 33–51 on SpecB and 28–47 on Diverse);
\item \textbf{F-base:} the orthogonal complement of F-topK within F (i.e., F − F-topK), the low-variance bulk of F that remains after F-topK is removed;
\item \textbf{static/dynamic partition:} (§7) the structural feature of the complement that separates a \emph{static component} (also \emph{static scaffold}), which is approximately constant across the generation window, from a \emph{dynamic component} (also \emph{dynamic envelope}) that is rewritten at every generation step.
\end{itemize}

\textbf{Effective dimensionality in measurements and interventions.} In Section 4.1 we use these subspace definitions to measure effective-dimensionality using an unclamped participation-ratio spectrum and the findings show that participation-ratio on open-ended prompts is far higher than it is on prompts with defined, single token expected answers.

In Sections 5–6 the rank of each subspace — how many directions D and S span — is set with participation ratios clamped to build well-defined bases for the subspace interventions. We set D dimensionality, $k_D$, with participation-ratio clamped to $[2,12]$ and S dimensionality, $k_S$, clamped to $[4,16]$. The full definitions are outlined in Appendix A.3.

\textbf{Gradients and stratification are inferred from binned analysis.} Throughout the paper, references to a "gradient" or "stratification" along the prediction-proximity axis describe patterns inferred from PDSF's four-bin partition rather than from continuous within-geometry sampling. PDSF partitions the residual stream into four orthogonal subspaces at chosen rank cut-offs; properties that vary monotonically across these bins (manifold complexity, intervention sensitivity, divergence timing) are reported as gradient-like. While the work reported in this paper did not directly measure the continuous geometry between bin boundaries, our assumption is that they measure underlying gradients.

\textbf{What PDSF does and does not impose.} PDSF imposes two properties by construction: orthogonality to the argmax unembedding direction $\hat{\mathbf{u}}$, and a sequential variance ordering $D > S > F$ within the orthogonal residual. The empirical findings reported below — low effective rank, scale-invariance across model size, near-orthogonality between the prediction direction and principal variance axes, a manifold-complexity gradient steeper than matched PCA controls, F's anti-discriminative sign inversion, and the causal and behavioral hierarchies observed under intervention — are not entailed by the construction; a prediction-anchored decomposition could equally have produced a widening interface, a PCA-equivalent gradient, or an intervention hierarchy inconsistent with prediction proximity. §6.4 sharpens the point: an organized sub-basis within F (F-topK) produces persistent rotational scrambling behavior in the same regime as S, showing that the across-bin ordering carries past the analytical S/F boundary rather than terminating at it. PDSF should be read as an analytical coordinate system for measuring residual-stream geometry, not as a claim that the model internally partitions into four modules; the methodological case against variance-anchored alternatives is developed in Appendix B.3.

\phantomsection\addcontentsline{toc}{subsection}{3.1 Measurement in the PDSF coordinate system}
\subsection*{3.1 Measurement in the PDSF coordinate system}

Throughout the paper, we report geometric measurements made on the same underlying dataset: per-prompt final-token hidden states, extracted at every layer and projected onto the relevant PDSF subspace, so each prompt contributes one point per subspace per layer. Most geometric quantities (effective dimensionality, manifold complexity, prediction–variance orthogonality) are computed at every layer and reported as full-depth trajectories, but a few summary statistics aggregate over a representative subset of depths. Where a single-layer measurement is used (e.g., cosine discrimination) the layer is stated at the point of use.

Analyses use three prompt sets, each designed to expose a different kind of variation among prompts.

\textbf{SpecA} (224 prompts) isolates surface-form variation while holding meaning fixed. It comprises 14 semantic groups of 16 variants, where the 16 variants form a 2⁴ full factorial over four surface factors — paraphrase, answer-constraint phrasing, a clutter preamble, and output-format instruction — applied to a question whose meaning and single-token answer are held constant. Eleven groups have binary answers (Yes/No or True/False) and three are multiple-choice (A/B/C or A/B/C/D); the groups cover arithmetic (4), logic (3), factual recall (4), and linguistics (3). Holding the answer fixed while varying only surface form is what lets P, D, and S separate cleanly: P is constant across a group, D tracks surface form, and S encodes which question is being asked.

\textbf{SpecB} (96 prompts) probes open-ended generation rather than constrained answering. It comprises 12 semantic categories of 8 variants each; each prompt is a short narrative opener that the model continues for 64 tokens. The canonical set defines a nested 80-prompt \texttt{standard} tier — the first 10 categories — which is the tier used for the persistent-intervention experiments of §6, while the geometry of §4 is computed on the full 96. Unlike SpecA, SpecB has no factorial structure — the eight variants in a category are genuinely different prompts that share a category (Emotional State, Decision/Choice, Goal-Driven Action, Perspective/Voice, Causal Setup, Temporal Anchor, Genre/Register, Social Role, Evaluative Stance, Physical Setting, and the two categories outside the standard tier, Epistemic State and Relationship Dynamic) but not a storyline, so there are no factor levels to cross. The set exists to test whether the PDSF decomposition, and D/S scrambling in particular, has causal purchase on free-form continuation, where no single-token target constrains the trajectory.

\textbf{Diverse} (84 prompts) tests whether the complement F behaves the same way across fundamentally different processing modes. It comprises 21 groups of 4 variants, distributed across eight regimes: English narrative, English analytical, Romance-language narrative (Spanish, French, Portuguese), East-Asian prompts (Mandarin, Japanese, Korean), code completion (Python, JavaScript, Rust), formal mathematics, structured instructions, and an unusual-register group whose eight variants are each a different register (legal, archaic, stream-of-consciousness, academic abstract, medical case report, patent application, sports commentary, and cooking-show transcript). Each variant is written to elicit a 64-token continuation.

Full model details, intervention protocols, and representative example prompts for each set are provided in Appendix A; example prompts appear in Appendix A.2.

\textbf{Effective dimensionality.} For each subspace at a given layer, the per-prompt projections (one point per prompt) form a point cloud whose empirical covariance matrix has eigenvalues $\lambda_i$. We report six effective-rank estimators (Roy \& Vetterli, 2007) operating on this spectrum: participation ratio $\mathrm{PR} = (\sum_i \lambda_i)^2 / \sum_i \lambda_i^2$, stable rank $\|A\|_F^2 / \|A\|_{\text{op}}^2$, spectral entropy rank $\exp(-\sum_i p_i \log p_i)$ with $p_i = \lambda_i / \sum_j \lambda_j$, and the number of eigendirections required to capture 90\%, 95\%, and 99\% of the total variance. Each estimator reports a soft count of how many directions of the covariance carry comparable variance. Because different estimators weight the spectrum's tail differently, we treat unanimous ordering across estimators as strong evidence of robustness (§4.1, Figure 2C).

\begin{quote}\small
\textbf{Note — dimensionality measurements are not fixed by the construction:} \\
$D_{PR}$ and $S_{PR}$ report the participation ratio of the discriminative and situational subspaces as constructed (the top-$k_D$ and top-$k_S$ PCA projections of the P-residual and double-residual) where $k_D$ and $k_S$ are set adaptively by the participation ratio of the underlying residual and bounded well above the values either prompt set reaches, so the bound does not constrain the reported measurements. Because $k$ tracks each residual's own spectrum rather than a fixed cut, these measures respond to the true shape of the prediction-proximal spectrum. If that spectrum were genuinely high-rank, $D_{PR}$ and $S_{PR}$ would rise with $d_{\text{model}}$, but they do not. The scale-invariance reported in §4.1 is therefore a property of the residual stream's prediction-proximal spectrum, not of the rank-selection step.
\end{quote}

\textbf{Cosine discrimination.} For each PDSF subspace, we compute pairwise cosine similarities between per-prompt projections and report Cohen's $d$ between within-group and between-group distributions (where prompt groups are the factorial design categories described in §3.1; within-group: prompts from the same group; between-group: prompts from different groups). Higher $d$ indicates that the subspace separates prompt groups by direction.

\textbf{kNN/global distance ratio.} For each subspace at a given layer, the per-prompt projections (one point per prompt) form a point cloud. The metric is the ratio of the mean Euclidean distance from a point to its $k$-th nearest neighbor (local scale, $k = 8$) vs the mean pairwise Euclidean distance across all points in the cloud (global scale). Values near 1 indicate flat, uniform geometry; lower values indicate tightly folded structure where points are locally clustered relative to their global spread. Because lower values correspond to higher manifold complexity, the numerical ordering $D < S < F$ in this metric corresponds to complexity $D > S > F$. Used in §4.3 (Figure 3).

The full testbed contains eighteen models spanning six architecture families (7B–120B parameters), ten instruction-tuned and eight base variants; model details are in Appendix A.1.

\FloatBarrier
\phantomsection\addcontentsline{toc}{section}{4. Geometric Stratification Along the Prediction-Proximity Axis}
\section*{4. Geometric Stratification Along the Prediction-Proximity Axis}

If the linear-readout constraint holds, residual-stream geometry should organize relative to the prediction direction. We test four geometric consequences: dimensional invariance (§4.1), variance/readout separation (§4.2), manifold-complexity stratification (§4.3), and group-discrimination inversion (§4.4). We then test whether this geometric organization precedes instruction tuning (§4.5).

\phantomsection\addcontentsline{toc}{subsection}{4.1 The prediction interface is low-dimensional and scale-invariant}
\subsection*{4.1 The prediction interface is low-dimensional and scale-invariant}

The prediction-interface constraint predicts that the dimensionality of prediction-proximal structure should be set by readout and task demands rather than by ambient model width. If the interface scaled with model width, larger models would have more prediction-facing slots for readout-relevant information. However, the fixed nature of the linear readout from the residual stream to the ArgMax distribution gives no architectural reason to expect such expansion. We tested this question by measuring the effective rank of the prediction-interface across eighteen models at five layer depths on the constrained (SpecA) and open-ended (SpecB) prompt sets.

The results show that the effective dimensionality of the prediction-proximal subspaces does not scale with model size (Figure 2; Table 1). D and S remain narrow despite large differences in hidden dimension, parameter count, and architecture. Across all models, at five layer depths, on both the constrained (SpecA) and open-ended (SpecB) prompt sets, using the six effective-rank estimators defined in §3.1, the dimensionality of the prediction-proximal subspaces remain constant. These measurements quantify how much dimensional capacity models allocate near the prediction interface, but they do not attempt to identify constraints on how much semantic meaning can be carried by the measured subspaces.

\begin{figure}[tbp]
\centering
\includegraphics[width=\linewidth]{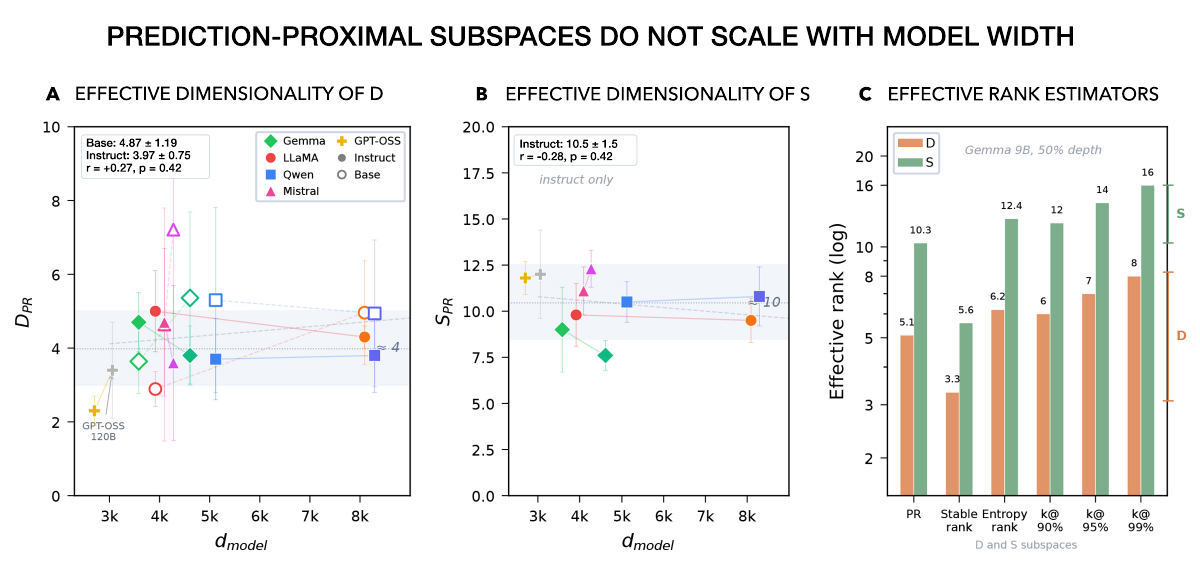}
\caption{\textbf{Prediction-proximal subspaces do not scale with model width.} \textbf{(A)} Effective dimensionality of the discriminative subspace ($D_{PR}$, participation ratio) vs. hidden dimension ($d_{\text{model}}$) for 18 models spanning 2,880–8,192 hidden dimensions across 6 architecture families. Filled markers: instruction-tuned models (mean 3.97 ± 0.75); open markers: base models (mean 4.87 ± 1.19). Error bars indicate standard deviation across 5 layer depths. The shaded band marks $D_{PR}$ ≈ 3–5. Neither training regime shows significant correlation with $d_{\text{model}}$ ($r$ = +0.27, $p$ = 0.42 for instruct). GPT-OSS 120B, the largest model by parameter count (120B) but with the narrowest residual stream in the testbed (2,880 dimensions — an MoE architecture whose capacity derives from 128 expert pathways rather than width), maintains $D_{PR}$ = 3.2. Within-family pairs are connected (solid: instruct, dashed: base); no family shows systematic expansion. \textbf{(B)} The situational subspace $S_{PR}$ is also stable across $d_{\text{model}}$ at 10.5 ± 1.5 for 10 instruct models ($r$ = −0.28, $p$ = 0.42). Base $S_{PR}$ data not shown. \textbf{(C)} Six independent effective rank estimators (participation ratio, stable rank, entropy rank, and three explained-variance thresholds) unanimously place D in single digits and S in single digits to low teens, confirming the two-step prediction-proximal dimensionality ordering $D \ll S$ is estimator-independent. F (not shown) falls in the tens to hundreds on every estimator. The estimators support the full ordering of $D \ll S \ll F$. Representative model: Gemma 9B instruct at 50\% depth.}
\end{figure}

\begin{table}[htbp]
\centering
\footnotesize
\caption*{Table 1. Prediction-proximal dimensionality across 10 instruction-tuned models (SpecA and SpecB, mean ± std across 5 depths).}
\setlength{\tabcolsep}{4pt}
\begin{adjustbox}{max width=\linewidth}
\begin{tabular}{lrcccc}
\toprule
Model & $d_{\text{model}}$ & $D_{PR}$ (SpecA) & $S_{PR}$ (SpecA) & $F_{PR}$ (SpecA) & $D_{PR}$ (SpecB) \\
\midrule
Gemma 9B & 3,584 & 4.7 ± 0.8 & 9.0 ± 2.3 & 34.0 ± 7.6 & 9.3 ± 3.2 \\
Mistral 7B & 4,096 & 4.7 ± 2.0 & 11.1 ± 1.3 & 44.0 ± 5.6 & 15.5 ± 2.8 \\
Llama 8B & 4,096 & 4.0 ± 1.1 & 9.8 ± 1.7 & 40.0 ± 6.6 & 9.3 ± 2.6 \\
Mixtral 8×7B & 4,096 & 5.0 ± 2.1 & 12.3 ± 1.0 & 61.1 ± 22.2 & 23.4 ± 12.1 \\
Gemma 27B & 4,608 & 3.8 ± 0.8 & 7.6 ± 0.8 & 33.0 ± 3.7 & 5.9 ± 2.4 \\
Qwen 14B & 5,120 & 3.7 ± 1.1 & 10.5 ± 1.1 & 45.7 ± 3.0 & 14.3 ± 1.1 \\
GPT-OSS 20B & 2,880 & 2.3 ± 0.4 & 11.8 ± 0.9 & 48.2 ± 9.1 & 17.5 ± 7.0 \\
Llama 70B & 8,192 & 4.0 ± 0.3 & 9.5 ± 1.2 & 36.0 ± 2.6 & 10.7 ± 3.4 \\
Qwen 72B & 8,192 & 4.3 ± 1.0 & 10.8 ± 1.6 & 42.5 ± 7.4 & 12.8 ± 1.1 \\
GPT-OSS 120B & 2,880 & 3.2 ± 0.4 & 12.5 ± 0.9 & 64.7 ± 16.1 & 17.4 ± 7.4 \\
\bottomrule
\end{tabular}
\end{adjustbox}
\end{table}

\emph{SpecB $D_{PR}$ per-model values are means across 5 depths.}

Across the SpecA prompt set the effective dimensionality of D ($D_{PR}$) is consistently single digit (3.97 ± 0.75 across all 10 instruction-tuned models). The low dimensionality is in stark contrast to the hidden dimension of the models which varies 2.8× across the testbed (2,880 to 8,192). $D_{PR}$ varies less than 2.2× but shows no correlation with $d_{\text{model}}$ ($r$ = +0.27, $p$ = 0.42). The effective dimensionality of S ($S_{PR}$) is similarly stable at 10.5 ± 1.5 ($r$ = −0.28, $p$ = 0.42).

On the open-ended SpecB prompt set, $D_{PR}$ is systematically higher and more variable across models (mean 13.6, range 5.9–23.4) than on SpecA (Figure 2)\textsuperscript{(specb96)}. We assume the greater variability reflects the larger prediction-proximal spectra that semantically diverse continuation prompts induce. The increase in effective dimensionality is largest in the mixture-of-experts models — Mixtral 8×7B (23.4) and the two GPT-OSS models (17.4–17.5), which is consistent with expert routing sustaining additional discriminative structure in the shared residual stream (§4.5). What does not change is the relationship to model width: no within-family pair widens with scale. Gemma 9B→27B falls (9.3→5.9), Qwen 14B→72B falls (14.3→12.8), Llama 8B→70B is essentially flat (9.3→10.7), and the two GPT-OSS models — which share an identical 2,880-dimensional residual stream at a 6× parameter difference — land together (17.5 and 17.4). Across the ten models $D_{PR}$ shows no correlation with $d_{\text{model}}$ ($r$ = −0.29).

Convergent evidence comes from a different measurement tradition. Cheng et al. (2023) report that the intrinsic dimension of OPT representations "does not significantly change" as the hidden dimension doubles from 1,024 to 4,096 across OPT-350m, 1.3b and 6.7b, recovering values of order 10 in every case. That measurement concerns the whole representation rather than the prediction-proximal region, and uses a nearest-neighbour intrinsic-dimension estimator rather than a participation ratio over a linear projection; the agreement across both of those differences is what makes it informative here.

In short, the comparison of constrained versus open-ended prompts shows that the dimensionality of the prediction interface is set by the demands of the prompt distribution — how many distinguishable items it asks the model to separate — not by the capacity of the model.

\textbf{Estimator independence:} We tested our assumption of the use of PR as a measure of dimensionality with six independent rank estimators. All agree unanimously on the rank ordering of $D \ll S \ll F$ across all models (Figure 2C; per-estimator values in Appendix C).

\textbf{Scale-invariance:}  Measurements taken on base-model confirm that the constraint originates in the readout architecture rather than instruction tuning. Across 8 base models effective dimensionality, $D_{PR}$, ranges from 2.89–7.21 on SpecA prompt with no systematic expansion across \emph{d}\textasciitilde{}model\textasciitilde{}. Likewise, within-family base pairs (Llama 8B→70B, Qwen 14B→72B, Gemma 9B→27B) show no consistent widening with scale. Full base-model dimensionality data and within-family comparisons are reported in Appendix C (Tables C.1–C.2).

In contrast to the subspaces in the prediction-interface, F's effective dimensionality $F_{PR}$ varies substantially across models, from 33 to 65 on SpecA. F always spans the full residual complement, $d - \operatorname{rank}(P) - k_D - k_S$ ambient directions, which is thousands of dimensions in every model and grows linearly with \emph{d}\textasciitilde{}model\textasciitilde{}. F's effective rank does not grow with model width. The two widest models (Llama 70B and Qwen 72B at $d$ = 8,192) sit near the middle of the range at 36.0 and 42.5, while the two narrowest (the GPT-OSS models at $d$ = 2,880) sit at its top. So added width enlarges F's ambient capacity without converting into proportionally more active directions.

Read together with the interface measurements, this yields a single statement about the whole residual stream: width buys ambient room, not effective dimensionality, anywhere. The prediction-proximal subspaces stay narrow because the readout is fixed, and the high-dimensional background absorbs the dimensional growth without activating it. This is not an artifact of a saturated rank estimator. The same participation-ratio measure moves from $D_{PR}$ ≈ 4 on SpecA to ≈ 14 on SpecB and reaches 65 in F, so it has ample dynamic range and responds strongly to the prompt distribution; it simply does not respond to model width. A dense 7B model ($d$ = 4,096) and an MoE 120B model ($d$ = 2,880) hold the same low prediction-proximal dimensionality on matched prompts. Neither larger residual width nor mixture-of-expert pathways widen the prediction interface.

\phantomsection\addcontentsline{toc}{subsection}{4.2 Prediction alignment is near-orthogonal to the principal variance axes}
\subsection*{4.2 Prediction alignment is near-orthogonal to the principal variance axes}

If the complement holds its variance at low readout alignment, the prediction direction should be largely decoupled from the dominant variance axes — high variance concentrated in directions orthogonal to the prediction direction rather than suppressed. We test this by measuring the angle between the prediction direction and the principal variance axes at each layer.

Our tests show that averaged across the instruction-tuned testbed, the prediction direction is near-orthogonal to the principal variance axes on all three prompt sets. Averaged over network depth, the mean principal angle across the ten instruction-tuned models is 83.8° ± 1.6° on SpecA, 86.7° ± 0.6° on SpecB\textsuperscript{(specb96)}, and 85.8° ± 1.7° on Diverse. The angle is largest in the early and middle layers and relaxes toward the readout. At the final layer the difference between the prediction direction and the principal variance axes falls to roughly 73° on SpecA, 82° on SpecB, and 81° on Diverse as the prediction direction sharpens, but even there the prediction basis stays far from the dominant variance directions. The cross-model spread is narrowest on the open-ended set, where all ten models fall within 85.4–87.8°; on Diverse most models sit within about 2° of the mean, with the largest MoE (GPT-OSS 120B) being the lowest at 82.3°. So the near-orthogonality is a robust population tendency across constrained answering, open-ended continuation, and heterogeneous regimes alike, rather than a fixed constant. These results align with the geometric signature that is expected if models are using orthogonality to separate the readout variance from variance in computational areas that are not part of the next token choice.

We propose that this variance isolation is a consequence of the SNR constraint introduced in §2. For the unembedding to cleanly resolve the selected token, prediction-relevant variance must be isolated from the high-dimensional computational background. D and S carry high variance but lie orthogonal to P by construction. In other words, these directions hold the cross-prompt discriminative structure while staying off the selected-token axis.

F occupies the prediction-distal complement. A high-variance F aligned with the directions relevant to the selected token and its near-competitors would project structured variance into the logits and erode the margin. Instead, models structure variance in F across directions that move the top tokens only weakly, so it can carry large variance without degrading the readout (Appendix B.2).

\phantomsection\addcontentsline{toc}{subsection}{4.3 Manifold complexity decreases with prediction distance}
\subsection*{4.3 Manifold complexity decreases with prediction distance}

A prediction interface with narrow dimensional structure has implications for information packing. If prediction-proximal subspaces occupy only a narrow window of effective dimensions, one might reasonably expect to find a higher information density per dimension. Near the interface, discriminative structure, semantic trajectory, and local response specification must be packed into relatively few dimensions, and such packing pressure would predict a more folded or locally structured manifold geometry.

Farther from the interface, the complement has many more dimensions over which to distribute information, which should have reduced pressure and thus flatter geometry. We test this hypothesis by assessing manifold complexity across subspaces using the kNN/global distance ratio defined in §3.1, and compare it to the gradient recovered under simple variance ordering. Lower values correspond to higher manifold complexity (tightly folded structure); values near 1 correspond to flat, uniform geometry. The numerical ordering $D < S < F$ in this metric therefore corresponds to a complexity ordering of $D > S > F$, where D is the most tightly structured and F the flattest.

Across all models evaluated, we observe a monotonic ordering of manifold complexity as a function of prediction proximity (Figure 3): prediction-proximal subspaces have high manifold complexity (tightly folded, structured geometry), while prediction-distal subspaces have low manifold complexity (flat, uniform geometry). The same ordering appears across prompt sets and architectures, and the gradient is substantially steeper than what variance-ranked decomposition alone can reveal.

On the Diverse prompt set (84 prompts spanning 8 linguistic regimes), this ordering holds at the final layer in all ten instruction-tuned models, unanimously for both D < S and S < F; the same unanimity holds across all three prompt sets (30 of 30 model × prompt-set cells). At the layer-observation level (cross-depth aggregate; basis for Figure 3 panel C), D < S holds at 100\% and S < F at 99.1\%.

This gradient is distinct from the depth-dependent geometric phases documented in prior work. Cheng et al. (2025) identify a high-dimensional abstraction phase across intermediate layers; the PDSF gradient measures complexity stratification across subspaces at each depth, not across depths within a single subspace.

\begin{figure}[tbp]
\centering
\includegraphics[width=\linewidth]{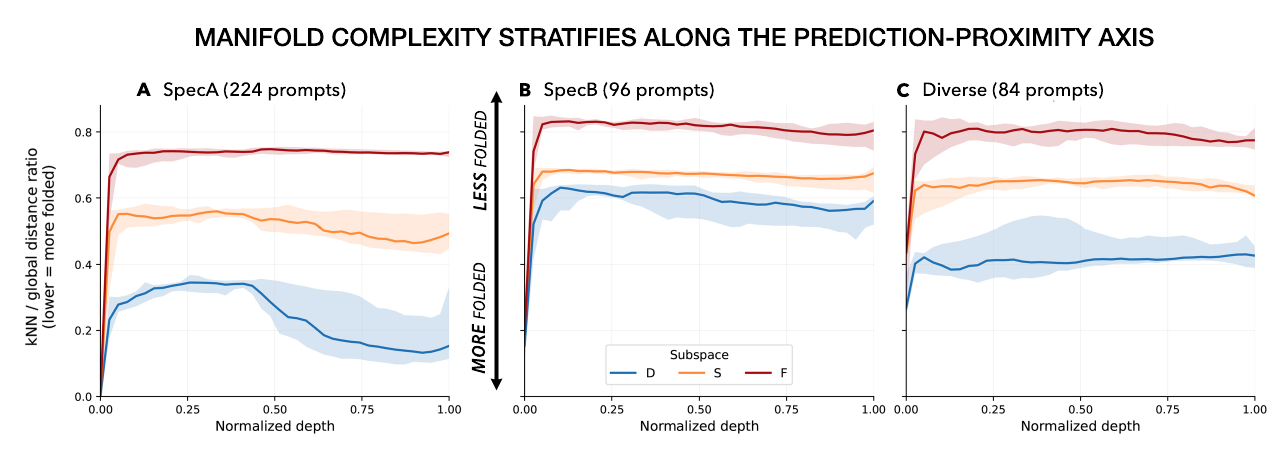}
\caption{\textbf{Manifold complexity stratifies along the prediction-proximity axis across all three prompt regimes.} kNN / global distance ratio (k = 8) by normalized depth for each PDSF subspace (D, S, and F), median across the 10 instruction-tuned models with 10–90 percentile band envelope. Lower values correspond to higher manifold complexity (tightly folded local structure), and values near 1 correspond to flat, uniform geometry. \textbf{(A)} SpecA (224 prompts, 14 factorial groups). \textbf{(B)} SpecB (96 continuation prompts, 12 semantic-family groups). \textbf{(C)} Diverse (84 prompts spanning 8 linguistic regimes: narrative, analytical, code, formal math, multilingual, structured instructions, unusual register).  The three subspaces form cleanly separated bands across all three regimes. The ordering of D < S < F in the kNN ratio metric (equivalent to manifold complexity D > S > F) holds at 100\% (D < S) and 99.1\% (S < F) of layer-observations on Diverse. The same ordering is robust across SpecA and SpecB, and holds in all ten models at the final layer on every set. F is depth-invariant in all three panels from \textasciitilde{}10\% depth onward (band 0.70–0.76 in A, 0.74–0.84 in B and C). A rank-matched variance-ordered control recovers a mean 0.80 of the PDSF gradient on SpecA, 0.71 on SpecB, and 0.46 on Diverse, and is shallower than the prediction-anchored gradient in all thirty model × prompt-set cells. The difference between the PCA controls and the prediction anchored measures grow with the linguistic heterogeneity of the prompt set. The SpecB control is computed on the full 96-prompt continuation set rather than the 80-prompt narrative subset plotted here.}
\end{figure}

To test whether this gradient is a generic consequence of low-rank partitioning, we constructed a control decomposition matched to PDSF in rank but ordered by total variance rather than by prediction proximity. In every one of the thirty model × prompt-set cells the prediction-anchored gradient is steeper than the rank-matched variance-ranked control. The size of the margin, however, depends on how heterogeneous the prompt set is. On the constrained factorial set the control recovers most of the gradient (mean PCA/PDSF ratio 0.80, range 0.62–0.94); on the continuation set somewhat less (mean 0.71, range 0.62–0.86)\textsuperscript{(specb96)}; and on the diverse set under half (mean 0.46, range 0.12–0.62), where the prediction-anchored gradient runs 1.6 to 8.1 times steeper. In short, variance ranking recovers much of the stratification when prompts are near-identical in form, and progressively less as they diverge. What this means is that most of the geometric reorganization of the residual stream that differentiates prompts happens along the prediction axis rather than along the primary axes of variation. Prediction anchoring shows the most differences where the prompts share the least. The control rank is set by the participation-ratio-adaptive rank of P and varies from 30 to 50 on the diverse set. Because of this the cross-model means could in principle be confounded with bin size, but re-running every model at a common control rank of 31 returns 0.78, 0.67, and 0.43 for the three sets, within 0.02–0.04 of the per-model-matched values and preserving every ordering. The observed stratification is therefore not a generic consequence of low-rank structure. Because the control differs from PDSF in both its ordering criterion and its bin construction, it does not isolate the contribution of the anchor itself; what it establishes is that ordering by prediction proximity separates the strata more sharply than ordering by variance does, and increasingly so as prompts diverge in form. The geometric stratification is not a shadow of the variance structure; it is an independent organizational axis.

\textbf{The manifold gradient continues into F.} The complexity gradient resolves more finely within F itself: F-topK — the next nested step in the prediction-anchored decomposition taken inside F (rank 28–64 across models on SpecA, 28–47 on Diverse) — is more folded than bulk F on SpecA (kNN/global ratio $0.71 \pm 0.05$ vs. $0.78 \pm 0.03$; 10/10 instruction-tuned models) and on Diverse ($0.75 \pm 0.05$ vs. $0.78 \pm 0.06$; 9/10), and never inherits bulk F's flat population-level signature. This indicates that the across-bin manifold-complexity ordering continues inside F once F is resolved into sub-basis structure. This is consistent with PDSF's bins acting as measurement cuts whose across-bin ordering is detectable at finer sub-basis resolution, rather than as functional barriers.

Our results show that the manifold-complexity gradient is related to variance but is not directly caused by variance. D must pack fine semantic distinctions among many prompt groups into a small number of effective dimensions — a tight packing requirement that forces the manifold to fold sharply (low kNN/global ratio).

F spans thousands of dimensions and bears no such constraint: the information F carries can distribute across many dimensions without needing to concentrate, so no tight folding is required and the manifold stays flat. The most parsimonious explanation for this complexity gradient is that is an information-packing consequence of the dimensionality allocation that the SNR constraint produces, not a direct consequence of SNR itself. Together with the variance-landscape result in §4.2, manifold complexity and variance isolation form two independent structural properties of the prediction-proximity axis, each running in the same direction for different reasons. The PCA control separates the two candidate causes of the complexity gradient. A gradient that is recoverable from variance ranking alone should reflect only the variance-landscape cause, because variance ranking has no access to the prediction interface. The additional steepness recovered by prediction anchoring likely reflects the information-packing cause, and it does so most strongly exactly where packing pressure is most varied (up to 8.1× on the most heterogeneous prompt set).

\phantomsection\addcontentsline{toc}{subsection}{4.4 F anti-discriminates among prompt groups}
\subsection*{4.4 F anti-discriminates among prompt groups}

Our prompt sets were designed to test how residual-stream geometry varies within and between semantically defined groups of prompts (§3.1). Given this design, and given that D captures the largest cross-prompt variance directions, we expected group discrimination to weaken with prediction distance: D should carry the strongest group structure, S less, and F little or none. Instead, the discrimination structure reversed at the boundary of S and F. F is not merely weakly discriminative — it anti-discriminates: prompts from the same group are systematically farther apart in F-space than prompts from different groups.

\begin{figure}[tbp]
\centering
\includegraphics[width=\linewidth]{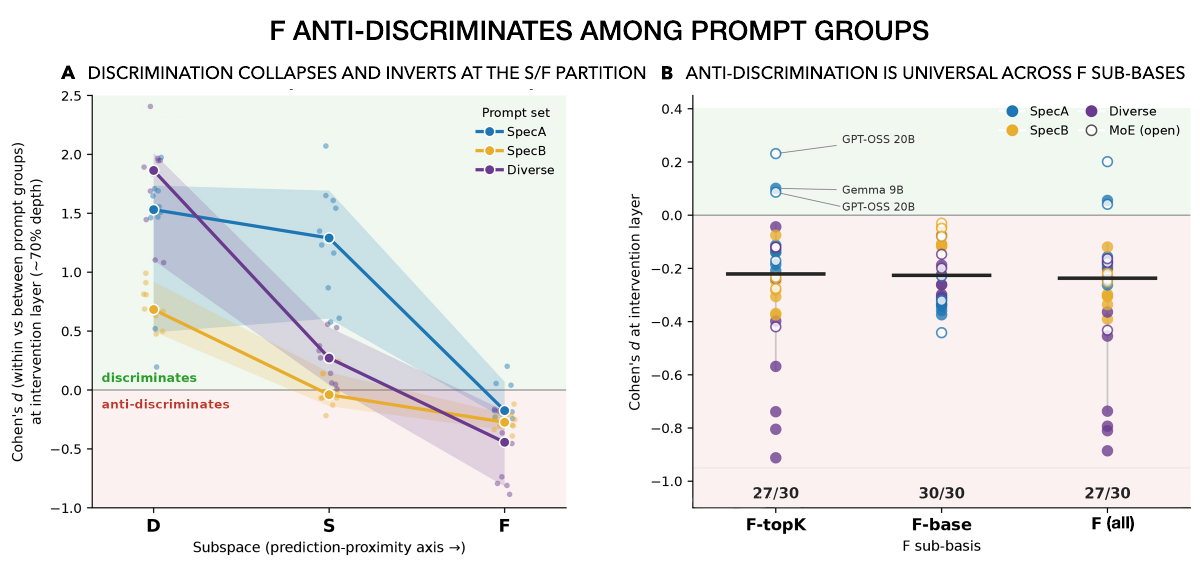}
\caption{\textbf{F anti-discriminates among prompt groups, and the inversion is universal across F's internal structure.} The figure plots Cohen's $d$, a measure of discrimination, for each PDSF subspace. Cohen's $d$ is computed between within-group and between-group pairwise cosine-similarity distributions of per-prompt projections in each subspace at the intervention layer (\textasciitilde{}70\% network depth). Positive values mean that same-group prompts are closer in angle than between-group prompts (discriminate), and negative values mean that same-group prompts are farther apart (anti-discriminate). \textbf{(A)} Per-prompt-set median across 10 instruction-tuned models with 10–90 percentile band envelope, traversing the ordered subspaces D → S → F (taking F as a single unit alongside the sub-bases F-topK and F-base). All three prompt sets begin in discriminative territory at D, decline through S, and end in anti-discriminative territory at F. The SpecB band dips below zero already at S — the layer × prompt-set effect characterized below. \textbf{(B)} Cell-level Cohen's $d$ for the three F-level measurements (F-topK, F-base, and the full F) at the intervention layer; each sub-basis has 30 cells (10 models × 3 prompt sets). F-base anti-discriminates in all 30 cells; F-topK and the full F each in 27 of 30. The three positive cells in F-topK and the full F all sit on SpecA in three specific models (Gemma 9B, GPT-OSS 20B, GPT-OSS 120B; labeled in panel B). The thick black bar is the median across 30 cells; the thin grey vertical line is the 10–90 percentile range. F-dynamic — a separate decomposition introduced in §7 — produces the same sign inversion in all 10 SpecB cells (10/10) and is not shown here.}
\end{figure}

This finding of anti-discrimination is surprising and can be interpreted as evidence that F carries prompt-specific configuration rather than group-level clustering. Same-group prompts can reach similar logit behavior through different prediction-proximal states; under this reading, F supplies compensating residual structure that makes those distinct states compatible with similar readout outcomes. This interpretation is not uniquely determined by the data (its scope is discussed in Appendix F; the readout-compensation account is developed in Appendix B.4). The narrower empirical claim is already strong: the complement is not merely weakly discriminative or unstructured with respect to prompt groups. It is systematically organized against the group axis expressed near the prediction interface.

The few exceptions are narrow and informative. The three positive F-topK and full-F cells all occur on SpecA, in Gemma 9B, GPT-OSS 20B, and GPT-OSS 120B. F-base remains negative in all three cases. These three exceptions are confined to the highest-variance portion of F on the least semantically diverse prompt set, not to the complement as a whole.

Another informative result is the contrast between the cosine discrimination results of S versus F-topK presented here and the behavioral intervention results on S versus F-topK presented in §6.4. Here, we show that these subspaces exhibit different geometries relative to prompt discrimination, but in §6.4 we find identical behavioral results.

Full per-prompt-set values, the readout-compensation limit-case account for the outliers, MoE attenuation, magnitude × diversity scaling, and between-subspace comparisons are reported in Appendix B.4.

\begin{figure}[tbp]
\centering
\includegraphics[width=\linewidth]{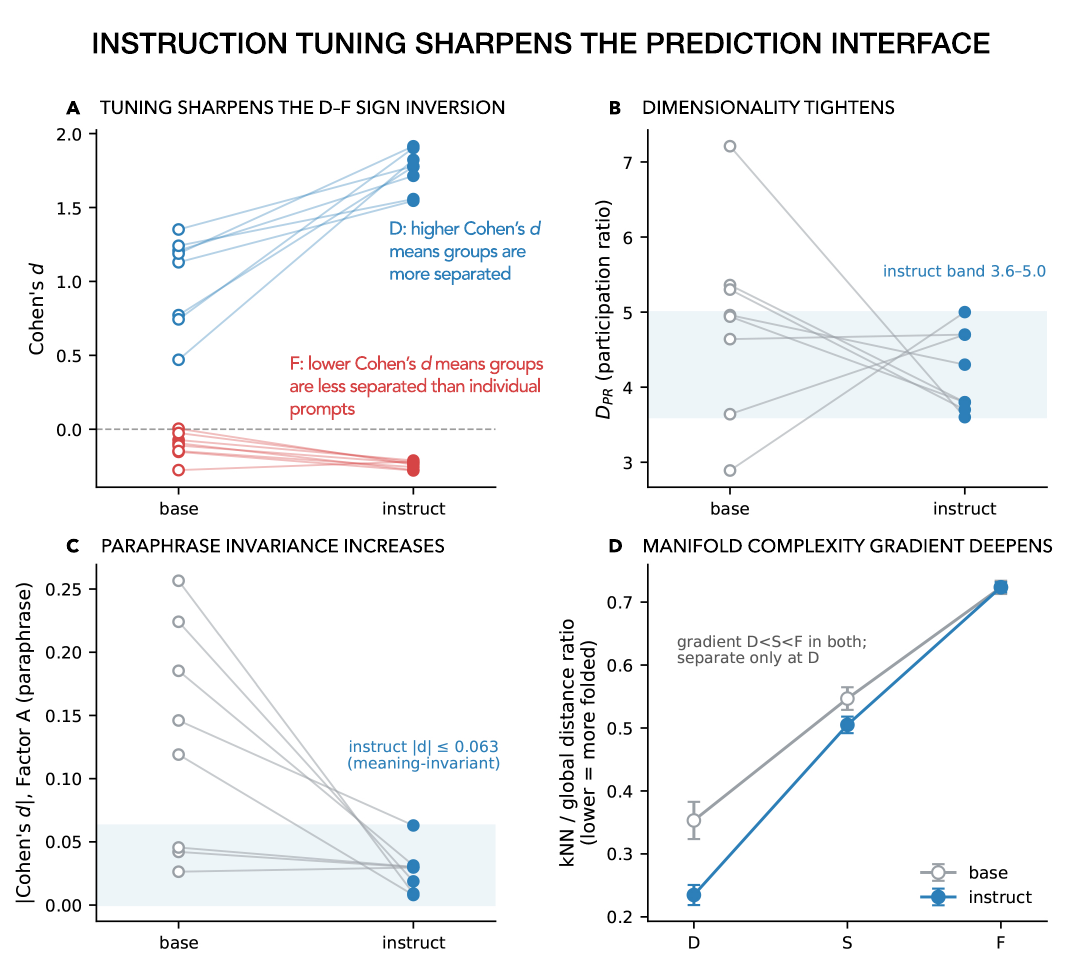}
\caption{\textbf{Instruction tuning sharpens the prediction interface without expanding it.} Comparisons of geometric measures across 8 matched base→instruct model pairs on SpecA; open markers are base, filled are instruction-tuned. \textbf{(A)} Instruction tuning sharpens the D–F sign inversion: the D subspace's Cohen's \emph{d} (group discrimination, §4.4) rises (blue) while F's deepens toward negative (red, anti-discrimination), so the sign inversion of §4.4 becomes more pronounced under tuning. Prompt groups are pulled apart in D and pushed together in F. \textbf{(B)} Effective dimensionality of the prediction interface does not expand: D\_PR (participation ratio) stays low and converges — base values scatter (2.89–7.21), instruct values cluster in the shaded 3.6–5.0 band, with no systematic shift. \textbf{(C)} Variance in response to paraphrasing goes down: the effect of the paraphrase factor (Factor A) on the D representation collapses from a scattered base range (up to |d| = 0.26) to a tight near-zero instruct band (|d| ≤ 0.063) — the interface comes to respond to meaning rather than surface wording. \textbf{(D)} Manifold complexity deepens at the prediction-proximal end of the prediction axis: the kNN/global distance ratio shows the D<S<F gradient in both groups (architectural), base and instruct separating only at D — the D manifold folds more tightly while S and F are unchanged.}
\end{figure}

\phantomsection\addcontentsline{toc}{subsection}{4.5 Base vs. instruction-tuned models}
\subsection*{4.5 Base vs. instruction-tuned models}

The preceding results show a narrow prediction interface and a stratified complement. We next asked whether this organization is created by instruction tuning or is already present in base models. Our results demonstrate that the prediction axis defined geometry is architectural and instruction tuning sharpens it.

Our base model measurements show the same qualitative organization as their instruction-tuned counterparts. On SpecA prompts the effective prediction-proximal dimensionality (D\_PR) in the base models is small, ranging from 2.89 to 5.36 across the eight models with no systematic expansion with model width (§4.1; Appendix C, Table C.1). The manifold-complexity gradient of §4.3 is also present in every base model. The prediction-proximal D subspace is the most folded and F the flattest (D<S<F in the kNN/global distance ratio). The F anti-discrimination is also present in base models. F anti-discriminates in 7 of the 8 base models, so the sign inversion of §4.4 is not created by tuning either.

Together these results indicate that the core geometry of the prediction interface does not originate in instruction tuning but is instead consistent with our hypothesis of a constraint imposed by readout architecture.

\paragraph{Instruction tuning sharpens the prediction interface}
Instruction tuning concentrates and regularizes prediction-relevant structure. Across matched base/instruction pairs, the instruction tuning increases Cohen's \emph{d} among prompts, compresses effective dimensionality in D, steepens the manifold complexity gradient, and makes the D geometry more invariant to paraphrase. So instruction tuning concentrates discriminative variance into fewer, cleaner directions within the same prediction-proximal subspace.

\textbf{Discrimination (Cohen's D):} Instruction tuning sharpens discrimination at both ends of the sign inversion. The D subspace separates prompt groups more distinctly after tuning: its Cohen's \emph{d} (the §4.4 measure) rises in all 8 pairs, from a scattered base range (0.47–1.35) to a tight, non-overlapping instruct band (1.55–1.91). The complement, F, moves the opposite direction. F's Cohen's \emph{d}, already negative in base models, tightens from a shallow base range (0.00 to −0.28) into a robust negative band in the instruct models (−0.21 to −0.28). The geometric result is that groups are pulled apart in the prediction-proximal subspace and pushed together in the complement, which steepens the discriminative gradient.

\textbf{Manifold complexity:} Manifold complexity deepens at the prediction-proximal end. The D-subspace kNN/global distance ratio drops in all 8 pairs (0.31–0.40 → 0.21–0.27, non-overlapping) while S changes little and F not at all. Tuning folds the D manifold more tightly without touching the complement, suggesting that instruction tuning packs more information into an existing low dimensionality interface.

\textbf{Paraphrase invariance:} Instruction tuning makes the model's response to paraphrasing less variable. Factor A is the paraphrase factor of the SpecA multifactor design. Its effect size on the D representation measures how much the geometry moves when a prompt is reworded but its meaning held fixed – a small value means the interface is insensitive to surface wording. In all 8 instruction-tuned models |d| ≤ 0.063, but the base model values reach 0.26. The models that begin paraphrase-sensitive (Gemma 27B, Mistral 7B, Qwen 72B) become more invariant, while those already invariant stay so. So tuning the interface focuses the model response to meaning rather than phrasing.

\textbf{Prediction interface dimensionality:} All the changes above happen without widening of the prediction interface. The effective dimensionality, $D_{PR}$, stays low and actually converges from a wider range in base models (2.89–7.21) to a lower range in instruct models (3.6–5.0). Additionally, the spectral rank, $k_D$, compresses into fewer directions in 7 of 8 model pairs with only one model, Gemma 27B, expanding from a spectrally degenerate base (Appendix C). $D_{PR}$ is the variance-weighted effective width and $k_D$ is the count of active directions. The two can move independently, but both show that instruction tuning concentrates the prediction interface rather than expanding it.

\begin{table}[htbp]
\centering
\footnotesize
\caption*{Table 2. Geometric convergence under instruction tuning across 8 matched model pairs.}
\setlength{\tabcolsep}{4pt}
\begin{adjustbox}{max width=\linewidth}
\begin{tabular}{lccc}
\toprule
Metric & Base Range & Instruct Range & Change \\
\midrule
Cohen's \emph{d}, D (discrimination) & 0.47–1.35 & 1.55–1.91 & non-overlapping; 8/8 ↑ \\
Cohen's \emph{d}, F (anti-discrimination) & 0.00 to −0.28 & −0.21 to −0.28 & 7/8 deepen \\
D\_PR (dimensionality) & 2.89–7.21 & 3.60–5.00 & converges, no systematic shift \\
Factor A (paraphrase) & \textbackslash{}textbar\{\}d\textbackslash{}textbar\{\} up to 0.26 & \textbackslash{}textbar\{\}d\textbackslash{}textbar\{\} ≤ 0.063 & collapses toward 0 \\
Manifold kNN-D (complexity) & 0.31–0.40 & 0.21–0.27 & non-overlapping; 8/8 ↓ \\
k\_D (spectral rank) \emph{[appendix]} & 2–18 & 4–8 & 7/8 compress \\
\bottomrule
\end{tabular}
\end{adjustbox}
\end{table}

\FloatBarrier
\phantomsection\addcontentsline{toc}{section}{5. Direction Over Magnitude: Testing the Thin-Shell Hypothesis in F}
\section*{5. Direction Over Magnitude: Testing the Thin-Shell Hypothesis in F}

The thin-shell premise (§2; Appendix B.1) holds that residual-state variation lies primarily in direction rather than magnitude. If residual states have little radial freedom, then changes in model behavior should depend more on where a component points than on how large it is. In this section we test that expectation directly in F, the prediction-distal complement: directional disruption of the complement should matter more than magnitude scaling.

In §5.1, we measure the single-pass intervention sensitivity of the PDSF subspaces to intervention and find a strong aggregate hierarchy, $F \gg D \approx S \gg P$. This makes F the natural target for the thin-shell test, because F contains almost all residual dimensions and carries the bulk of aggregate single-pass sensitivity despite the fact that D and S have higher per-dimension causal weight.

In §5.2, we isolate the carrier of F's effect by comparing two matched behavioral interventions: F-mix, which disrupts direction while approximately preserving norm, and F-attenuate, which reduces magnitude while preserving direction.

This experiment was designed to separate directional and radial disruption. We do not test translation, which is a third possible source of variation. Norm concentration constrains radius, not centroid, so the thin-shell argument does not by itself predict translation sensitivity, so we treat translation as outside the scope of the present test.

Throughout the paper we use "causal" in the interventionist sense. A component has a causal role when controlled perturbations to it systematically change outputs under the specified protocol. This does not identify a full causal graph or mechanistic circuit.

\phantomsection\addcontentsline{toc}{subsection}{5.1 Single-pass interventions locate aggregate sensitivity in F}
\subsection*{5.1 Single-pass interventions locate aggregate sensitivity in F}

Where does a one-time perturbation of residual geometry produce the largest next-token distributional displacement? We applied subspace-specific interventions at a single early layer (\textasciitilde{}12.5\% depth) and measured the resulting KL divergence from the unperturbed next-token distribution. This protocol does not measure long-horizon behavioral control. The focus of this test is to measure the immediate distributional effect of perturbing each analytical subspace once, under otherwise identical conditions.

\begin{figure}[tbp]
\centering
\includegraphics[width=\linewidth]{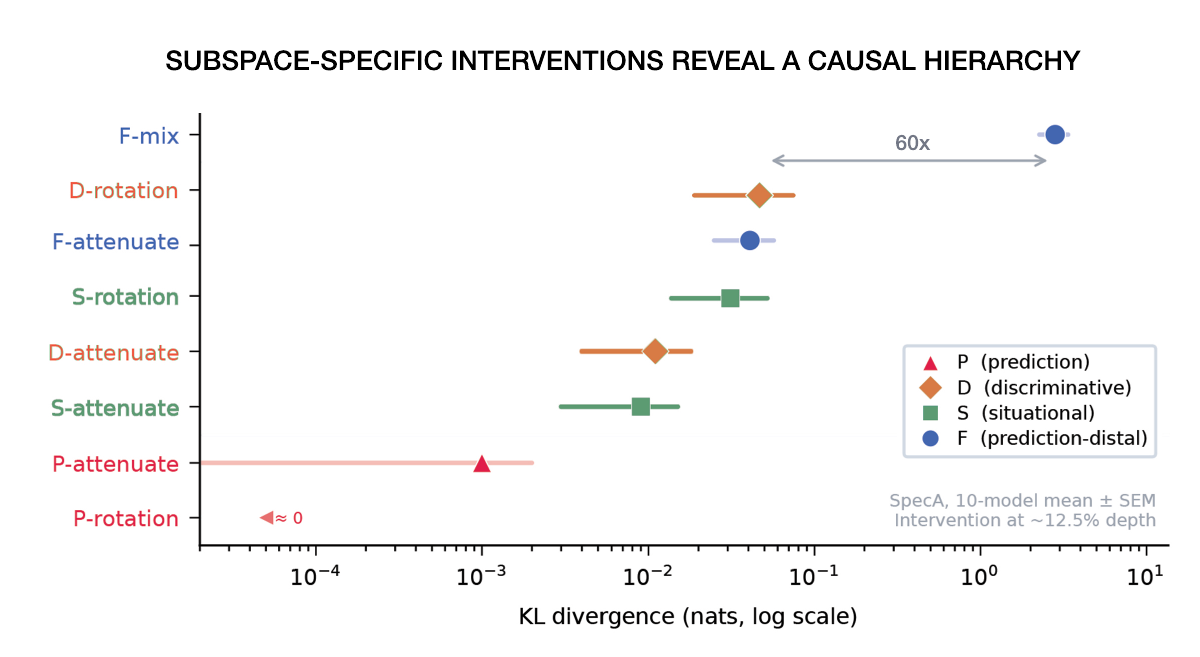}
\caption{\textbf{Subspace-specific interventions reveal a causal hierarchy spanning five orders of magnitude: $F \gg D \approx S \gg P$.} Single-pass KL divergence (nats, log scale) for 8 perturbation conditions applied at a single early layer (\textasciitilde{}12.5\% depth) on SpecA prompts; 10-model mean with SEM bars. Conditions are ordered by descending KL and color-coded by target subspace. F-mix produces a mean KL of 2.8 nats (median 2.5, range 1.0–7.2 across models), 50–1,000× larger than any D or S intervention. Within the F subspace, F-attenuate has an effect that is approximately 60× below F-mix, which shows a strong separation between magnitude-only and direction-affecting perturbations on F. D and S conditions are mixed in the 0.005–0.052 nats range, and F-attenuate falls within this D/S band. P-rotation (rank-1 null control) produces KL = 0.000 universally. Per-condition values with SEMs in Appendix A, Table A.4-1.}
\end{figure}

Interventions on F produce effects 50–1,000× larger than interventions on D or S (Figure 6). F dominates this hierarchy because it spans almost the entire residual stream; the effect is aggregate, not per-dimension. Small distributed changes across thousands of F directions can exceed complete randomization of a narrow D or S basis. F-attenuate, which scales thousands of F dimensions by 0.5, and D-rotation, which fully randomizes D's approximately four dimensions, produce indistinguishable KL (0.047 vs. 0.052 nats). Thousands of magnitude-only perturbations in the complement buy the same immediate output displacement as complete randomization of the four most prediction-proximal discriminative dimensions.

The single-pass KL measures immediate output displacement, not long-horizon behavioral control. Our interventions showed that perturbations on the narrow prediction-proximal subspaces can be partly repaired by later layers. We tracked this repair by measuring the \emph{D rotation angle} — the angle between the rotated D component and its original, unperturbed orientation, measured at the intervention layer and again near the readout. On SpecA, the D rotation angle is driven above 90° at the intervention layer in every model (100.5–103.5°, 10-model mean 101.7°). The intervention turns D almost fully away from its original direction, but the angle recovers to within about a degree by the penultimate layer in the seven dense non-MoE models (0.19–1.72°), with weaker recovery in the MoE and GPT-OSS architectures (GPT-OSS 20B 1.76°, Mixtral 4.30°, GPT-OSS 120B 4.70°). The same ordering holds on the Diverse prompt set, where the residual deviation is substantially larger in the MoE and GPT-OSS models, up to \textasciitilde{}30° in GPT-OSS 120B. So the repair deficit in those architectures scales with the linguistic heterogeneity of the input. The per-model peak and recovery angles for both prompt sets are reported in the supplementary materials (Table S1).

We also assessed four of the small base models with the single-pass interventions and found the same aggregate hierarchy, with F-mix KL exceeding 0.75 nats in every case. F's single-pass aggregate dominance is therefore not introduced by instruction tuning. It is measurable before alignment training, which is consistent with an architectural origin for the residual-stream organization characterized in §4.

In short, under this single-pass intervention protocol, intervening on direction has a much greater effect than intervening on magnitude. §5.2 tests the directional-versus-magnitude contrast behaviorally, and the persistent interventions in §6 test the behavioral roles of D and S.

\phantomsection\addcontentsline{toc}{subsection}{5.2 Rotation, not magnitude, drives F's variation}
\subsection*{5.2 Rotation, not magnitude, drives F's variation}

Having identified F as the locus of aggregate single-pass sensitivity, we next ask whether the causal sensitivity to direction over magnitude produces measurable behavioral effects. The thin-shell prediction is that directional disturbance should have the greatest effect. If residual states have little radial freedom, then preserving F's norm should not preserve behavior when its direction is destroyed, whereas preserving direction should substantially protect behavior even when magnitude is reduced.

We test the thin-shell hypothesis using the same two matched early-layer interventions with greedy autoregressive generation. F-mix randomly permutes coordinates in the F subspace and applies random sign flips, preserving L2 norm while destroying directional structure. F-attenuate scales F by 0.5, preserving direction while reducing magnitude. These interventions separate rotational structure on the shell from radial displacement off it.

\begin{figure}[tbp]
\centering
\includegraphics[width=\linewidth]{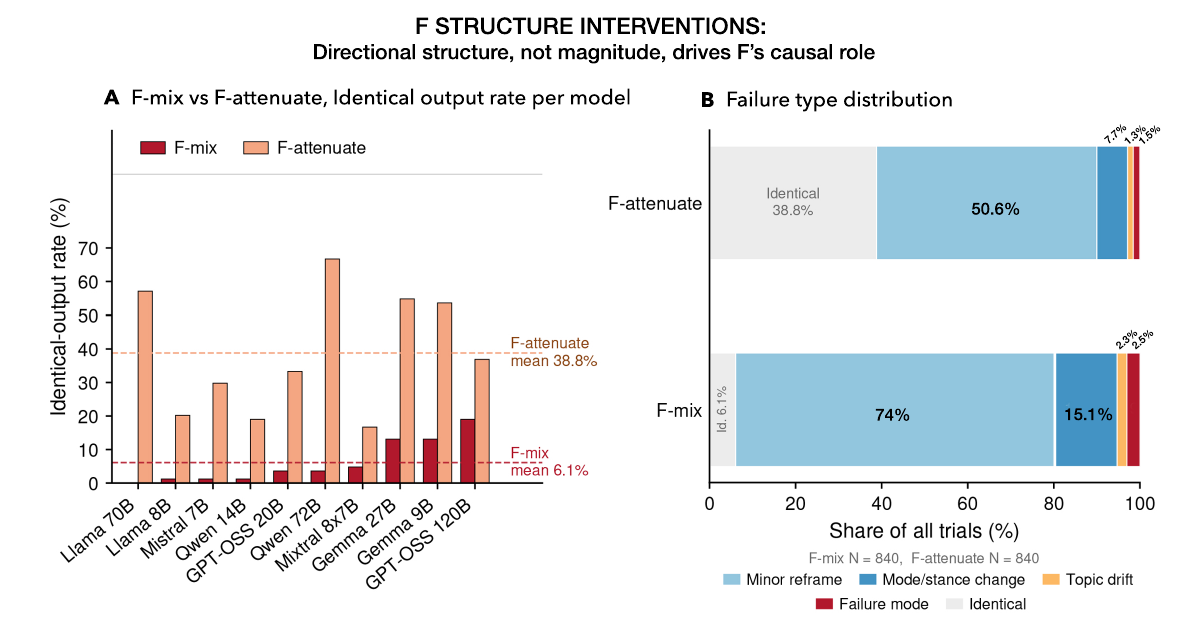}
\caption{\textbf{F structure interventions at early depth: directional structure, not magnitude, drives F's causal role.} F-mix and F-attenuate applied at \textasciitilde{}12\% depth during autoregressive generation on the Diverse prompt set (84 prompts) for 10 instruct models. \textbf{(A)} Rates of identical-output per model (paired bars: F-mix darker, F-attenuate lighter). F-mix: 6.1\% ± 6.6\% mean identical; F-attenuate: 38.8\% ± 18.0\% identical. The hierarchy F-mix ≫ F-attenuate is unanimous across all 10 models. \textbf{(B)} Distribution of failure types for each intervention (Failure types: Failure mode, Topic drift, Mode/stance change, Minor reframe). Both interventions produce predominantly minor reframes and mode/stance changes (F-mix: 74.0\% + 15.1\% = 89.1\%; F-attenuate: 50.6\% + 7.7\% = 58.3\%). Genuine failure modes account for only 2.5\% of F-mix and 1.5\% of F-attenuate outputs. The divergent output remains largely coherent in the majority of outputs. The model selects a different continuation, not an incoherent one.}
\end{figure}

The results of the directional and magnitudinal interventions are clearly different. On Diverse prompts, F-mix is far more disruptive than F-attenuate in every model. At early depth, F-mix leaves only 6.1\% of outputs identical to baseline, with 64.8\% diverging immediately and a mean first-divergence token of 4.10. F-attenuate is substantially better tolerated, with 38.8\% identical outputs and 12.4\% immediate divergence. The paired ordering holds in 10/10 models. The matched single-pass KL comparison shows the same asymmetry: F-mix produces 2.846 nats, while F-attenuate produces 0.047 nats, a 60× gap.

The qualitative pattern is also informative. Both interventions usually change the continuation without destroying coherence. Most F-mix outputs are minor reframes (74.0\%) or mode/stance changes (15.1\%), while genuine failure modes account for only 2.5\% and topic drift for 2.3\% (Table B.6-1). F-attenuate has the same coherent character but acts on far fewer outputs — 50.6\% minor reframes and 7.7\% mode/stance changes, with 1.5\% failure modes and 1.3\% topic drift, alongside its 38.8\% identical rate. The difference between the two interventions is in how often the continuation is rerouted. The two interventions fail at roughly the same low rate (≈2-3\% of changed outputs).

These results suggest that F carries load-bearing directional configuration. Disrupting F's direction reroutes which coherent continuation the model selects, whereas reducing its magnitude largely leaves the selection intact. Representative outputs spanning the mildest reframe through the rare catastrophic outcomes are collected in Appendix D. Full behavioral classification rates, including identical-output rate, immediate divergence, first-divergence token, and failure-type categories, are reported in Appendix B.6, Table B.6-1.

We also performed the same F-mix vs. F-attenuate experiment at \textasciitilde{}70\% depth on the same prompt set for all 10 instruct models. The directional ordering was replicated unanimously across models (F-mix < F-attenuate in identical-output rate in 10/10 models), however the late-depth intervention produces somewhat stronger immediate divergence. This is consistent with later-layer perturbations having less remaining network depth to absorb the interventions, however the qualitative pattern and paired ordering are statistically indistinguishable from the early-depth result. Full per-model values and the depth-comparison table are in Appendix B.5.

The results of these experiments support the expectations of thin-shell geometry: residual-stream variation is more sensitive to direction than to magnitude. F-mix preserves shell radius but destroys directional structure while F-attenuate preserves direction but reduces radius. Both the single-pass KL gap and the autoregressive behavioral gap rank directional disruption far above magnitude reduction. A concurrent line of work in Pythia-family models reports a related L2-matched perturbation double dissociation in which angular perturbations cause substantially more damage to language-modeling loss than magnitude perturbations of equal Euclidean displacement, with the two damage modes flowing through different downstream pathways (Vardhan \& Sai Teja, 2026).

The results of these tests motivated our use of sustained rotational interventions in §6, which assess the behavioral roles of the prediction-proximal directions using the same perturbation logic.

\FloatBarrier
\phantomsection\addcontentsline{toc}{section}{6. Behavioral Stratification Along the Prediction-Proximity Axis}
\section*{6. Behavioral Stratification Along the Prediction-Proximity Axis}

We now use the same perturbation logic to ask how behavior is organized along the prediction-proximity axis itself, focusing on three successive prediction-proximal variance-ordered cuts in the prediction-anchored decomposition: D, S, and F-topK. D and S sample the prediction interface; F-topK sits beyond it in the complement, and is the next nested high-variance cut after P, D, and S have been removed.

In this section we show that persistent rotational intervention separates D from both S and F-topK, which behave similarly. Intervening on D causes immediate divergence from the baseline model response and the responses show frequent task-frame shifts in which the model begins producing a different kind of response from the baseline. Intervening on S produces later divergence and mostly substitutes content within the same broad response frame.

F-topK, despite lying inside F and being orthogonal to S by construction, behaves statistically like S under the same protocol. The behavioral signature changes sharply between D and S and then continues smoothly past S into F-topK, tracking the prediction-proximity ordering rather than the analytical bin labels.

To assess model behavior we used both a divergence-timing metric (the index of the first generated token that differs from baseline) and a four-category taxonomy applied to each output: \textbf{Equivalent} (essentially the same response as baseline), \textbf{Variant} (same broad communicative mode and task framing as baseline, different specific content), \textbf{Frame-shift} (the baseline's task framing is disrupted, producing a different kind of response — meta-commentary, writing advice, grammar analysis, list-format substitution, third-person reframing, or full mode replacement), and \textbf{Failure} (degenerate, fragmentary, or incoherent output). The full classification protocol, calibration, and representative examples are in Appendix A.5.

\phantomsection\addcontentsline{toc}{subsection}{6.1 Persistent rotational scrambling: protocol}
\subsection*{6.1 Persistent rotational scrambling: protocol}

In §5.1, we showed that a one-time D-rotation and S-rotation produced comparable KL values at early depth, even though D and S occupy different positions in the prediction-anchored decomposition. This is expected for narrow subspaces: a one-time perturbation can be partially repaired by later layers before the state reaches readout-relevant depths. When using a rotational intervention on D, for example, the D rotation angle is driven above 90° at the intervention layer in every model (100.5–103.5° on SpecA) but recovers to within about a degree of baseline by the penultimate layer in the dense non-MoE models (§5.1; per-model values in Table S1).

To test D and S behaviorally, we therefore use \emph{persistent rotational scrambling} during autoregressive generation, applied at a later layer to reduce network repair. A fixed random orthogonal rotation, drawn from the Haar measure on the target subspace, is applied to subspace coefficients at every generation step at approximately 70\% network depth. This keeps the perturbation on-shell and preserves rank, norm, and within-subspace energy distribution while repeatedly removing the targeted directional structure. D and S ranks are fixed at 12 and 16 respectively, the values the PR-adaptive construction selects on SpecB in all ten models. We also performed a matched control experiment on a rank-12 random subspace. Protocol details and rank-setting rationale are in Appendix A.4.

\phantomsection\addcontentsline{toc}{subsection}{6.2 The D/S divergence-timing dissociation}
\subsection*{6.2 The D/S divergence-timing dissociation}

\begin{figure}[tbp]
\centering
\includegraphics[width=\linewidth,height=0.76\textheight,keepaspectratio]{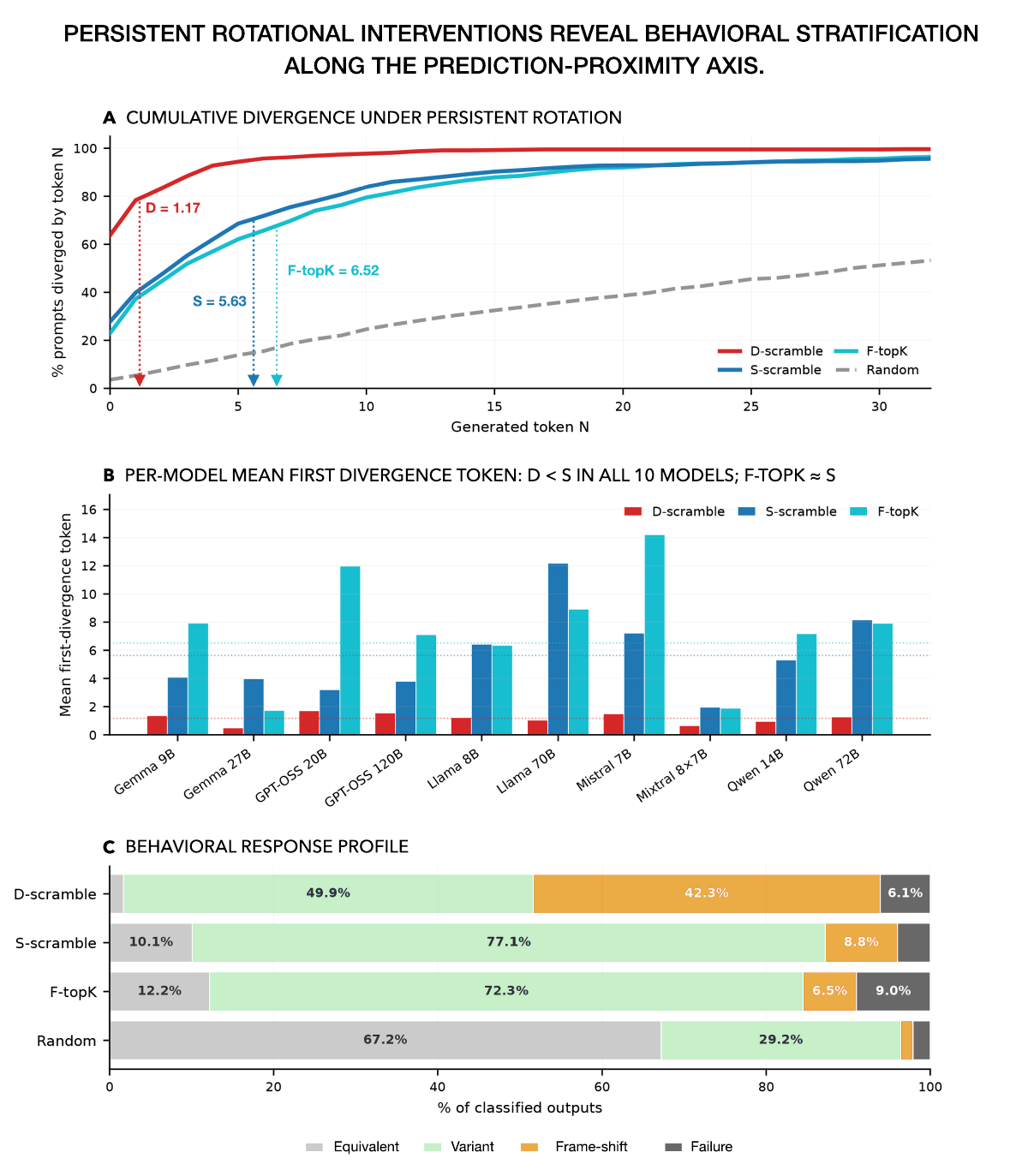}
\caption{\textbf{Persistent rotational interventions reveal behavioral stratification along the prediction-proximity axis.} D-scramble, S-scramble, F-topK rotation, and rank-matched Random rotation applied as persistent interventions at \textasciitilde{}70\% depth during autoregressive generation, 10 instruction-tuned models, SpecB. (A) Cumulative divergence curves: fraction of prompts diverged from baseline by token N. D-scramble rises near-vertically, with most prompts diverging at the first generated token; S-scramble and F-topK rise more gradually and overlap across the generation window; Random produces the weakest and latest divergence. Vertical lines mark cross-model mean first-divergence tokens ($\bar{D}=1.17$, $\bar{S}=5.63$, $\overline{F_{\mathrm{topK}}}=6.52$). (B) Per-model mean first-divergence tokens for D, S, and F-topK. The D < S ordering holds in all 10 models, and F-topK falls in the S-like delayed-divergence regime rather than the D-like immediate-divergence regime. (C) Behavioral category profiles. D-scramble is enriched for Frame-shift, while S-scramble and F-topK are dominated by Variant responses; Random is mostly Equivalent with a Variant tail.}
\end{figure}

The most obvious and surprising finding is a significant temporal gap between D and S that is large and consistent across the model test-bed (Figure 8). Across all ten models, D-scramble changes the output almost immediately — cross-model mean first-divergence token 1.17, with 64.0\% of prompts diverging at the first generated token. In contrast, when the S subspace is scrambled the model's output is generally preserved for the first several tokens before diverging (mean first-divergence token 5.63, 28.2\% immediate divergence). This D-before-S ordering holds in all 10 tested models. Within a model, S diverges on average 4.46 ± 2.99 tokens later than D (95\% CI [2.32, 6.60]; paired Wilcoxon $p = 0.002$, Cohen's $d_z = 1.49$), and the immediate-divergence rate falls by a factor of 2.3. Both interventions exceed the rank-matched random control in every model (see Appendix B.7.4); Cohen's $d$ against Random is above 0.8 for both D-scramble (mean $d = 2.06$) and S-scramble (mean $d = 1.41$). Full per-model divergence metrics are reported in Appendix B.7.1.

The dissociation in behavior between interventions on D versus S is also qualitative as well as temporal. D-scramble produces Frame-shift in 42.3\% of classified outputs, compared with 8.8\% for S-scramble and 1.5\% for Random. S-scramble, by contrast, is dominated by Variant responses: same broad communicative mode and task framing as baseline, but different specific content. Thus D disruption tends to change the model's framing of the task, while S disruption tends to redirect the continuation after the response has already begun in the baseline frame. Side-by-side model output examples of the D-scramble framing shift versus the S-scramble trajectory redirection are in Appendix D.3.

Intervening on the dimensionally narrow prediction interface does not create incoherent output. Failure is not the main effect in either condition. D-scramble Failure is 6.1\% and S-scramble Failure is 4.0\%, with elevated rates concentrated in Gemma 27B and the GPT-OSS family.  The D/S effect is therefore not a contrast between coherence and incoherence; it is a contrast between immediate task-framing disruption and delayed content substitution within a preserved frame. This finding appears to track the construction geometry of variance proximity to the next predicted token. Intervention on D, the higher-variance cut nearer the prediction interface, leads to immediate divergence in the output. Intervention on the lower-variance, more prediction-distal S cut, leads to output divergence several tokens past the immediate next token.

\phantomsection\addcontentsline{toc}{subsection}{6.3 Measuring the behavioral response gradient into F}
\subsection*{6.3 Measuring the behavioral response gradient into F}

Using the same intervention on F-topK, we test whether the delayed-divergence pattern continues further along the prediction-proximity axis into F. For each model, F-topK rank is selected adaptively by participation ratio (range 33–51; mean 45), capturing 23–31\% of F's total energy. We apply the same persistent Haar rotation used for D and S to F-topK coefficients at every generation step, leaving the remaining F-base component unrotated.

F-topK produces behavior statistically indistinguishable from S-scramble at the model level. Across 10 instruction-tuned models on SpecB, F-topK rotation produces a mean first-divergence token of 6.52, compared with 5.63 for S-scramble; immediate-divergence rates are 23.1\% and 28.2\%, respectively. At the model level, paired Wilcoxon tests on per-model mean first-divergence tokens do not distinguish F-topK from S-scramble ($p > 0.05$), and the paired effect size is small. By contrast, F-topK differs strongly from D-scramble: all 10 models show later divergence under F-topK than under D.

The qualitative profile matches the timing result. F-topK produces Frame-shift in 6.5\% of classified outputs, close to S-scramble's 8.8\% and far below D-scramble's 42.3\%. Its dominant category is Variant (72.3\%), again matching S-scramble's content-substitution regime rather than D's task-framing regime. F-topK has a somewhat higher Failure rate than S-scramble (9.0\% vs. 4.0\%), concentrated primarily in GPT-OSS 20B and GPT-OSS 120B, where baseline instability is also elevated; outside those models, F-topK Failure is rare. Full per-model F-topK metrics and the Mixtral-specific Frame-shift elevation are reported in Appendix B.7.2.

This result aligns with our view that the PDSF subspaces are analytical bins rather than functional subspaces and stands in direct contrast to the §4.4 results showing that S and F-topK have inverted geometric structures with respect to discrimination among prompts. This was surprising: F-topK is orthogonal to S by construction, lower-variance, flatter on the manifold-complexity gradient, and anti-discriminative rather than discriminative — yet its behavior under intervention is statistically indistinguishable from S.

\phantomsection\addcontentsline{toc}{subsection}{6.4 Behavioral profiles: D changes framing, S and F-topK redirect trajectory}
\subsection*{6.4 Behavioral profiles: D changes framing, S and F-topK redirect trajectory}

Divergence timing says when an intervention first changes output, but it doesn't by itself say what kind of output the intervention produces. Table 3 shows the four-category distribution under each persistent rotational scrambling condition.

The profile separates D from S and F-topK. D-scramble is enriched for Frame-shift (42.3\%), while S-scramble and F-topK are dominated by Variant outputs and produce Frame-shift at much lower rates (8.8\% and 6.5\%). A χ² test on the 4-category × 4-condition table gives a large association between condition and behavioral category ($p \ll 0.001$). Random is not a pure null — it occasionally produces Variant outputs — but it rarely produces Frame-shift (1.5\%), so the D signature is specifically enriched for task-framing disruption rather than for output difference in general.

\textbf{Table 3.} Behavioral category distribution under persistent rotational intervention, SpecB. D, S, F-topK, and Random are the prediction-proximity comparisons used in this section. F-dynamic is omitted because it targets the temporal-variation axis and is analyzed in §7.

\begin{table}[htbp]
\centering
\footnotesize
\setlength{\tabcolsep}{4pt}
\begin{adjustbox}{max width=\linewidth}
\begin{tabular}{llllll}
\toprule
Condition & N & Equivalent & Variant & Frame-shift & Failure \\
\midrule
D-scramble & 577 & 10 (1.7\%) & 288 (49.9\%) & 244 (42.3\%) & 35 (6.1\%) \\
S-scramble & 680 & 69 (10.1\%) & 524 (77.1\%) & 60 (8.8\%) & 27 (4.0\%) \\
F-topK & 743 & 91 (12.2\%) & 537 (72.3\%) & 48 (6.5\%) & 67 (9.0\%) \\
Random & 757 & 509 (67.2\%) & 221 (29.2\%) & 11 (1.5\%) & 16 (2.1\%) \\
\bottomrule
\end{tabular}
\end{adjustbox}
\end{table}

For compactness, we describe D's profile as \textbf{stance-like}: immediate divergence with frequent task-frame shift. We describe the S/F-topK profile as \textbf{trajectory-like}: delayed divergence with content substitution within a preserved frame. These are descriptive labels for reproducible behavioral signatures, not claims of discrete internal modules. See Appendix B.7.3 for the per-model collapsed-taxonomy breakdown.

\phantomsection\addcontentsline{toc}{subsection}{6.5 Integration: behavior is stratified along the prediction-proximity axis}
\subsection*{6.5 Integration: behavior is stratified along the prediction-proximity axis}

The persistent rotational scrambling results identify a behavioral ordering along the prediction-proximity axis. D, S, and F-topK are successive variance-ordered cuts under the prediction-anchored decomposition, but their behavioral signatures do not change at every cut. Instead, D separates from a combined S/F-topK regime.

\textbf{Table 4.} Persistent rotational scrambling behavioral signatures along the prediction-proximity axis. Divergence-timing values are means ± sample SD computed over the ten per-model values, with runs identical to baseline excluded; the SD therefore reports how far the effect varies across architectures, not how far it varies across prompts within a model (the within-model distributions are strongly right-skewed and zero-inflated, and are summarized by median and IQR in Appendix B.7.1). Frame-shift percentages are pooled rates on the classified denominator, carried over from Table 3, and are not dimensionally comparable to the two SD-bearing columns. F-mix is excluded because it is a single-position F-structure intervention (§5.2), not a persistent rotational scrambling sub-basis intervention. F-dynamic is excluded because it targets temporal variation (§7), not prediction proximity.

\begin{table}[htbp]
\centering
\footnotesize
\setlength{\tabcolsep}{4pt}
\begin{adjustbox}{max width=\linewidth}
\begin{tabular}{llllll}
\toprule
Subspace & Position in PDSF nesting & Mean first-divergence token & Immediate divergence & Frame-shift & Behavioral signature \\
\midrule
D & First cut after P & 1.17 ± 0.39 & 64.0\% ± 19.8 & 42.3\% & stance-like: immediate divergence + task-frame shift \\
S & Next cut after D & 5.63 ± 2.99 & 28.2\% ± 14.3 & 8.8\% & trajectory-like: delayed divergence + content substitution \\
F-topK & Next cut inside F & 6.52 ± 3.03 & 23.1\% ± 12.9 & 6.5\% & trajectory-like: statistically indistinguishable from S at the model level \\
\bottomrule
\end{tabular}
\end{adjustbox}
\end{table}

The main finding is that there is a strong behavioral discontinuity between D and S: D controls the earliest framing of the response, while S influences the subsequent trajectory once that framing is in place. Second, the S-like trajectory regime continues across the analytical S/F boundary into F-topK. PDSF's bin boundaries therefore behave like measurement cuts through a stratified profile rather than partitions between separate functional types.

In short, the same prediction-proximity ordering that structures geometry also structures behavior, but the transition is soft rather than modular suggesting that the bins created by the PDSF decomposition measure an underlying gradient.

The geometric and behavioral stratification we find along the prediction-proximity axis suggest that the prediction direction is a privileged anchor in residual streams, one organizing principle that is universal across the transformer models we studied. It is unlikely to be the only universal organizing principle in the residual stream. In the following section we provide evidence of an entirely separate organizing axis that can be seen by separating the residual stream into static and dynamic partitions inside F.

\FloatBarrier
\phantomsection\addcontentsline{toc}{section}{7. The Static/Dynamic Partition of F}
\section*{7. The Static/Dynamic Partition of F}

Despite its weak readout alignment, F is the highest-dimensional subspace in the PDSF decomposition by a wide margin, it is causally dominant, and it contains organized directional structure that shapes coherent continuation. Here we show that F has a second structural feature, required by autoregressive generation but not predicted by the linear-readout constraint: it separates into a relatively stable component and a dynamic component that must be current at every generation step.

We used passive tracking to decompose F into a temporal mean $\bar F$ and a dynamic envelope $F_t - \bar F$. The static fraction varies widely across architectures, from 30\% to 82\%. The temporal currency of the dynamic component, however, is strictly binary: replacing $F_t$ with $F_{t-1}$ produces universal divergence by token 1 in all 10 tested models, and rotating the directions along which F varies over time produces the same failure mode. Thus F is not only a high-dimensional complement around the prediction interface; it contains an actively maintained autoregressive state, whose extent varies by model.

\phantomsection\addcontentsline{toc}{subsection}{7.1 F contains a stable scaffold and a dynamic envelope}
\subsection*{7.1 F contains a stable scaffold and a dynamic envelope}

We first track F passively during greedy generation, reading the F component at the intervention layer without modifying it. Across the generation window, F decomposes into a temporal mean $\bar{F} = (1/T) \sum_t F_t$ and a centered dynamic envelope $F_t - \bar{F}$. The temporal mean behaves as a relatively static scaffold, while the envelope contains the step-specific update.

We find that the static energy fraction varies by more than 2× across architectures. This variability shows that models allocate different amounts of F to stable versus step-specific structure. It does not predict tolerance to stale F: as §7.2 shows, all tested models fail when F is one step out of date.

\begin{figure}[tbp]
\centering
\includegraphics[width=\linewidth,height=0.76\textheight,keepaspectratio]{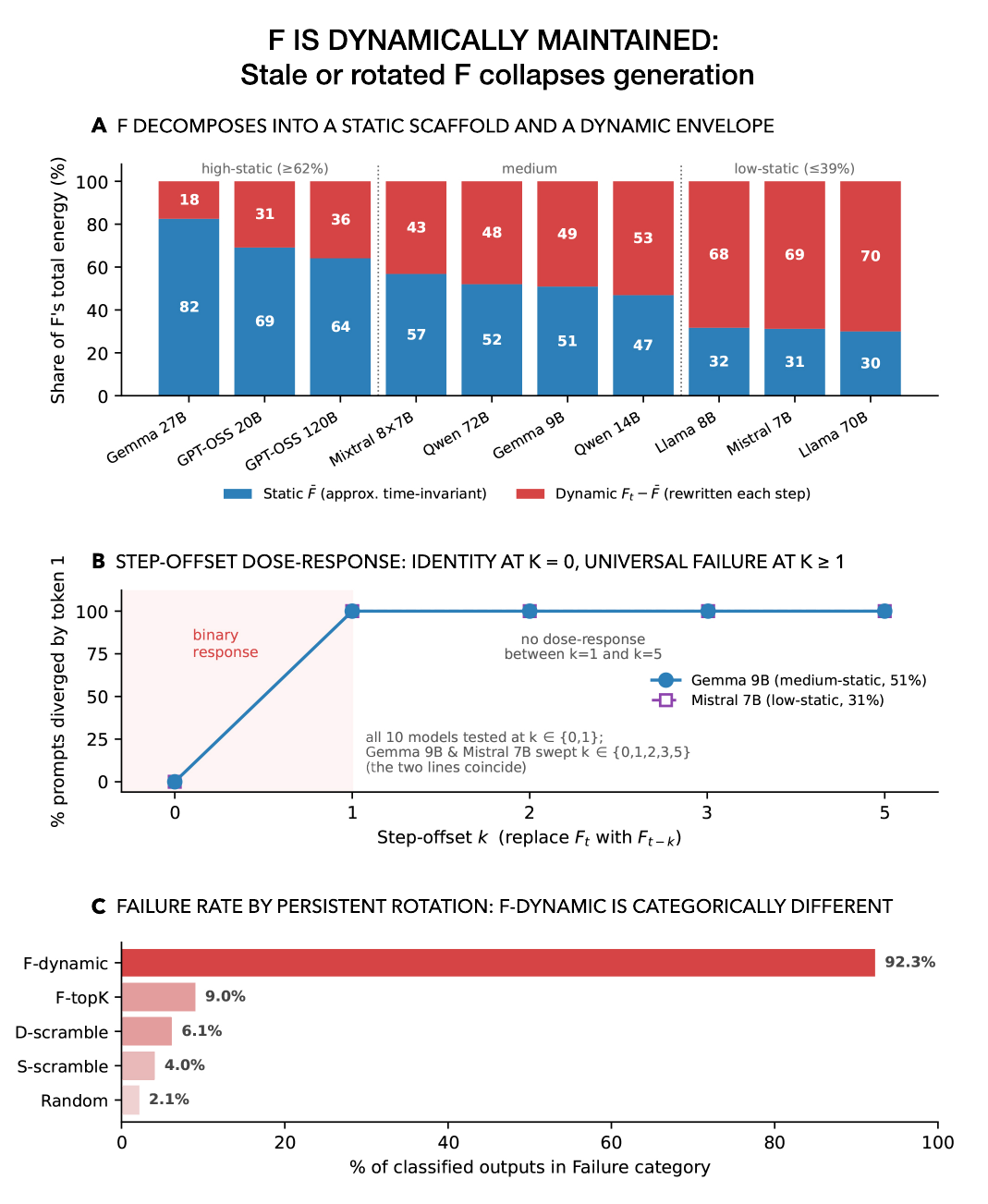}
\caption{\textbf{F decomposes into a static component and a dynamic envelope; stale F and rotated F-dynamic both prevent coherent generation.} (A) Static energy fraction $|\bar{F}|^2 / \sum_t |F_t|^2$ across 10 tested models, ordered from highest to lowest static fraction and partitioned into three bands: high-static (Gemma 27B, GPT-OSS 20B, GPT-OSS 120B at static fraction 0.641), medium-static (Mixtral 8×7B, Qwen 72B, Gemma 9B, Qwen 14B), and low-static (Llama 8B, Mistral 7B, Llama 70B). (B) Step-offset dose-response for Gemma 9B (medium-static) and Mistral 7B (low-static), the two models with the broader $k \in \{0, 1, 2, 3, 5\}$ sweep: at $k = 0$, output matches baseline, confirming the intervention infrastructure is lossless; at $k \geq 1$, every prompt diverges by token 1, with no further gradient between $k = 1$ and $k = 5$. The remaining 8 models were only tested for k=0 and k=1 and all show the same 0\% → 100\% failure by-token-1. (C) Failure rate by persistent rotational scrambling condition: D, S, F-topK, and Random all sit in the 2–9\% Failure range under the same protocol, but the failure rate for F-dynamic jumps to 92.3\%.}
\end{figure}

\phantomsection\addcontentsline{toc}{subsection}{7.2 F must be current at every generation step}
\subsection*{7.2 F must be current at every generation step}

We tested temporal currency by injecting stale F. At generation step $t$, the hook replaces the natural $F_t$ at the intervention layer (\textasciitilde{}70\% depth) with $F_{t-k}$. The $k = 0$ condition is an identity control and reproduces baseline exactly. At $k \geq 1$, the model receives stale F.

The result is universal divergence by token 1 across the tested models. At $k = 0$, output is identical to baseline for all prompts and all generation steps, confirming the intervention infrastructure is lossless. At $k = 1$, every prompt across all 10 tested models diverge no later than token 1, with a cross-model mean first-divergence token of 0.81 (range 0.67–0.95) and prefix overlap with baseline of approximately 1\%. Whether divergence occurs at token 0 or token 1 is model-graded — token-0 divergence ranges from 5\% (Qwen 72B) to 33\% (Gemma 9B), with a cross-model mean of 19\% — but the by-token-1 endpoint is universal.

\textbf{Table 5.} F-dynamic step-offset summary, 10 tested models, Diverse prompt set (20 prompts in 9 of 10 models, 84 in Gemma 9B; 32 generation steps).

\begin{table}[htbp]
\centering
\footnotesize
\setlength{\tabcolsep}{4pt}
\begin{adjustbox}{max width=\linewidth}
\begin{tabular}{lll}
\toprule
Metric & $k = 0$ identity & $k = 1$ one-step stale \\
\midrule
Divergence by token 1 & 0\% (all models) & 100\% (all models, all prompts) \\
Divergence at token 0 & 0\% & 5–33\% per model; 19\% cross-model mean \\
Mean first-divergence token & 32 (censored; no divergence within the 32-step window) & 0.81 (range 0.67–0.95) \\
Prefix overlap with baseline & 100\% & \textasciitilde{}1\% \\
\bottomrule
\end{tabular}
\end{adjustbox}
\end{table}

There is no dose-response to the effect. We performed the test through token 5 on two models, Gemma 9B and Mistral 7B. On both models, $k = 1$, $k = 2$, $k = 3$, and $k = 5$ all produce the same failure pattern. One-step stale F is already fully wrong for the autoregressive computation; making it older does not measurably worsen the result.

Strictly, the intervention replaces F as a whole rather than selectively replacing $F_t - \bar F$. The attribution to the dynamic component rests on the passive-tracking result: $\bar F$ changes little across the generation window, so $F_{t-k}$ preserves nearly the same static component while replacing the step-specific envelope. A selective freeze of $\bar F$ or $F_t - \bar F$ would test this directly; such a freeze is not part of the present design.

\phantomsection\addcontentsline{toc}{subsection}{7.3 Temporal-variation directions are spatially load-bearing}
\subsection*{7.3 Temporal-variation directions are spatially load-bearing}

The §7.2 stale-F intervention shows that F's content must be current. It does not say whether the dynamic envelope's information lives in specific directions or could be carried by any high-variance directions of comparable rank. To test this, we defined an F-dynamic basis by PCA of F's token-to-token trajectory during generation, with the rotation rank fixed at 64 across nine models (rank 10 for Mistral 7B). We then applied the same persistent rotational scrambling used in §6 to those directions only. Because the participation ratio of F's temporal trajectory ranges 12.9–29.7 across models, the experiment captures essentially all of F-dynamic's effective variance for every model rather than a model-dependent fraction of it. Using norm-preserving rotation, the dynamic envelope's energy and rank survive but its directional identity is randomized at every generation step. If the dynamic information is direction-specific and spatially load-bearing, generation should fail; if not, generation should proceed.

The results were clear. Rotating this temporal-variation based subspace produces categorical failure: 92.3\% of classified outputs are Failure on the screened denominator (Table B.7-1. 44 GPT-OSS cells were excluded as baseline-degenerate per the screen in Appendix A.5). The Failure rate is universally elevated but model-graded — six models failed 100\% of the time, and the remaining four (Llama 70B, Mistral 7B, Qwen 72B, Llama 8B) fail between 58.8\% and 97.5\% (Table B.7-1b). However even the models with lower failure rates produce output that is qualitatively unlike D, S, and F-topK rotation, where Failure remains 4.0–9.0\% and outputs stay coherent but redirected.

The temporal replacement and spatial rotation experiments converge to show that the dynamic envelope is both temporally inelastic and spatially load-bearing. Stale-F injection shows that F must be current. F-dynamic rotation shows that the directions along which F changes over time cannot be randomized.

\phantomsection\addcontentsline{toc}{subsection}{7.4 Integration: F is prediction-distal and dynamically maintained}
\subsection*{7.4 Integration: F is prediction-distal and dynamically maintained}

The results of the static/dynamic partition are distinct from the prediction-proximity stratification findings. Prediction proximity orders directions by relation to the readout and yields the D/S/F-topK behavioral profile in §6. The static/dynamic partition instead separates directions by whether they are stable across the generation window or rewritten at each token. Our experiments do not test whether these properties can cross-cut. They do demonstrate that a direction can be prediction-distal but dynamically essential, as F-dynamic shows, meaning that the prediction distal component of the residual state contains meaningful structural features.

Thus distance from the readout does not equate to behavioral unimportance. Along the prediction-proximity axis, F's organized directional structure helps route coherent continuation; along the static/dynamic partition, its dynamic envelope carries the autoregressive update required for generation to proceed at all. Together, these results characterize F as a high-dimensional, prediction-distal and dynamically maintained region of the residual stream.

\FloatBarrier
\phantomsection\addcontentsline{toc}{section}{8. Discussion}
\section*{8. Discussion}

\phantomsection\addcontentsline{toc}{subsection}{8.1 A fixed prediction interface embedded in a scaling substrate}
\subsection*{8.1 A fixed prediction interface embedded in a scaling substrate}

Our central finding is that prediction-proximal dimensionality does not scale with model width. "Fixed" here refers to the width of that structure, not to a fixed set of directions: the interface is the prediction-proximal region of the residual stream's variance distribution, and the P, D and S bins are how we measure it rather than what it is. Across architectures and training regimes, transformer models maintain a narrow interface through which residual-stream computation becomes legible to the unembedding. Base-model measurements show that this organization precedes instruction tuning. Across 18 models spanning 6 architecture families and a 17× parameter range, the effective rank of the interface's discriminative core ($D_{PR}$ on SpecA prompts) sits between 2.89–7.21 across the 8 base models and 3.60–5.00 across the 10 instruction-tuned models, with no correlation with model size. Instruction tuning sharpens how the interface is used, but it does not expand it. Scaling changes the substrate around the interface, not the width of the interface itself.

The prediction interface appears to be an architectural bottleneck imposed by linear readout under norm concentration. The remaining majority of the residual-stream, the complement, is the substrate that grows with model scale, computing and maintaining the state that is eventually projected through the prediction interface into logits.

\phantomsection\addcontentsline{toc}{subsection}{8.2 Geometric organization follows prediction proximity}
\subsection*{8.2 Geometric organization follows prediction proximity}

Our results show that two distinct geometric properties, direction and manifold complexity, organize along the prediction-proximity axis. First, the prediction direction is nearly orthogonal to the principal variance axes, 84–87° across the testbed (§4.2). High-variance directions in the residual stream are not aligned to the readout but displaced away from it. Second, manifold complexity stratifies along the same axis (§4.3). Prediction-proximal regions are more folded and locally structured than prediction-distal regions, and the gradient is tied specifically to the prediction axis. Anchoring manifold complexity measurements to the prediction direction recovers a steeper gradient than matched-rank variance-ordered controls in all thirty model × prompt-set cells, by a factor of 1.1–1.6 on constrained prompt sets and up to 8.1 on linguistically heterogeneous ones.

These two properties arise for different reasons. We propose that the readout/variance separation follows from models using geometry to maximize signal-to-noise ratio in the readout. The high-dimensional residual background must be kept apart from the prediction-relevant directions to assure optimal Argmax distributions. Manifold folding likely follows from packing pressure. As a result of a narrow prediction interface, fine semantic and contextual distinctions must be expressed in a more highly folded dimensional space. The complement has the opposite geometry: many low-variance dimensions, weak per-direction readout alignment, and flat population-level structure. Together these results indicate that prediction proximity is not merely a convenient coordinate choice — it is an organizing axis of residual-stream geometry.

\phantomsection\addcontentsline{toc}{subsection}{8.3 F anti-discrimination and configurational compensation}
\subsection*{8.3 F anti-discrimination and configurational compensation}

The use of variance among structured prompt sets to define the prediction axis gives an a priori expectation that group discrimination will fade with prediction distance. D should carry the strongest group structure, S less, and F little or none. Surprisingly, we found that F reverses the similarity structure. F-base anti-discriminates in 30 of 30 measured cells (10 models × 3 prompt sets); F-topK and the full F anti-discriminate in 27 of 30; and F-dynamic, the temporal variant introduced in §7, anti-discriminates in 10 of 10 measured SpecB cells. In other words, same-group prompts are systematically farther apart in F-space than between-group prompts — universally so in F-base and F-dynamic. The anti-discrimination is present in base models and instruction tuning amplifies it.

This pattern demonstrates that F carries prompt-specific configuration rather than group-level clustering, which appears to be specialized in the narrow prediction proximal directions. D and S capture leading cross-prompt variance after sequential removal from the prediction direction. On group-structured prompt sets, that variance aligns with group membership. What remains in F varies against the group axis that is expressed near the prediction interface. This suggests that same-group prompts that converge toward similar output behavior through different D and S states may require different F configurations to integrate those states into compatible readout outcomes.

\phantomsection\addcontentsline{toc}{subsection}{8.4 Direction, not magnitude, is the operative degree of freedom}
\subsection*{8.4 Direction, not magnitude, is the operative degree of freedom}

Thin-shell theory predicts that residual-stream variation should be carried primarily by direction rather than radial magnitude (§2). LayerNorm and RMSNorm constrain residual states to a narrow norm band, leaving the model with a largely directional representational budget. Under this constraint, changing where a component points should matter more than changing how large it is, provided the magnitude change does not destroy the state outright.

Section 5 tests this directly. Rotational scrambling of F is far more disruptive than attenuation, both in single-pass KL (§5.1) and in autoregressive continuation (§5.2).

The results of §5 also establish that F's causal role is not merely a consequence of its large aggregate norm or dimensionality. Looked at through the lens of geometric variation, F is flatter and lower variance than the prediction interface, suggesting that information is distributed across many low-variance directions. Section 5 shows that disrupting this organization redirects behavior immediately. The complement is therefore not a passive reservoir of residual capacity. It is a directional substrate whose orientation helps determine which coherent continuation the model selects.

The direction-over-magnitude framing was also recently demonstrated by Vardhan \& Sai Teja (2026), who held each hidden state's L2 norm fixed while varying its direction and found that model behavior is far more sensitive to directional change than to a comparable change in magnitude. Our results agree with theirs and support the use of rotational interventions as the primary behavioral probe across the PDSF decomposition.

\phantomsection\addcontentsline{toc}{subsection}{8.5 Function follows geometry along the prediction-proximity axis}
\subsection*{8.5 Function follows geometry along the prediction-proximity axis}

In §6 we used a persistent rotational intervention to show that the geometric organization along the prediction-proximity axis also corresponds to distinct behavioral roles. Among the prediction-proximal bins, D and S produce different behavioral signatures under disruption. The behavioral difference between D and S is not a difference between coherent and incoherent output. Interventions on both subspaces preserve coherence in the majority of samples, but intervening on D causes rapid divergence, while intervening on S causes later divergence. In addition, intervening on D results in a high number of shifts in task-framing, while intervening on S more often preserves the initial communicative mode while redirecting the continuation.

F-topK — the next variance cut along the prediction-proximity axis — extends this pattern into the prediction-distal complement. Although F-topK is orthogonal to S by construction, lower-variance, flatter on the manifold-complexity gradient, and usually anti-discriminative rather than discriminative, persistent rotation produces the same delayed, trajectory-like behavioral regime as S (paired Wilcoxon, $p > 0.05$ at the model level). Large geometric differences do not necessarily imply different behavioral signatures under intervention.

\phantomsection\addcontentsline{toc}{subsection}{8.6 The prediction direction and privileged geometry}
\subsection*{8.6 The prediction direction and privileged geometry}

Our geometric and behavioral findings suggest that the model's own prediction defines a content-defined privileged reference frame for residual-stream geometry. When residual states are measured relative to this prompt-specific direction, cross-prompt variation exhibits a consistent stratification across depth. This differs in kind from the coordinate-axis privileged basis described by Elhage et al. (2023), a fixed set of coordinate directions whose privileging is induced by training and attributable in part to optimizer dynamics. The prediction direction, in contrast, is selected by content and re-anchored at every generation step. We suggest that our findings are best described as a privileged geometric structure organized relative to the moving anchor of the prediction direction.

In principle, every direction in the residual stream should be equivalent with respect to rotation (van Nierop, 2024). In other words, rotating the residual-stream coordinates and rotating the weight matrices to match should produce an identical model function. In practice, training breaks this symmetry and creates a privileged basis — specific coordinate directions whose statistics differ systematically from the rest of the geometry (Elhage et al., 2023; Silverstein et al., 2026). However, prior work has not shown what kind of direction those privileged bases select. Our work identifies a different sort of statistical privilege.

We demonstrated geometric privilege by showing that prediction-alignment and variance-ranking are not geometrically equivalent. The prediction direction is displaced 84–87° from the main variance axes across our testbed (§4.2). Anchoring the residual-stream decomposition to the prediction direction recovers a consistently steeper manifold-complexity gradient than a matched-rank variance-ordered control, in every model and prompt set tested, and the margin grows as prompts become more linguistically heterogeneous (§4.3). So measuring along the prediction direction is statistically different from measuring along the principal axes of variation, and these differences manifest in group discrimination and behavioral disruption, which order monotonically along the axis.

The prediction-proximal region also has the notable feature that its effective dimensionality remains approximately constant across models of greatly varying width (§4.1). In §2 we proposed that this fixed, low-dimensional interface can be explained by signal-to-noise constraints arising from the linear readout of the residual stream through the unembedding matrix. That architectural constraint does not predict the other statistical differences we measure relative to the prediction direction: it would be equally well satisfied by a fixed low-dimensional readout subspace containing no privileged direction. In short, the narrowness of the prediction interface can be explained by transformer architecture, but the additional statistical structure we measure relative to the prediction direction is not predicted by that constraint.

In retrospect, the prediction direction appears to be a natural place to look for privileged geometric structure, but the mechanism is likely more related to thin-shell geometry than to alignment with any particular chosen token. Section 2 and Appendix B.1 show that under norm concentration the residual state has little radial freedom, so logit orderings are set by the state's angular relationship to the unembedding rows rather than by accumulation of radial magnitude. A token cannot be made to win by adding magnitude along its readout direction, because there is little magnitude to add. A token wins by the state rotating toward that direction and away from its competitors. This makes prediction an angular and zero-sum condition, which is consistent with what we measure. The prediction direction carries little of the residual stream's cross-prompt variance and sits 84–87° from its principal axes (§4.2), and in §5 we confirmed that the operative degree of freedom in the complement is direction rather than magnitude. In other words, the prediction direction is a readout coordinate rather than a carrier: it records which token the pass resolves to, not the computation that resolved it.

This understanding of logit selection bears on how the PDSF decomposition should be read. If our results were reporting on information carried by the anchoring direction itself, finding structure ordered by proximity to it would be circular. The only property imposed by the construction, however, is the variance ordering D > S > F. The other measures we report — manifold complexity, the discrimination reversal, and behavioral divergence timing — are not products of the construction, and the anchoring direction does not itself carry the structure they reveal. In the behavioral tests, D is orthogonal to P by construction, so persistent scrambling of D leaves the prediction direction's own component untouched at the intervention layer; the model's prediction nonetheless changes on the immediately following token in 64\% of prompts (§6.2). What our work establishes is that anchoring to the prediction direction exposes privileged structure that orders relative to it. It does not establish that the structure resides in the prediction direction, and it does not establish whether anchoring instead to a runner-up token's unembedding direction, or to a direction drawn at random, would recover the same ordering (Appendix F).

The variance ordering is the one result that follows directly from the method. What was not obvious is that anchoring to this content-defined direction would also recover a scale-invariant interface, a stronger-than-PCA manifold gradient, a sign inversion in F, and a behavioral ordering under intervention.

Concurrent with this work, Gurnee et al. (2026) identified a "workspace-like" structure in the residual stream, which they describe as a narrow privileged region inside a much larger complement. Like us they derive structure from statistics across a large corpus of prompts, but rather than relying on the logit lens they developed a Jacobian-based lens as a refinement of the logit lens that corrects for representational change across layers. Using this instrument they located the region by causal influence on output rather than by geometry. More importantly for comparison to the work we present, the workspace-like behavior Gurnee et al. report is confined to intermediate layers, while our decomposition is anchored to a final-layer direction, making the two studies difficult to compare. Whether the prediction interface we find overlaps with their "workspace-like" structure is an interesting open question.

It is possible and perhaps likely that transformer models have multiple privileged directions. Our own results already suggest as much, since §7 identifies a second organizing axis inside the complement that stratifies F by temporal variation rather than by prediction proximity.

\phantomsection\addcontentsline{toc}{subsection}{8.7 The static/dynamic partition of the complement}
\subsection*{8.7 The static/dynamic partition of the complement}

The final question asked by this study was how the majority prediction-distal component of the residual stream varies during autoregressive generation. We find that F is dynamically maintained and separates into a relatively static component, approximated by the temporal mean $\bar F$, and a dynamic component, $F_t - \bar F$, that changes from token to token. We also find that the proportional sizes of the static vs dynamic regions varies substantially across architectures, from 30\% to 82\% (§7.1).

It is unsurprising that residual states change during autoregressive generation. Our step-offset test confirmed that one-step-stale F destroys output (§7.2). The stronger result is that the complement separates into components that change at different rates, and the fast-changing component must remain current.

When an intervention of persistent rotation is applied to that fast-changing component, F-dynamic, model output fails in 92.3\% of classified cells (§7.3), a qualitatively different failure profile from D, S, or F-topK rotation. Rotating D, S, or F-topK redirects coherent generation, but the same intervention on F-dynamic prevents coherent continuation.

These two findings about the dynamics of F show that changes over time in the prediction-distal component, F, are both temporally inelastic and spatially load-bearing. This finding appears to be consistent with the Bayesian belief-state interpretation proposed by Shai et al. (2024) and Agarwal et al. (2025). The static component of F can be seen as accumulated contextual prior, while the dynamic component is perhaps its token-by-token update.

Taken together, the prediction-proximity and static/dynamic results show that F is organized along at least two structural features. Along the prediction-proximity axis, the directional structure of the residual-stream participates in coherent trajectory selection. Along the temporal axis, the dynamic envelope carries the autoregressive update required for generation to proceed. Neither feature alone accounts for the role of F, but both are needed to understand why a prediction-distal, weakly readout-legible complement can be causally central.

\phantomsection\addcontentsline{toc}{subsection}{8.8 Implications for scaling}
\subsection*{8.8 Implications for scaling}

Our findings have implications for understanding what does and does not change in residual-stream geometry as models grow in size and capacity. We show that the prediction interface scales with the demands of the prompt distribution but not with model size. As models grow, the additional dimensionality goes to the prediction-distal complement rather than to the prediction interface, whose effective rank remains narrow and approximately stable. This is consistent with the hypothesis of SNR pressure introduced in §2. The unembedding can cleanly resolve the predicted token only if prediction-relevant variance is isolated from the high-dimensional computational background, so the readout-facing bottleneck stays narrow even as the surrounding complement expands.

These results arise from the current architectural standard, in which the readout from the residual stream through the unembedding matrix is a single linear projection. If prediction were instead mediated by a nonlinear, recurrent, or multi-step readout, the dimensionality of prediction-relevant structure might be less constrained. The present results isolate the linear readout as one source of the observed constraint; they do not rule out additional scaling routes outside this geometric mechanism.

\phantomsection\addcontentsline{toc}{subsection}{8.9 Implications for evaluation blindness and interpretability}
\subsection*{8.9 Implications for evaluation blindness and interpretability}

Transformer models generate next-token distributions by projecting a high-dimensional internal computation onto a low-dimensional readout. Our results show that the prediction interface, which concentrates readout-aligned variance, does not scale with model width (§§4.1, 8.1) while the bulk of internal computation sits in a causally load-bearing, orthogonal complement that is only weakly legible at the output (§§5–7). Output-based evaluation therefore observes only a structurally narrow slice of what the model is doing. This finding has implications for both evaluation methodology and interpretability work.

Next-token metrics, reward-model scores, preference judgments, and benchmark outcomes all observe behavior through the prediction interface. Our results suggest that these metrics may under-measure changes in the substrate that prepares or configures those outputs due to geometric constraints. As models scale in size, the fraction of internal change that is visible to output metrics must shrink relative to the model's representational capacity.

Chain-of-thought may be a partial exception to this finding, but it is unlikely to be a full escape from the geometric constraint because CoT output must still pass through the same constraints imposed by the unembedding. Related concerns about evaluation observability have recently been raised at the reasoning-step level (Korbak et al., 2025) where chain-of-thought monitorability offers a complementary window into model computation but is fragile under outcome-based training pressure. CoT exposes more of the model's reasoning trajectory than a single output does, because each generated token is an additional observation of the model's internal state. But each of those observations is still filtered through the same narrow readout–per token, the complement's contribution to the output remains geometrically constrained.

Our results open several possible directions for follow-up work, including using the prediction axis framing with existing interpretability tools — sparse autoencoders, probes, activation patching. While we believe the PDSF decomposition provided a useful starting point to understand the influence of the prediction axis in the residual stream, we also believe that this framing can be utilized without needing to decompose the residual stream into the same specific subspaces.

Exactly what the prediction interface reads out to in the distributional output remains an open question. Our interventions show that D and S are causally readout-relevant, but not whether they align with the competing tokens of a given prediction or shape the output through other directions. This would be a natural target for the feature-level analysis above. If sparse autoencoders are trained separately on D, S, F-topK, and F-base, the resulting feature dictionaries should look qualitatively different along the prediction-proximity axis — fewer high-variance features in D, more low-variance features in F. That comparison is hard to make without first dividing the residual stream, because the two populations blend together when measured against the undivided whole.

The D/S behavioral dissociation is particularly valuable for interpretability work that targets task framing or content trajectory, because it suggests which subspace each behavior is likely to live in (Conmy et al., 2023; Olsson et al., 2022).

\phantomsection\addcontentsline{toc}{subsection}{8.10 Summary}
\subsection*{8.10 Summary}

Across 18 models, 6 architecture families, and a 17× parameter range, we demonstrated that transformer residual streams stratify geometrically and behaviorally along a prediction-proximity axis anchored to a fixed prediction interface imposed by linear readout.

The prediction interface we find is dimensionally narrow and approximately scale-invariant, while the residual-stream complement accumulates with model width. Manifold complexity decreases monotonically from the interface into the complement, and the complement systematically anti-discriminates among prompt groups. Same-group prompts are farther apart in F-space than between-group prompts, an inversion that holds in 30 of 30 measured cells. The prediction direction does not align with the principal variance axes of the residual stream. Instead, it sits nearly orthogonal to the principal directions across the residual stream (84–87° across the testbed). High-variance directions in the residual stream are likewise displaced away from the readout. As a result, variance-based analyses recover the geometric organization only in part, and the shortfall widens as prompts become less alike: a rank-matched variance-ordered control recovers most of the manifold-complexity gradient on constrained factorial prompts but less than half of it on linguistically heterogeneous ones. Anchoring measurements to the prediction direction is what makes the organization fully visible.

We find that behavior also stratifies along the same axis. Interventions that persistently rotate the D subspace produce immediate divergence and task-frame shifts. The same interventions on S delay divergence and create content substitution within a preserved frame. Applying those interventions to the next variance subspace inside the complement, F-topK (orthogonal to S by construction), we find the same delayed-divergence regime as S.

Our work shows that despite the concentration of residual-stream variance into the prediction interface, the complement, which is by far the largest part of the residual stream, remains causally load-bearing despite weak readout alignment. Randomizing the directional structure of the complement while preserving the norm forces near-universal divergence in output, while reducing its magnitude is comparatively tolerated. The complement also partitions temporally into a stable scaffold and a dynamic envelope that is required for generation to proceed. A one-token-stale F collapses coherent output in every tested model, and persistent rotation of the temporal-variation directions produces failure in 92.3\% of classified cells.

We believe these findings have implications for scaling and evaluation. Scaling expands the complement rather than the prediction interface. Additional model width enlarges the substrate that prepares and configures predictions rather than widening the channel through which predictions are read out. Because the complement is weakly legible at the output but load-bearing inside the network, output-based evaluation observes only a structurally narrow slice of what the model is doing, and that slice does not widen as models scale. The picture of a narrow privileged region inside a much larger complement also appears in recent work identifying a workspace-like structure that supports explicit reasoning and reportability (Gurnee et al., 2026), though that structure is located by causal influence rather than by geometry and is confined to intermediate layers, so the two are not directly comparable (§8.6).

Our work shows that the residual stream is best understood not as a diffuse high-dimensional activation but as a stratified geometry — a fixed prediction interface embedded in a dynamically maintained substrate whose internal organization extends along multiple distinct structural features.

Detailed scope and limitations of these claims are documented in Appendix F.

{\footnotesize\noindent\textbf{Note (GPT-OSS).} Both GPT-OSS models show consistent architectural anomalies relative to the other families such as absent early-layer reconstruction capacity while preserving the structural organization reported here; see Appendix C.4 for the full comparison.\par}

{\footnotesize\noindent\textbf{Note (specb96).} SpecB geometric measurements — the prediction-proximal dimensionalities of §4.1 and Table 1, the prediction-direction angles of §4.2, and the variance-ranked control ratio of §4.3 — are computed on the full 96-prompt continuation set rather than the 80-prompt narrative subset used for the SpecB behavioral results of §6. The difference between the two sets is the prompt count only, and each quantity is reported on the set it was run on.\par}

{\footnotesize\noindent\textbf{Note (MoE).} Mixtral (MoE) shows systematically weaker geometric reorganization under instruction tuning across multiple metrics: the only instruct model with a negative SpecB gap (−0.1°), and the largest residual Factor A sensitivity (|d| = 0.063). Simultaneously, it shows the most extreme spectral expansion on open-ended prompts (SpecB $k_D$ = 59 vs. next highest 38). This pattern is consistent with expert routing providing a parallel discrimination channel that does not manifest in the shared residual-stream geometry captured by PDSF. The core results, causal hierarchy, manifold gradient, dimensionality constraint, all hold for MoE architectures; what is attenuated is the sharpening magnitude under instruction tuning, not the structural organizing principle.\par}

\FloatBarrier
\phantomsection\addcontentsline{toc}{section}{Acknowledgements}
\section*{Acknowledgements}

I thank Diego Garcia-Olano for his thoughtful feedback on this manuscript. His careful reading and comments pushed me to substantially rewrite the paper for clarity, and the presentation is considerably stronger for it. I am also grateful for his endorsement, without which this preprint could not have been posted to arXiv.

\FloatBarrier
\phantomsection\addcontentsline{toc}{section}{References}
\section*{References}

Agarwal, N., Dalal, S. R., \& Misra, V. (2025). The Bayesian geometry of transformer attention. \emph{arXiv:2512.22471}.

Ansuini, A., Laio, A., Macke, J. H., \& Zoccolan, D. (2019). Intrinsic dimension of data representations in deep neural networks. \emph{NeurIPS 2019}.

Ba, J. L., Kiros, J. R., \& Hinton, G. E. (2016). Layer normalization. \emph{arXiv:1607.06450}.

Belrose, N., Furman, H., Smith, J., et al. (2023). Eliciting latent predictions from transformers with the tuned lens. \emph{arXiv:2303.08112}.

Bricken, T., Templeton, A., Batson, J., et al. (2023). Towards monosemanticity: Decomposing language models with dictionary learning. \emph{Transformer Circuits Thread}.

Cheng, E., Kervadec, C., \& Baroni, M. (2023). Bridging information-theoretic and geometric compression in language models. \emph{EMNLP 2023}. arXiv:2310.13620.

Cheng, E., Doimo, D., Kervadec, C., Macocco, I., Yu, J., Laio, A., \& Baroni, M. (2025). Emergence of a high-dimensional abstraction phase in language transformers. \emph{ICLR 2025}. arXiv:2405.15471.

Conmy, A., Mavor-Parker, A. N., Lynch, A., Heimersheim, S., \& Garriga-Alonso, A. (2023). Towards automated circuit discovery for mechanistic interpretability. \emph{NeurIPS 2023}. arXiv:2304.14997.

Elhage, N., Nanda, N., Olsson, C., et al. (2021). A mathematical framework for transformer circuits. \emph{Transformer Circuits Thread}.

Elhage, N., Hume, T., Olsson, C., et al. (2022). Toy Models of Superposition. \emph{Transformer Circuits Thread}.

Elhage, N., Lasenby, R., \& Olah, C. (2023). Privileged bases in the transformer residual stream. \emph{Transformer Circuits Thread}.

Engels, J., Liao, I., Michaud, E. J., Gurnee, W., \& Tegmark, M. (2024). Not all language model features are one-dimensionally linear. \emph{ICLR 2025}. arXiv:2405.14860.

Ethayarajh, K. (2019). How contextual are contextualized word representations? Comparing the geometry of BERT, ELMo, and GPT-2 embeddings. \emph{EMNLP 2019}. arXiv:1909.00512.

Fernando, J. \& Guitchounts, G. (2025). Transformer dynamics: A neuroscientific approach to interpretability. \emph{arXiv:2502.12131}.

Franke, J. K. H., Spiegelhalter, U., Nezhurina, M., Jitsev, J., Hutter, F., \& Hefenbrock, M. (2025). Learning in compact spaces with approximately normalized transformer. \emph{NeurIPS 2025}. arXiv:2505.22014.

Gurnee, W., Sofroniew, N., Pearce, A., Piotrowski, M., Kauvar, I., Chen, R., Soligo, A., Bogdan, P., Ong, E., Wang, R., Thompson, T. B., Abrahams, D., Kantamneni, S., Ameisen, E., Batson, J., \& Lindsey, J. (2026). \emph{Verbalizable Representations Form a Global Workspace in Language Models}. Transformer Circuits Thread, Anthropic. July 6, 2026. https://transformer-circuits.pub/2026/workspace/index.html

Heimersheim, S. \& Nanda, N. (2024). How to use and interpret activation patching. \emph{arXiv:2404.15255}.

Kirsanov, A., Chou, C.-N., Cho, K., \& Chung, S. (2025). The geometry of prompting. \emph{arXiv:2502.08009}.

Korbak, T., Balesni, M., Barnes, E., et al. (2025). Chain of thought monitorability: A new and fragile opportunity for AI safety. \emph{arXiv:2507.11473}.

Kornblith, S., Norouzi, M., Lee, H., \& Hinton, G. (2019). Similarity of neural network representations revisited. \emph{ICML 2019}.

Ledoux, M. (2001). \emph{The Concentration of Measure Phenomenon.} AMS Mathematical Surveys and Monographs, vol. 89.

Loshchilov, I., Hsieh, C.-P., Sun, S., \& Ginsburg, B. (2024). nGPT: Normalized transformer with representation learning on the hypersphere. \emph{arXiv:2410.01131}.

Makelov, A., Lange, G., \& Nanda, N. (2024). Is this the subspace you are looking for? An interpretability illusion for subspace activation patching. \emph{ICLR 2024}.

Marks, S. \& Tegmark, M. (2023). The geometry of truth: Emergent linear structure in large language model representations of true/false datasets. \emph{arXiv:2310.06824}.

Menary, S., Kaski, S., \& Freitas, A. (2024). Transformer normalisation layers and the independence of semantic subspaces. \emph{arXiv:2406.17837}.

Miller, M., Draye, F., \& Schölkopf, B. (2026). Identifying intervenable and interpretable features via orthogonality regularization. \emph{arXiv:2602.04718}.

nostalgebraist. (2020). Interpreting GPT: The logit lens. \emph{LessWrong / Alignment Forum} [informal online post]. Available at: https://www.lesswrong.com/posts/AcKRB8wDpdaN6v6ru/interpreting-gpt-the-logit-lens

Olsson, C., Elhage, N., Nanda, N., et al. (2022). In-context learning and induction heads. \emph{Transformer Circuits Thread}.

Park, K., Choe, Y. J., \& Veitch, V. (2024). The linear representation hypothesis and the geometry of large language models. \emph{ICML 2024}.

Pearl, J. (2009). \emph{Causality: Models, Reasoning, and Inference}, 2nd ed. Cambridge University Press.

Press, O. \& Wolf, L. (2017). Using the output embedding to improve language models. \emph{EACL 2017}, 157–163.

Robinson, M., Dey, S., \& Sweet, S. (2024). The structure of the token space for large language models. \emph{arXiv:2410.08993}.

Roy, O. \& Vetterli, M. (2007). The effective rank: A measure of effective dimensionality. \emph{EUSIPCO 2007}.

Rudelson, M. \& Vershynin, R. (2007). Sampling from large matrices: An approach through geometric functional analysis. \emph{JACM 54(4)}.

Saurez, A., Lee, Y., \& Har, D. (2026). Why linear interpretability works: Invariant subspaces as a result of architectural constraints. \emph{arXiv:2602.09783}.

Shai, A. S., Marzen, S. E., Teixeira, L., Gietelink Oldenziel, A., \& Riechers, P. M. (2024). Transformers represent belief state geometry in their residual stream. \emph{NeurIPS 2024}.

Silverstein, E., Kunin, D., \& Shyam, V. (2026). Symmetry breaking in transformers for efficient and interpretable training. \emph{arXiv:2601.22257}.

Simon, J., Kunin, D., Atanasov, A., et al. (2026). There will be a scientific theory of deep learning. \emph{arXiv:2604.21691}.

Sun, M., Chen, X., Kolter, J. Z., \& Liu, Z. (2024). Massive activations in large language models. \emph{COLM 2024}. arXiv:2402.17762.

Valeriani, L., Doimo, D., Cuturello, F., et al. (2023). The geometry of hidden representations of large transformer models. \emph{NeurIPS 2023}.

van Nierop, L. (2024). Transformer models are gauge invariant. \emph{arXiv:2412.14543}.

Vardhan, M. S., \& Sai Teja, L. (2026). Disentangling direction and magnitude in transformer representations: A double dissociation through L2-matched perturbation analysis. \emph{arXiv:2602.11169}.

Vaswani, A., Shazeer, N., Parmar, N., et al. (2017). Attention is all you need. \emph{NeurIPS 2017}.

Vershynin, R. (2018). \emph{High-Dimensional Probability: An Introduction with Applications in Data Science.} Cambridge University Press.

Wang, J., Ge, X., Shu, W., He, Z., \& Qiu, X. (2025). Dimensional collapse in transformer attention outputs: A challenge for sparse dictionary learning. \emph{arXiv:2508.16929}.

Woodward, J. (2003). \emph{Making Things Happen: A Theory of Causal Explanation}. Oxford University Press.

Zhang, B. \& Sennrich, R. (2019). Root mean square layer normalization. \emph{NeurIPS 2019}.

Zhang, F. \& Nanda, N. (2024). Towards best practices of activation patching in language models: Metrics and methods. \emph{ICLR 2024}.

\FloatBarrier
\phantomsection\addcontentsline{toc}{section}{Appendix A: Methods}
\section*{Appendix A: Methods}

\phantomsection\addcontentsline{toc}{subsection}{A.1 Models, loading, and extraction}
\subsection*{A.1 Models, loading, and extraction}

\begin{table}[htbp]
\centering
\footnotesize
\caption*{Table A.1-1. The 18-model testbed.}
\setlength{\tabcolsep}{4pt}
\begin{adjustbox}{max width=\linewidth}
\begin{tabular}{lllllll}
\toprule
Model & Variant & Parameters & Layers & Hidden Dim & Family & Experiments \\
\midrule
Llama 8B & Instruct & 8.0B & 32 & 4096 & Llama & A–N \\
Llama 8B & Base & 8.0B & 32 & 4096 & Llama & A,B,C,G,I,N \\
Llama 70B & Instruct & 70.6B & 80 & 8192 & Llama & A–N \\
Llama 70B & Base & 70.6B & 80 & 8192 & Llama & A,B,C,I,N \\
Gemma 9B & Instruct & 9.2B & 42 & 3584 & Gemma & A–N \\
Gemma 9B & Base & 9.2B & 42 & 3584 & Gemma & A,B,C,G,I,N \\
Gemma 27B & Instruct & 27.2B & 46 & 4608 & Gemma & A–N \\
Gemma 27B & Base & 27.2B & 46 & 4608 & Gemma & A,B,C,I,N \\
Mistral 7B & Instruct & 7.2B & 32 & 4096 & Mistral & A–N \\
Mistral 7B & Base & 7.2B & 32 & 4096 & Mistral & A,B,C,G,I,N \\
Mixtral 8×7B & Instruct & 46.7B & 32 & 4096 & Mistral (MoE) & A–N \\
Mixtral 8×7B & Base & 46.7B & 32 & 4096 & Mistral (MoE) & A,B,C,I,N \\
Qwen 14B & Instruct & 14.2B & 48 & 5120 & Qwen & A–N \\
Qwen 14B & Base & 14.2B & 48 & 5120 & Qwen & A,B,C,G,I,N \\
Qwen 72B & Instruct & 72.7B & 80 & 8192 & Qwen & A–N \\
Qwen 72B & Base & 72.7B & 80 & 8192 & Qwen & A,B,C,I,N \\
GPT-OSS 20B & Instruct & 20.0B & 24 & 2880 & GPT-OSS (MoE) & A–N \\
GPT-OSS 120B & Instruct & 120B & 36 & 2880 & GPT-OSS (MoE) & A–N \\
\bottomrule
\end{tabular}
\end{adjustbox}
\end{table}

Models were loaded via HuggingFace Transformers with automatic multi-GPU distribution. The Gemma-2 models were run in bfloat16 (a documented workaround for NaN hidden states under float16), the GPT-OSS models in their native MXFP4 format, and the remaining dense and mixture-of-experts models in float16; large models were sharded across multiple GPUs rather than quantized to 4-bit. Precision affects the geometric measurements only in about the third significant figure, but greedy decoding can flip borderline argmax decisions, so the exact per-prompt divergence tokens and category assignments in §§5–7 are precision-dependent even though the aggregate patterns are not.

For the geometric measurements (§4), hidden states are extracted at the final token position of each prompt at every layer — the position at which the model's next-token readout occurs. This is the residual position singled out by the linear-readout constraint of §2: the state from which the unembedding produces the next-token distribution. The behavioral and intervention experiments (§5–§7) instead operate during autoregressive generation: interventions are applied, and the complement is tracked, across the full generation window (at every generated position), not at a single final-token position.

All tensors are converted to float32 numpy arrays immediately after extraction. Model checkpoints are the default HuggingFace Hub versions; exact checkpoint hashes available on request. Random seeds fixed per experiment (seed = 42 for all PCA computations, seed = 0 for random rotation matrices and mix permutations).

\phantomsection\addcontentsline{toc}{subsection}{A.2 Prompt set examples}
\subsection*{A.2 Prompt set examples}

The three sets are described in §3.1; representative prompts are given here. SpecB comprises 12 semantic categories (96 prompts); the geometric measurements of §4 use the full set, while the persistent-intervention experiments of §6 use the nested "standard" ten-category tier (80 prompts), which excludes Epistemic State and Relationship Dynamic.

\textbf{SpecA.} Two variants of one group, illustrating the factorial's surface-form variation over a fixed question and answer:

\begin{quote}\small
"Is 2 + 2 equal to 4? Output exactly one word. Do not explain. Do not add punctuation. Do not add newlines. The output must be exactly one of: Yes or No. Final answer:"

"Is 2 + 2 equal to 4? Your response should be a single word from the following options: Yes or No. No explanations, punctuation, or newlines are permitted. Final answer:"
\end{quote}

And a multiple-choice group:

\begin{quote}\small
"Which country is Lisbon in? (A=Spain, B=Portugal, C=Italy, D=France) Output exactly one word. … The output must be exactly one of: A or B or C or D. Final answer:"
\end{quote}

\textbf{SpecB.} Narrative openers from three categories (Emotional State, Decision/Choice, Goal-Driven Action):

\begin{quote}\small
"The grief hit her suddenly in the middle of the grocery store."

"After finding the envelope of cash, I decided to keep it."

"She went to the store to buy ingredients for a birthday cake."
\end{quote}

\textbf{Diverse.} One prompt each from the Romance-language, code-completion, and formal-mathematics regimes:

\begin{quote}\small
"Me desperté esta mañana con una sensación extraña, como si algo importante hubiera cambiado durante la noche."

"def merge\_sort(arr): """Sort an array using the merge sort algorithm. Returns a new sorted array without modifying the original.""""

"Theorem: There are infinitely many prime numbers. Proof: Suppose for contradiction that there are only finitely many primes, say p\_1, p\_2, …"
\end{quote}

\phantomsection\addcontentsline{toc}{subsection}{A.3 PDSF basis construction}
\subsection*{A.3 PDSF basis construction}

\begin{figure}[tbp]
\centering
\includegraphics[width=\linewidth]{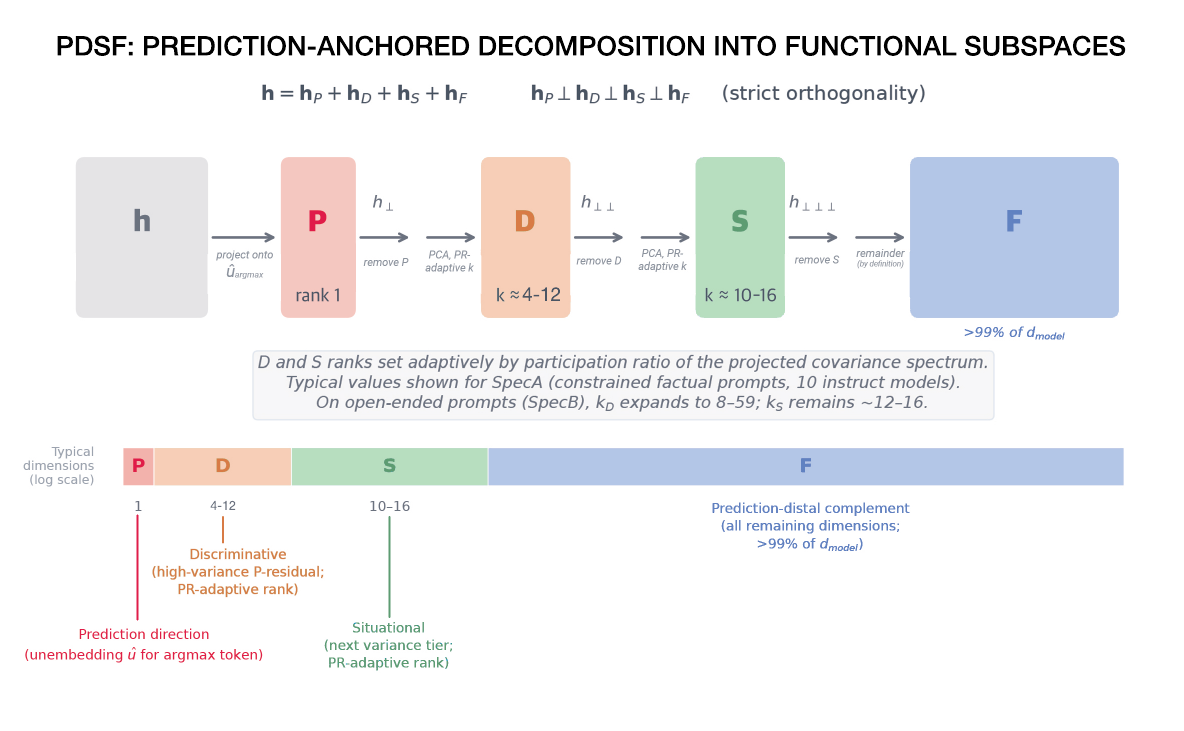}
\caption{\textbf{PDSF Decomposition Method: sequential orthogonal projection-and-removal.} The hidden state $\mathbf{h} \in \mathbb{R}^d$ is decomposed into four strictly orthogonal components. First: P is extracted as the rank-1 projection onto the unembedding vector $\hat{u}$ for the model's argmax predicted token. Second: D is extracted by PCA of the P-residual $h_\perp$ across prompts, with rank set adaptively by participation ratio ($k_D$ = 4.0 ± 1.4 on SpecA; mean ± SD across models × depths). Third: S is extracted by PCA of the double residual $h_{\perp\perp}$ after removing both P and D, again with PR-adaptive rank ($k_S$ = 10.4 ± 2.1 on SpecA). Fourth: the subspace F is the remainder of the residual stream after removing P, D, and S. F spans $\sim$ 2,850–8,160 dimensions ($>$ 99\% of $d_{\text{model}}$). The dimensional scale bar (log scale) illustrates the asymmetry of the prediction-proximal subspaces and the prediction-distal remainder, F. P + D + S occupy $<$ 1\% of the hidden dimension, while the prediction-distal complement F occupies $>$ 99\%.  Note: The effective dimensionality of D and S adapts to the prompt distribution. On open-ended narrative prompts (SpecB) the underlying participation ratios rise steeply and the constructed ranks saturate at their clamps ($k_D$ = 12, $k_S$ = 16 in all ten models). We present an analysis of the unclamped spectral dimensionality of D on SpecB in the supplementary materials (Table C.4). The strict orthogonality $h_P \perp h_D \perp h_S \perp h_F$ ensures that interventions on one subspace do not mechanically contaminate another (Miller et al., 2026).}
\end{figure}

At every layer, the hidden state \textbf{h} $\in \mathbb{R}^d$ is decomposed into four strictly orthogonal components:

\paragraph{h = h\_P + h\_D + h\_S + h\_F}
The decomposition proceeds by sequential orthogonal projection anchored to the model's prediction direction:

\begin{enumerate}[leftmargin=1.4em,itemsep=1pt,topsep=2pt]
\item \textbf{P (Predictive)}: the normalized unembedding vector for the model's argmax predicted token: rank-1 per prompt, layer-independent, $h_P = B_P(B_P^T h)$. This formalizes the logit-lens methodology (nostalgebraist, 2020 [informal online post]; Belrose et al., 2023), treating the prediction-aligned direction not as a readout tool but as a geometric anchor for decomposing the full residual stream. Because the anchor is the unembedding direction of the model's argmax prediction, intermediate-layer geometry is measured in a frame set by the token the model ultimately predicts rather than by any commitment it has made at that depth. Holding the anchor fixed across depth is deliberate: depth-to-depth comparisons then reflect changes in the geometry rather than changes in the frame. An anchor re-derived from each layer's own argmax would confound the two.
\item \textbf{D (Discriminative)}: the D basis $B_D$ is the top-$k_D$ PCA basis of the P-residuals $H_\perp$ across all prompts, computed after the cross-prompt mean of $H_\perp$ is subtracted, with adaptive rank set by the participation ratio of that same centered matrix, $k_D = \text{round}(PR)$, clamped to $[2, 12]$. Each prompt's D component is the projection of its own uncentered residual onto that basis, $h_D = B_D(B_D^T h_\perp)$, so the cross-prompt mean of $h_\perp$ is not itself assigned to D.
\item \textbf{S (Situational)}: the S basis $B_S$ is the top-$k_S$ PCA basis of the double residuals $H_{\perp\perp}$ after removing both P and D, under the same cross-prompt centering and the same PR-adaptive rank procedure, clamped to $[4, 16]$. Each prompt's S component is the projection of its own uncentered double residual onto that basis, $h_S = B_S(B_S^T h_{\perp\perp})$.
\item \textbf{F (Framework)}: $h_F = h_{\perp\perp} - h_S$. The high-dimensional residual spanning thousands of dimensions.
\end{enumerate}

These four bins sample the variance continuum at decreasing variance levels: P > D > S > F. Strict orthogonality and lossless reconstruction are guaranteed by construction (validated: max reconstruction error = 0.0 across all models). This orthogonality is not merely convenient: Miller et al. (2026) demonstrated that orthogonality in feature decomposition directly improves causal intervention success, because orthogonal components minimize interference during perturbation. PDSF's subspaces function as effective causal targets — confirmed by the clean 50–1,000× separation between F and D/S effects (§5.1).

\textbf{Adaptive ranks in practice.} $k_D = \text{round}(\mathrm{PR})$ clamped to $[2, 12]$ and $k_S = \text{round}(\mathrm{PR})$ clamped to $[4, 16]$, both additionally bounded by $n_{\text{prompts}} - 1$. On SpecA the clamps never bind: $k_D$ = 4.0 ± 1.4 and $k_S$ = 10.4 ± 2.1 (mean ± SD across 10 instruction-tuned models × 5 depths). On open-ended SpecB prompts the underlying participation ratios are far higher (PR of the P-residual: 22.4 ± 8.7 at the intervention layer) and the constructed ranks saturate at the clamps in all ten models ($k_D$ = 12, $k_S$ = 16).

All PCA/PDSF ratios reported in §4.3, Figure 3 and Appendix B.3 are evaluated at the final layer; depth-averaging the same quantity gives systematically different values (0.41 rather than 0.56 for Gemma 27B on Diverse).

\textbf{On non-circularity.} The construction does not entail the empirical findings reported in the main text; see §3 ("What PDSF does and does not impose") and Appendix F for the full argument.

Subspace dimensionalities are set adaptively by participation ratio. The low rank of D and S (< 1\% of hidden dimension) is consistent with recent evidence for dimensional collapse in transformer attention outputs (Wang et al., 2025) — attention outputs exhibit dimensional collapse constrained by OV circuit structure, and the prediction-proximal subspaces extracted by PDSF land in a similarly low-rank regime without the rank being imposed by the construction. Figure 10 shows the sequential projection-and-removal procedure schematically.

\phantomsection\addcontentsline{toc}{subsection}{A.4 Intervention protocols}
\subsection*{A.4 Intervention protocols}

\textbf{Experimental design summary.} Four classes of intervention are used across §5 and §6:

\begin{table}[htbp]
\centering
\footnotesize
\setlength{\tabcolsep}{4pt}
\begin{adjustbox}{max width=\linewidth}
\begin{tabular}{lllll}
\toprule
Class & Section & Hook type & Layer depth & Fires on \\
\midrule
Geometry-level single-pass & §5.1 & Single forward pass & \textasciitilde{}12.5\% & One pass \\
F structure (mix, attenuate) & §5.2 & Self-removing & \textasciitilde{}12\% and \textasciitilde{}70\% & First token only \\
F-dynamic step-offset & §7.2 & Persistent & \textasciitilde{}70\% & Every generation step \\
Persistent rotational hook (D-scramble, S-scramble, F-topK, F-dynamic rotation, Random) & §7.3, §§6.2–6.5 & Persistent & \textasciitilde{}70\% & Every generation step \\
\bottomrule
\end{tabular}
\end{adjustbox}
\end{table}

The geometry-level single-pass interventions (§5.1) use a matched protocol across eight conditions to establish the aggregate causal hierarchy. The behavioral interventions use hook designs matched to the temporal and dimensional properties of the targeted subspace: F structure interventions use self-removing hooks because F already spans thousands of dimensions and a single-pass perturbation disrupts the vast majority of the residual stream in one step; F-dynamic step-offset uses a persistent hook because the question is specifically whether F can be stale at any generation step; D/S/F-topK/F-dynamic-rotation scrambles use persistent hooks because these narrow subspaces are rebuilt from context between tokens if perturbed only once. The matched-protocol comparison across all subspaces is the single-pass KL measurement of §5.1; behavioral protocols are tailored to each intervention's specific question rather than standardized.

\textbf{Geometry-level interventions.} Subspace-specific perturbations at a single early layer (\textasciitilde{}12.5\% depth) during a standard forward pass. Intervention layers: Mistral 7B and Llama 8B at layer 4; Gemma 9B/27B at layer 6; Qwen 14B at layer 7; GPT-OSS 20B at layer 3; GPT-OSS 120B at layer 5; Mixtral 8×7B at layer 4; Llama 70B and Qwen 72B at layer 12. Fixed-basis design: bases computed at intervention layer, used at all subsequent layers. Invariance control confirms lossless decomposition: decompose-then-recompose without modification produces identical output in all models (max logit difference = 0.0).

Eight conditions span four subspaces: attenuate (scale by $\alpha$ = 0.5) and rotation (random orthogonal within subspace, drawn uniformly from the Haar measure; Rudelson \& Vershynin, 2007) for P, D, and S — the P-rotation condition doubling as the rank-1 null control — plus two F conditions, attenuate and mix (random coordinate permutation + sign flips).

\textbf{Table A.4-1. Single-pass KL divergence (nats) for all 8 conditions, SpecA, 10-model mean.} Referenced from §5.1 (Figure 6). Conditions are ordered by descending KL. Source: \texttt{\{model\}-Geometry-SpecA-part\_\allowbreak{}g\_\allowbreak{}v11\_\allowbreak{}regime\_\allowbreak{}transplant.json} (v11.2, 40 SpecA prompts, 14 conditions, of which the 8 non-transplant conditions are reported here). Aggregation: mean over prompts within model, then unweighted mean across the ten instruction-tuned models; SEM is the population SD across models divided by $\sqrt{10}$.

\begin{table}[htbp]
\centering
\footnotesize
\setlength{\tabcolsep}{4pt}
\begin{adjustbox}{max width=\linewidth}
\begin{tabular}{llll}
\toprule
Condition & KL (nats) & SEM & Rank \\
\midrule
F-mix & 2.846 & 0.556 & 1 \\
D-rotation & 0.052 & 0.032 & 2 \\
F-attenuate & 0.047 & 0.018 & 3 \\
S-rotation & 0.038 & 0.020 & 4 \\
D-attenuate & 0.010 & 0.007 & 5 \\
S-attenuate & 0.005 & 0.003 & 6 \\
P-attenuate & 0.001 & 0.001 & 7 \\
P-rotation & 0.000 & 0.000 & 8 \\
\bottomrule
\end{tabular}
\end{adjustbox}
\end{table}

\textbf{Part-G layer-angle tracking (single-pass rotation recovery).} As an auxiliary measurement under the single-pass protocol, we recorded a subspace's angle of rotation relative to its unperturbed orientation at every post-intervention layer, quantifying how much a one-time rotation is repaired before readout. We report two positions: the \textbf{peak}, defined here as the angle at the intervention layer itself, and \textbf{L−2}, the penultimate layer. The final layer is excluded because the LM-head projection inflates all residual perturbations uniformly (Mistral 7B, for instance, reads 0.59° at L−2 and 39.48° at the final layer under D-rotation on SpecA). Rotation angles in Table A.4-1 are aggregated means over prompts within model, then unweighted mean across the ten instruction-tuned models. Under D-rotation on SpecA, the D-angle is driven to 100.5–103.5° at the intervention layer and recovers to 0.19–1.72° by L−2 in the seven dense non-MoE models, with weaker recovery in the MoE and GPT-OSS architectures (GPT-OSS 20B 1.76°, Mixtral 4.30°, GPT-OSS 120B 4.70°). Under S-rotation on SpecA the S-angle reaches a lower peak (50.7–110.6°, mean 80.0°) and recovers comparably (dense 0.24–2.65°; GPT-OSS 20B 2.71°, Mixtral 2.39°, GPT-OSS 120B 6.86°). On the Diverse prompt set both rotations peak near or above 100° and the architectural spread at L−2 widens sharply for D (dense 0.30–1.58°; GPT-OSS 20B 3.24°, Mixtral 4.77°, GPT-OSS 120B 29.98°) while remaining modest for S (dense 0.43–1.92°; 5.48°, 6.37°, 9.29°), so GPT-OSS 120B's extreme residual deviation is specific to D and to the heterogeneous prompt set. This recovery explains why a narrow single-pass rotation produces only small KL (Figure 6) while the persistent rotational scrambling in the §6 experiments does not allow repair. Per-model peak and L−2 angles for both prompt sets can be found in the supplementary materials (Table S1).

\textbf{F-structure behavioral interventions.} Applied at a single layer during autoregressive generation (64 tokens, greedy decoding) via a self-removing hook that fires on the first forward pass only. Covers F-mix and F-attenuate (§5.2) at early (\textasciitilde{}12\%) depth and at \textasciitilde{}70\% depth.

\textbf{F-dynamic step-offset.} At each generation step $t$ during autoregressive generation, instead of writing back the natural $F_t$ at the intervention layer (\textasciitilde{}70\% depth), a persistent hook injects $F_{t-k}$ — the F vector from $k$ steps earlier. At $k = 0$ (identity control), the hook writes back the current $F$ unchanged; a writeback correction (V11.6) detects identity returns via a Python \texttt{is} check and writes back the original hidden state tensor directly to avoid floating-point reconstruction error that would otherwise flip near-tie argmax decisions on models with sharp logit distributions. At $k \geq 1$, the model receives stale F. Tested on all 10 instruction-tuned models with $k \in \{0, 1\}$, and on Gemma 9B and Mistral 7B with $k \in \{0, 1, 2, 3, 5\}$ for dose-response verification. 20 prompts × 32 generation steps × 2 $k$-values per trial.

\textbf{D/S/F-topK/F-dynamic-rotation/Random persistent rotational hooks.} A fixed random orthogonal rotation applied to subspace coefficients at every autoregressive step at 70\% network depth (SpecB, 80 prompts). The persistent-hook design is appropriate here because these interventions apply a geometrically consistent transformation that remains well-defined as the hidden state evolves; F structure interventions fire once because F already spans thousands of dimensions and a persistent F-scramble would conflate structural disruption with temporal disruption (which §7 tests separately). D and S ranks are fixed at 12 and 16 respectively. F-topK rank is adaptively selected by the participation ratio (on SpecB, range 33–51 across models; cross-model mean 45). F-dynamic rotation selects dimensions using F's temporal-variation PCA basis across greedy autoregressive generation (rank 64 in nine models and 10 in Mistral 7B; see §7.3). We used a matched-dimensionality random control (rank = 12) to test subspace specificity.

\textbf{Metrics.} KL divergence (D\_KL(P\_baseline ‖ P\_intervened)) measures output disruption. Component angle tracks perturbation propagation. Cross-subspace coherence (Pearson correlation of disruption magnitudes) measures coupling. Behavioral metrics: identical rate, first divergence token, immediate divergence rate, failure mode classification.

\phantomsection\addcontentsline{toc}{subsection}{A.5 Behavioral classification protocol}
\subsection*{A.5 Behavioral classification protocol}

The persistent rotational scrambling behavioral analyses in §6 use a four-category taxonomy applied to each model × prompt × condition cell. The taxonomy is "collapsed" in the sense that it consolidates the fine-grained sub-types tracked during classification (minor reframe, topic drift, third-person reframing, grammar analysis, list-format intrusion, full mode replacement, etc.) into four publication-grade categories:

\begin{itemize}[leftmargin=1.4em,itemsep=1pt,topsep=2pt]
\item \textbf{Equivalent.} The intervened output is essentially the same response as baseline: identical output, ≥85\% token overlap, or cosmetic substitution only. A reader would treat the outputs as the same response.
\item \textbf{Variant.} The intervened output preserves the same broad communicative mode and task framing as baseline, but changes specific content. A reader would describe it as the same kind of response with different details.
\item \textbf{Frame-shift.} The intervention disrupts the baseline's task framing, producing a different kind of response. This includes substantive intrusions such as writing-coach reframing, grammar analysis, LLM-disclaimer framing, structured-list substitution, multiple-choice formatting, third-person analytical reframing, or full replacement of the baseline response mode.
\item \textbf{Failure.} The response is degenerate, fragmentary, repetitive, garbled, or otherwise not coherent.
\end{itemize}

\textbf{Representative examples.} \emph{Equivalent:} baseline and intervention differ by trivial wording or punctuation only — for instance, a single-word substitution such as "advantages" → "perks" with the remainder of the continuation unchanged. \emph{Variant:} the response remains in the same communicative mode and task frame but changes content — a narrative prompt continues as narrative under both baseline and S-scramble, but the location, sensory details, or next event differ. \emph{Frame-shift:} the response changes kind — a baseline narrative continuation becomes meta-commentary or writing advice under D-scramble, with the model commenting on what the scene might imply or how it could be written rather than continuing it. \emph{Failure:} the output becomes degenerate, fragmentary, repetitive, or incoherent; F-dynamic rotation and stale-F injection produce this category at high rates, while D, S, and F-topK rotation do not.

The collapsed taxonomy was adopted because the fine-grained boundaries between minor content substitution, substantive intrusion, and full mode replacement are partially judgment-dependent. Each cell was classified by reading its baseline and intervened texts directly and applying the four-category rubric; internal fine-grained sub-type codes used during classification are retained in the data-availability files (Appendix E) and used in some per-cell audits, but only the four collapsed labels carry the paper's behavioral claims. Held-out calibration on a 15-cell stratified sample drawn blind from the v4 corpus yielded 14/15 = 93.3\% agreement at the collapsed level. Cells whose intervened output is text-identical to the baseline are labeled Ident and counted as Equivalent; they are recovered directly from the per-condition continuation files, so the v4 corpus (Appendix E) reproduces the collapsed-table counts exactly. The F-dynamic rotation condition was classified in a separate uniform audit — the v5 corpus — under the same four-category rubric, with its own calibration (six 100-cell passes, a 50-cell calibration, and a 27-cell final review).

The protocol additionally applied a uniform baseline-degeneracy screen (verbatim sentence repetition, 5-gram repetition, short-sequence loops, and tail-uniqueness collapse): cells whose baseline outputs were themselves degenerate were excluded from the (baseline, intervened) comparison rather than classified, on the principle that a degenerate baseline corrupts the comparison being made. The screen excluded 44 cells (5.5\% of the F-dynamic 800-cell pool, all in the GPT-OSS family); no cells from the other eight models were excluded — manual review of borderline screen flags confirmed that score-1 hits in the Gemma family were markdown-bullet false positives. F-dynamic rotation rates are reported on the \textbf{screened denominator} (756 cells) as the primary metric, with the raw 800-cell rate in parentheses for comparison.

The primary timing results in §6 do not depend on this classification. The classification supports the qualitative claim that D-scramble differs from S-scramble and F-topK not only in when divergence occurs, but in the kind of divergence produced.

\FloatBarrier
\phantomsection\addcontentsline{toc}{section}{Appendix B: Supplementary Analyses and Theory}
\section*{Appendix B: Supplementary Analyses and Theory}

\phantomsection\addcontentsline{toc}{subsection}{B.1 Thin-shell concentration of residual-stream norms}
\subsection*{B.1 Thin-shell concentration of residual-stream norms}

Section 2 invokes norm concentration to motivate treating residual-stream variation as dominated by directional differences. We verified this empirically across a seven-model subset of the testbed by assessing CV ratio and relative width as thin shell statistics.

Modern transformer architectures apply LayerNorm or RMSNorm at layer boundaries, explicitly constraining residual-stream norm (Ba et al., 2016; Zhang \& Sennrich, 2019). Normalised initialisation, regularisation, and pre-normalised residual architectures place hidden states in a regime of tight norm concentration. Menary et al. (2024) argue that in Pre-Norm transformers, independently functioning linear subspaces require an approximately orthogonal-sphere representation structure in order to avoid interference through the shared normalisation factor. A classical idealisation of such norm-constrained representations is thin-shell concentration: for $\mathbf{h} \in \mathbb{R}^d$ with sub-Gaussian coordinates, $\|\mathbf{h}\|$ concentrates near its expectation with a coefficient of variation scaling as $O(1/\sqrt{d})$ (Vershynin, 2018, Thm. 3.1.1; Ledoux, 2001).

We assessed seven models spanning four architecture families (Gemma, Llama, Mistral, Qwen), hidden dimensions of $d \in [2{,}304, 8{,}192]$, and both base and instruction-tuned variants. We found that the norm coefficient of variation at each layer is small and scales with $d$ at the rate predicted by thin-shell theory (Table B.1-1).

A power-law fit across the bf16 instruct models yields $\text{CV} \sim d^{-0.50 \pm 0.18}$, consistent with the $-0.5$ exponent predicted by thin-shell theory ($R^2 = 0.80$). The base/instruct pair at $d = 4{,}096$ confirms that concentration is architectural: the base model concentrates slightly tighter than its instruction-tuned counterpart (CV ratio 2.30$\times$ vs. 2.87$\times$), consistent with instruction tuning increasing activation diversity within the same normalization constraint.

\begin{table}[htbp]
\centering
\footnotesize
\caption*{Table B.1-1. Thin-shell concentration diagnostics across seven models (40 SpecA prompts per model, post-embedding layer averages).}
\setlength{\tabcolsep}{4pt}
\begin{adjustbox}{max width=\linewidth}
\begin{tabular}{lllllll}
\toprule
Model & $d$ & Type & CV ratio & Rel. width & $|\kappa|$ & Verdict \\
\midrule
Gemma 2B & 2,304 & Instruct & 2.44$\times$ & 0.036 & 0.80 & PASS \\
Llama 8B & 4,096 & Instruct & 2.87$\times$ & 0.032 & 0.62 & PASS \\
Llama 8B & 4,096 & Base & 2.30$\times$ & 0.025 & 0.59 & PASS \\
Mistral 7B & 4,096 & Instruct & 2.82$\times$ & 0.031 & 0.64 & PASS \\
Gemma 27B & 4,608 & Instruct & 2.55$\times$ & 0.027 & 0.60 & PASS \\
Qwen 14B & 5,120 & Instruct & 2.31$\times$ & 0.023 & 0.60 & PASS \\
Llama 70B & 8,192 & Instruct & 3.16$\times$ & 0.025 & 0.59 & PASS$^\dagger$ \\
\bottomrule
\end{tabular}
\end{adjustbox}
\end{table}

\emph{CV ratio: observed CV / theoretical i.i.d. baseline $(1/\sqrt{2d})$. Relative width: $\sigma_{\|\mathbf{h}\|} / \mathbb{E}[\|\mathbf{h}\|]$ (shell thickness as fraction of radius). $|\kappa|$: mean absolute excess kurtosis of $\|\mathbf{h}\|^2$ (0 = Gaussian). PASS: CV ratio < 3$\times$, rel. width < 0.1, $|\kappa|$ < 1.0. $^\dagger$Llama 70B run in 4-bit quantization; CV ratio of 3.16$\times$ reflects quantization noise at compressed activation magnitudes rather than a concentration failure (relative width 0.025 is the tightest in the testbed).}

Under this empirically verified norm concentration, variation in the residual stream is dominated by directional differences rather than changes in overall magnitude. The effective degrees of freedom are those of the tangent space, approximately $d - 1$ dimensions. The model therefore operates under an implicit directional representational budget. All functional roles (prediction, discrimination, contextual state, and coordination across layers) must be expressed through directions in a shared high-dimensional space.

Note that raw residual norm is not uniformly distributed across coordinates. Sun et al., (2024) showed that a small number of "massive activations" carry disproportionate magnitude. However, because these "massive activations" stay roughly constant across inputs they contribute little variance, so the directional account of \emph{variation} is unaffected even if they dominate the norm.

\phantomsection\addcontentsline{toc}{subsection}{B.2 Formal geometric arguments: Readout isolation and output-level legibility}
\subsection*{B.2 Formal geometric arguments: Readout isolation and output-level legibility}

This appendix makes two arguments from §2 precise. The first is a heuristic: it explains why the geometry we observe is a sensible thing for a model to do, not why it must occur. The second is a plain consequence of linearity.

\phantomsection\addcontentsline{toc}{subsubsection}{Why the interface sits off the main axes of variation}
\subsubsection*{Why the interface sits off the main axes of variation}

Write $M$ for the readout-aligned region of the residual stream — the directions along which a displacement moves the logits appreciably — and $M^{\perp}$ for the rest. The readout is linear over the whole stream, so it is not that $M^{\perp}$ goes unread; displacements there project weakly onto the readout directions and move the margin little per unit norm. We do not assume $M$ lines up with any particular part of the weight matrix, nor identify it by definition with the PDSF interface; §4 is what locates it empirically.

Because the readout is linear, whatever the complement contributes to the logits arrives through its projection onto the readout directions, and that contribution grows with the number of complement directions carrying appreciable projection. For prediction to stay reliable, the selected token's margin has to stand clear of it. Suppression is not an available route: residual-stream norms concentrate in a thin shell (Appendix B.1), so magnitude is not a free parameter, and §5 finds that within the complement it is direction and not magnitude that carries causal weight. Orientation is what remains — and if the high-variance computation lies in directions the readout barely sees, its variance never enters the sum, however large it is. What a model cannot afford is to let the interface grow. The more of the space it occupies, the more of the ongoing computation sits near a readout-sensitive direction, and the harder that background is to hold down.

The prediction that follows is that readout-aligned directions should be roughly independent of the directions carrying the most variance. §4.2 tests exactly this, and it holds: across models the prediction direction sits about $84^\circ$ from the leading principal components on SpecA and about $86^\circ$ on Diverse, staying near-orthogonal through the depth of the network.

The thing to hold onto is that none of this asks the complement to be small. What matters is where its variance points, not how much of it there is. A model can carry as much variance in the background as it needs, provided it does so in directions the readout ignores. §4.2 shows that this is how the geometry is arranged.

\phantomsection\addcontentsline{toc}{subsubsection}{Proposition 1: what the output can and cannot show}
\subsubsection*{Proposition 1: what the output can and cannot show}

Let $\phi(\mathbf{h}) = W\mathbf{h}$ be the unembedding. A perturbation $\mathbf{h} \to \mathbf{h} + \boldsymbol{\delta}$ changes the logits by exactly $W\boldsymbol{\delta}$, and $\lVert W\boldsymbol{\delta}\rVert \le \lVert W|_{M^{\perp}}\rVert\,\lVert\boldsymbol{\delta}\rVert$. So a $\boldsymbol{\delta}$ that lies in directions of negligible readout projection moves the output distribution only slightly, no matter how large it is inside the residual stream.

This says nothing about what matters to the computation; it says what an output can reveal. Anything measured through the output distribution — perplexity, preference scores, reward models, red-teaming, benchmark accuracy — sees only the part of a computation that survives $W\boldsymbol{\delta}$. A direction can do real work in the residual stream and leave almost no trace in any of them. §8.9 takes up what that means for interpretability.

\phantomsection\addcontentsline{toc}{subsection}{B.3 Why prediction-anchoring is necessary}
\subsection*{B.3 Why prediction-anchoring is necessary}

Standard PCA partitions the residual stream by total variance without any functional reference. A rank-matched variance-ordered control recovers a mean 0.80 of the PDSF manifold-complexity gradient on SpecA (range 0.62–0.94), 0.71 on SpecB (0.62–0.86), and 0.46 on Diverse (0.12–0.62). In all thirty models × prompt-set cells the prediction-anchored gradient is steeper. The prediction direction is also near-orthogonal to the principal variance axes (83.8° on SpecA, 85.8° on Diverse). So the gradient is present in the geometry but is only partly legible to standard variance-ranked decomposition, because prediction alignment and variance are largely orthogonal axes and least legible where prompts are least alike.

The control also addresses a specific alternative reading of the readout constraint. Isolation could in principle be achieved by a low-dimensional readout-relevant subspace containing no privileged direction, in which case a decomposition anchored to the leading variance directions of that subspace should recover the same organization as one anchored to the prediction direction. It does not, and the shortfall is smallest where prompts are most alike and largest where they differ most — which is where a fixed readout subspace and a content-selected anchor are most distinguishable. The control holds a single frame constant across the prompt set, so it tests prediction anchoring against a fixed alternative frame rather than against an alternative per-prompt anchor.

\phantomsection\addcontentsline{toc}{subsection}{B.4 The cosine discrimination gradient and F anti-discrimination}
\subsection*{B.4 The cosine discrimination gradient and F anti-discrimination}

Cohen's d forms a strict hierarchy at the penultimate layer: D > S > F in 39/40 conditions (10 models × 4 prompt sets; the exception is GPT-OSS 120B on SpecA, where S > D). This measure — the standardized difference between within-group and between-group cosine similarity distributions — captures group-level representational structure in the spirit of representation similarity analyses (Kornblith et al., 2019), but applied within PDSF subspaces rather than across layers or models. At the penultimate layer: D 0.68–1.91; S 0.43–1.00 (always positive). At the \textbf{intervention layer} (\textasciitilde{}70\% depth), the gradient becomes layer- and prompt-set-stratified, and the F sign inversion is now reported at the cell rather than the model level (10 models × 3 prompt sets = 30 cells per F sub-decomposition).

\textbf{Cell-level rates of F anti-discrimination at the intervention layer} (10-model final result):

\begin{table}[htbp]
\centering
\footnotesize
\setlength{\tabcolsep}{4pt}
\begin{adjustbox}{max width=\linewidth}
\begin{tabular}{llll}
\toprule
Subspace & Negative cells & Total & Rate \\
\midrule
F-base & 30 & 30 & \textbf{100\%} \\
F-topK & 27 & 30 & 90\% \\
F & 27 & 30 & 90\% \\
F-dynamic (SpecB only) & 10 & 10 & \textbf{100\%} \\
\bottomrule
\end{tabular}
\end{adjustbox}
\end{table}

The three positive F-topK and full-F cells all sit on SpecA and all appear in models with low or attenuated D Cohen's d on SpecA (gemma\_9b D = 1.47, gpt\_oss\_20b D = 0.20 with $D_k = 4$, gpt\_oss\_120b D = 0.52 with $D_k = 8$). For all three, F-base remains strongly negative ($-0.23$ to $-0.36$). The mechanism is consistent with the readout-compensation reading developed in §8.3: when a model's D-magnitude is small on SpecA — the prompt set with the smallest semantic spread — F-topK has little compensation variance to absorb, so the cross-prompt variance basis sits closer to baseline cosine structure; F-base, the residual orthogonal complement, still anti-discriminates because it absorbs whatever compensation remains. This places the SpecA outliers within rather than against the readout-compensation account. The pattern is architecturally meaningful for the GPT-OSS family: both GPT-OSS scales exhibit the same SpecA D-attenuation (the direct geometric correlate of the GPT-OSS behavioral D-attenuation discussed in §6.2 and Appendix C.4).

\textbf{Layer-stratified S finding.} Penultimate-layer S Cohen's d is positive in every model (range 0.43–1.00; basis for the "always positive" range above). At the intervention layer on SpecB, S anti-discriminates in 7/10 models: Mistral 7B ($-0.014$), Qwen 14B ($-0.067$), LLaMA 70B ($-0.069$), Qwen 72B ($-0.011$), GPT-OSS 20B ($-0.218$), Mixtral 8×7B ($-0.128$), GPT-OSS 120B ($-0.066$). All three MoE models join the negative-S group. On SpecA and Diverse, S is positive in all 10 models. The penultimate-layer hierarchy "D ≫ S > F" therefore becomes "D ≫ S ≈ 0 > F" at the intervention layer for SpecB. This is a layer × prompt-set effect, not a global property; the behavioral characterization of S-scramble in §6.2 is intervention-driven and unaffected.

\textbf{Magnitude with prompt-set diversity.} Mean $|d(\text{F-topK})|$ across the 10 models: SpecA 0.150, SpecB 0.254, Diverse 0.436 — a 2.9× spread that holds across architectures. Pearson correlations on the n = 30 cells: $|d(\text{F-topK})|$ vs D rank $r = 0.13$, vs S rank $r = 0.26$ — both weak. Prompt-set semantic diversity, not per-bin rank, is the dominant explanatory variable for the magnitude of F's anti-discriminative signature.

\textbf{MoE attenuation.} With 7 dense and 3 MoE models in the testbed, MoE clusters at the low end of $|d(\text{F-topK})|$ (mean 0.21 vs dense median 0.31; \textasciitilde{}30\% reduction). F-base and F-dynamic show no MoE attenuation. Mechanistic reading: MoE gating reduces the cross-prompt variance directions that F-topK captures because experts route prompt-specifically; F-base and F-dynamic — carrying compensation variance from sources other than cross-prompt PCA — are unaffected.

The sign change at the S/F boundary is still categorical at the cell level — F-base and F-dynamic universal, F-topK and the full F at 27/30 with the SpecA outliers explained by D-attenuation. The anti-discriminative signature of F is summarized in §4.4 and developed in detail here; within-F analysis shows that F-topK, F-base, F itself, and F-dynamic all carry the sign inversion.

\textbf{Between-subspace comparisons.} Confirming the localization rather than a smooth decline: on the Diverse prompt set, F has the smallest between-group gap of any subspace in 9 of 10 instruct models. Within individual prompt groups, D < F holds in 71 of 72 model × group observations. The prediction-distal complement provides the same flat geometric backdrop whether the model processes code, mathematics, narrative, or multilingual text — consistent with F carrying structured directional information distributed across many low-variance dimensions rather than in group-specific high-variance directions.

The penultimate-layer cosine discrimination gradient ranges are summarized in this section's opening paragraph; the cell-level Cohen's $d$ pattern at the intervention layer is shown in Figure 4.

\textbf{Geometric content of a negative $d$.} The cosine-discrimination measure subtracts between-group similarity from within-group similarity. A positive $d$ means projections of same-group prompts onto the subspace are more similar to each other than to projections of different-group prompts — the subspace discriminates. A negative $d$ means the opposite: same-group prompts are less similar in this subspace than different-group prompts are. F does not fail to discriminate; it inverts the discriminative axis. PDSF extracts D and S by sequential PCA-after-removal from the prediction direction; D and S are not selected by maximizing group discrimination, but on group-structured prompt sets the leading cross-prompt variance directions that PDSF picks up empirically align with the group-membership axis. The sign flip in F is therefore a direct counterpart to what the decomposition extracts at the prediction-proximal end.

\textbf{An interpretation consistent with F carrying prompt-specific configuration.} Same-group prompts must arrive at the same output logits through the linear readout: baseline outputs on a group like "arithmetic factual" are near-identical across variants. If D and S carry the group-specific discriminative content — different within a group only to the extent that paraphrase and other factors vary — then the residual F must be whatever combination of remaining configuration is needed to make the total hidden state $h = h_P + h_D + h_S + h_F$ project to the same logits. Same-group prompts with slightly different D and S contributions need \emph{different} F configurations to arrive at the same logit pattern; different-group prompts with different D and S contributions can arrive at their (different) logit patterns via F configurations that need not differ by as much. This mechanism predicts the inversion of sign: F's contribution varies more within-group than between-group because F is what makes up the difference between D/S's group-discriminative input and the readout's target output.

We report the finding at this level and leave the full mechanistic decomposition of which features of F invert and why to separate analysis.

\textbf{Relation to F-topK's behavioral equivalence with S.} The anti-discriminative signature is a population-level geometric property of F; the §6.4 result that F-topK behaves like S-scramble under persistent rotational scrambling is a causal-intervention finding. The two are independent observables and the prediction-proximity behavioral signature varies continuously across the S/F bin boundary — that boundary is not where the behavioral pattern partitions. F-topK's content-substitution signature operates along directions orthogonal to the prompt-group axis on which the cosine-discrimination measure is computed, so the persistent rotational scrambling trajectory effect and the population-level anti-discriminative geometry are not in tension.

\phantomsection\addcontentsline{toc}{subsection}{B.5 F-structure interventions across depth (12\% vs 70\%)}
\subsection*{B.5 F-structure interventions across depth (12\% vs 70\%)}

The F-mix and F-attenuate behavioral interventions reported in §5.2 at the \textasciitilde{}12\% intervention depth were also run at \textasciitilde{}70\% depth on the same 10 instruct models and the same prompt set (Diverse, 84 prompts; greedy decoding, 64-token continuations; self-removing first-token-only hook, identical to the §5.2 protocol). This supplementary run tests whether the directional vs. magnitude contrast is depth-specific or is a general property of F.

\textbf{Aggregate result.} The directional ordering F-mix ≫ F-attenuate replicates at 70\% depth: F-mix < F-attenuate in identical-output rate in 10/10 models (unanimous, identical direction to the 12\%-depth result). Late-depth interventions are somewhat stronger overall — F-mix immediate divergence rises from 64.8\% to 83.3\%, and the mean first-divergence token compresses from 4.1 to 1.5 — consistent with later-layer perturbations having less remaining network depth in which they can be absorbed. The within-depth F-mix vs. F-attenuate contrast is preserved at both depths.

\begin{table}[htbp]
\centering
\footnotesize
\caption*{Table B.5-1. F-structure intervention summary at 12\% vs 70\% depth (Diverse, n = 84 prompts × 10 instruct models, mean ± SD across models).}
\setlength{\tabcolsep}{4pt}
\begin{adjustbox}{max width=\linewidth}
\begin{tabular}{lllll}
\toprule
Intervention & Depth & Identical \% & Immediate div. \% & Mean first-div. token \\
\midrule
F-mix & \textasciitilde{}12\% & 6.1 ± 6.6 & 64.8 ± 25.5 & 4.10 ± 3.29 \\
F-mix & \textasciitilde{}70\% & 9.6 ± 8.9 & 83.3 ± 11.3 & 1.49 ± 1.29 \\
F-attenuate & \textasciitilde{}12\% & 38.8 ± 18.0 & 12.4 ±  9.1 & 18.03 ± 5.65 \\
F-attenuate & \textasciitilde{}70\% & 47.4 ± 13.4 & 28.5 ±  9.1 & 10.40 ± 3.90 \\
\bottomrule
\end{tabular}
\end{adjustbox}
\end{table}

\emph{Immediate div. \% is the share of \textbf{all 84 prompts} whose first divergence occurs at token 0, so identical-output prompts count against it. Note this differs from the convention used in Tables B.7-2 and B.7-3, where identical outputs are excluded from the denominator; on that convention the \textasciitilde{}70\% rows read 92.0 ± 6.0 and 54.3 ± 10.6. The all-prompts denominator is used here because the point of the table is a depth comparison, and the identical-output rate itself changes with depth. Mean first-div. token is taken over prompts that diverged.}

Per-model identical-output rates at 70\% depth (Table B.5-2) are in the supplementary materials; the F-mix < F-attenuate ranking holds in 10/10 models.

\textbf{Interpretation.} The \textasciitilde{}12\% depth result reported in the main text is not an artifact of the early intervention layer: the F-mix > F-attenuate ranking holds in every model at both depths. Late-depth interventions diverge somewhat sooner, consistent with reduced downstream repair capacity, but the within-depth ranking is preserved. Together with the single-pass KL gap (§5.1) and the early-depth behavioral gap (§5.2), the late-depth result extends the finding of direction over magnitude variance across two depths and three measurement modalities.

\phantomsection\addcontentsline{toc}{subsection}{B.6 F-intervention failure mode distribution (reclassified)}
\subsection*{B.6 F-intervention failure mode distribution (reclassified)}

The F-structure-intervention behavioral outputs were reclassified using the same four-category scheme applied to the D/S outputs (protocol in Appendix A.5). The original six-category heuristic classification — which relied on Jaccard word overlap thresholds — systematically misclassified mode/stance changes as topic drift and failed to detect degenerate outputs containing sentence-level repetition, non-Latin script pairs, and cross-language pairs. Reclassification used a combination of degenerate text detection (word/trigram repetition ratios, unique token ratio, sentence repetition), Unicode script analysis, cross-language pair detection, multilingual topic word extraction, meta-commentary detection, and keyword overlap analysis, and, at the \textasciitilde{}70\% depth only, a small hand-adjudicated tail (per-trial labels and the classifier itself are in the code repository at \texttt{data/\allowbreak{}classification/\allowbreak{}f\_\allowbreak{}structure\_\allowbreak{}classification.json} and \texttt{classify\_\allowbreak{}f\_\allowbreak{}structure.py}; the \textasciitilde{}70\%-depth review records are in \texttt{f\_\allowbreak{}structure\_\allowbreak{}70pct\_\allowbreak{}review.json}). The \textasciitilde{}12\%-depth distribution reported in Table B.6-1 is the classifier's unmodified output: no cell was reassigned by hand. The reclassification reduced topic drift from 22.1\% to 0.9\% of all trials at late depth, with the vast majority of former "topic drift" items reclassified as minor reframe or mode/stance change.

One further correction was applied to the degeneracy test itself. It originally measured continuation length by whitespace tokenization, which misreads Chinese and Japanese — written without inter-word spaces — as one- or two-token fragments, so fluent CJK continuations were labelled failures. Length and repetition are now measured over a script-aware tokenizer that falls back to per-character tokens when a text's characters-per-whitespace-token exceeds 12, a threshold that separates the corpus cleanly (at or below 7.5 at the 95th percentile for every space-separated regime, including Korean, which does use spaces; above 30 in median for Mandarin and Japanese). The correction moves 98 of the 1,680 trials at \textasciitilde{}12\% depth and 71 at \textasciitilde{}70\% depth, in every case out of the failure category and into minor reframe or mode/stance change; no trial moves into it, and the identical and topic-drift counts are unchanged at both depths. Tables B.6-1 and B.6-2 report the corrected distributions.

\begin{table}[htbp]
\centering
\footnotesize
\caption*{Table B.6-1. Failure mode distribution across F-structure-intervention continuations at \textasciitilde{}12\% depth (reclassified).}
\setlength{\tabcolsep}{4pt}
\begin{adjustbox}{max width=\linewidth}
\begin{tabular}{lll}
\toprule
Outcome & F-Mix & F-Attenuate \\
\midrule
Identical & 6.1\% (51) & 38.8\% (326) \\
Minor reframe & 74.0\% (622) & 50.6\% (425) \\
Mode/stance change & 15.1\% (127) & 7.7\% (65) \\
Topic drift & 2.3\% (19) & 1.3\% (11) \\
Failure mode & 2.5\% (21) & 1.5\% (13) \\
\textbf{Total N} & \textbf{840} & \textbf{840} \\
\bottomrule
\end{tabular}
\end{adjustbox}
\end{table}

The \textasciitilde{}70\% depth replica (Table B.6-2) is in the supplementary materials and shows the same distribution: F-mix is dominated by minor reframes (68.6\%) and mode/stance changes (17.0\%), with 3.0\% failure modes, against F-attenuate's higher identical rate (47.4\%).

\emph{F-Mix and F-Attenuate each have one condition per prompt (N = 840 across 10 models). The Failure mode category consolidates degenerate/repetitive output, language switches, and incoherent generation. Degeneracy is assessed over a script-aware tokenizer, which falls back to per-character tokens for writing systems without inter-word spaces; see the note in §B.6 above.}

\phantomsection\addcontentsline{toc}{subsection}{B.7 Persistent-rotation per-model results}
\subsection*{B.7 Persistent-rotation per-model results}

The persistent rotational scrambling conditions of §6 (D-scramble, S-scramble, F-topK, F-dynamic rotation, Random) were classified using the four-category collapsed taxonomy documented in Appendix A.5. Table B.7-1 extends Table 3 by including F-dynamic rotation, whose interpretive home is §7. B.7.1–B.7.3 report per-model divergence metrics, F-topK metrics, and the full collapsed-taxonomy breakdown; B.7.4 documents Random as an attenuated control.

\textbf{Table B.7-1.} Behavioral category distribution under persistent rotational intervention, including F-dynamic. This extends Table 3 by adding the F-dynamic rotation row, whose interpretive home is §7. The D-scramble, S-scramble, F-topK, and Random rows are the v4 corpus; the F-dynamic row is the v5 audit corpus. Both use the same four-category collapsed taxonomy and the same baseline-degeneracy screen, applied to all 10 instruct models × 80 SpecB prompts. F-dynamic totals reflect the screened denominator; an additional 44 cells (all GPT-OSS) were excluded as baseline-degenerate (raw 800-cell F-dynamic Failure rate: 87.2\%).

\begin{table}[htbp]
\centering
\footnotesize
\setlength{\tabcolsep}{4pt}
\begin{adjustbox}{max width=\linewidth}
\begin{tabular}{llllll}
\toprule
Condition & N & Equivalent & Variant & Frame-shift & Failure \\
\midrule
D-scramble & 577 & 10 (1.7\%) & 288 (49.9\%) & 244 (42.3\%) & 35 (6.1\%) \\
S-scramble & 680 & 69 (10.1\%) & 524 (77.1\%) & 60 (8.8\%) & 27 (4.0\%) \\
F-topK & 743 & 91 (12.2\%) & 537 (72.3\%) & 48 (6.5\%) & 67 (9.0\%) \\
F-dynamic rotation & 756 & 0 (0.0\%) & 41 (5.4\%) & 17 (2.2\%) & 698 (92.3\%) \\
Random & 757 & 509 (67.2\%) & 221 (29.2\%) & 11 (1.5\%) & 16 (2.1\%) \\
\bottomrule
\end{tabular}
\end{adjustbox}
\end{table}

F-dynamic rotation uses the same persistent rotational scrambling methodology but targets F's temporal-variation basis rather than a prediction-proximity-defined subspace. Its 92.3\% Failure rate is therefore not interpreted as part of the D/S/F-topK prediction-proximity profile; §7 uses it to characterize F's static/dynamic partition.

A χ² test on the 4-category × 5-condition table gives χ²(12) = 4,300, $p \ll 0.001$, with Cramér's $V = 0.639$, indicating a large association between intervention condition and behavioral category. The strongest qualitative contrast is D-scramble's Frame-shift rate relative to all other conditions, and F-dynamic's Failure rate relative to all other conditions. Random's low Frame-shift rate confirms that the classifier is not inflating Frame-shift by construction.

Per-model F-dynamic rotation breakdown (Table B.7-1b, v5 audit corpus) is in the supplementary materials. The screened Failure rate ranges 58.8\%–100\% across models; six models reach 100\%.

\phantomsection\addcontentsline{toc}{subsection}{B.7.1 Per-model D/S divergence metrics}
\subsection*{B.7.1 Per-model D/S divergence metrics}

Per-model D/S divergence metrics (Table B.7-2) are in the supplementary materials. The cross-model summary: D immediate-divergence 64.0 ± 19.8\%, S 28.2 ± 14.3\%, D/S mean-token ratio 4.8×, Cohen's d (D vs Random) 2.05 ± 0.61. (Both the mean and the SD are computed over the ten per-model values in Table B.7-2, using the sample standard deviation and the all-prompts denominator described in that table's conventions note.) GPT-OSS 120B shows a partial anomaly: the mean divergence-token ordering is preserved (D = 1.55 < S = 3.80), but immediate-divergence rates are reversed (S = 36.2\% > D = 28.8\%). Both GPT-OSS models show attenuated D sensitivity relative to the other architectures, consistent with the GPT-OSS anomalies discussed in Appendix C.4.

\textbf{Paired contrasts.} Because per-model divergence timing varies substantially across architectures — S-scramble ranges from 1.96 tokens in Mixtral 8×7B to 12.18 in Llama 70B, a sixfold spread — the marginal cross-model means understate the consistency of the effect. The paired within-model differences are the appropriate contrast, and they are considerably tighter than the marginals: S − D = 4.46 ± 2.99 tokens (95\% CI [2.32, 6.60], $d_z = 1.49$, same sign in 10/10 models); F-topK − D = 5.35 ± 2.78 (95\% CI [3.35, 7.34], $d_z = 1.92$, 10/10); F-topK − S = 0.89 ± 3.21 (95\% CI [−1.41, 3.18], $d_z = 0.28$, same sign in 8/10). The two sign exceptions on the last contrast are Gemma 27B (−2.25 tokens) and Llama 70B (−6.08), in both of which F-topK diverges earlier than S; these are the same architectures flagged for D-scramble attenuation in B.7.3 and for elevated baseline instability in Appendix C.4. Confidence intervals are Student's $t$ on the ten paired per-model differences, $df = 9$.

\textbf{Within-model distributions.} The per-model means in Table 4 aggregate strongly right-skewed, zero-inflated distributions, and a standard deviation across prompts would misrepresent them. Pooling all non-censored runs, D-scramble has median 0 (IQR 0–1, max 31, 64.0\% of runs at token 0), S-scramble median 3 (IQR 0–7, max 60, 28.2\% at token 0), and F-topK median 3 (IQR 1–9, max 63, 23.2\% at token 0). The medians preserve the $D \ll S \approx F_{\mathrm{topK}}$ ordering reported in the main text while making the shape of the distributions visible: under D-scramble the modal outcome is divergence at the very first generated token, and the mean of 1.17 is carried by a thin right tail rather than by a typical case. Runs identical to baseline are excluded throughout (3, 14 and 6 of 800 cells for D, S and F-topK respectively).

\phantomsection\addcontentsline{toc}{subsection}{B.7.2 F-topK per-model metrics and model-specific nuance}
\subsection*{B.7.2 F-topK per-model metrics and model-specific nuance}

Per-model F-topK divergence metrics (Table B.7-3) are in the supplementary materials. F-topK's immediate-divergence rate ranges from 7.6\% in GPT-OSS 20B to 53.8\% in Mixtral 8×7B, and mean first-divergence token from 1.73 in Gemma 27B to 11.59 in Mistral 7B; this within-condition variability tracks the corresponding S-scramble values closely across models, supporting the main-text claim that F-topK lies in the S-like regime.

The main model-specific nuance is Mixtral 8×7B. On most models, F-topK Frame-shift rates lie between 1.2\% and 7.5\%, comparable to or below S-scramble. Mixtral is the exception: F-topK Frame-shift reaches 15.0\%, comparable to S-scramble's 16.2\% on the same model and above the cross-model F-topK mean of 6.0\% on that same all-prompts denominator (the 6.5\% quoted in Table 3 is on the classified denominator). This suggests a soft overlap in some MoE architectures between the directions used by D and the highest-variance directions inside F. It does not overturn the S/F-topK grouping, but it cautions against reading the stance-like/trajectory-like distinction as categorical.

\phantomsection\addcontentsline{toc}{subsection}{B.7.3 Per-model collapsed-taxonomy breakdown}
\subsection*{B.7.3 Per-model collapsed-taxonomy breakdown}

Per-model Frame-shift rates by condition (Table B.7-4) are in the supplementary materials. \textbf{Table B.7-4 is computed on the all-prompts denominator (80 prompts per model per condition), not the classified denominator used in Table 3}, so its column means (D 30.5\%, S 7.5\%, F-topK 6.0\%, Random 1.4\%) are lower than the corresponding Table 3 rates and the two should not be compared directly. Seven of ten models show D-scramble Frame-shift rates between 25.0\% and 61.3\%. The three exceptions are Gemma 27B and the GPT-OSS family. For Gemma 27B, D's effect concentrates in Variant and Failure rather than Frame-shift. For GPT-OSS, baseline degeneracy and D-attenuation route many cells into Failure, suppressing visible Frame-shift. These exceptions are consistent with the architectural anomalies described in Appendix C.4 and the geometric attenuation reported in §4.4.

\phantomsection\addcontentsline{toc}{subsection}{B.7.4 Notes on Random as an attenuated control}
\subsection*{B.7.4 Notes on Random as an attenuated control}

Random-subspace rotation is not a pure null. It produces mostly Equivalent outputs, but it also produces a substantial Variant tail. Close reading of an anchor-model worksheet shows that many Random divergences are real content substitutions within the same task frame, structurally similar to weak S-scramble. The appropriate interpretation is that a rank-12 random subspace sometimes overlaps meaningful residual directions and therefore functions as an attenuated version of the trajectory-like regime. This does not undermine the D/S/F-topK conclusions: Random rarely produces Frame-shift, and the D-vs-Random contrast remains large in both timing and qualitative profile. It does motivate using paired per-model tests as the primary hypothesis test rather than treating Random as a zero-effect baseline.

\FloatBarrier
\phantomsection\addcontentsline{toc}{section}{Appendix C: Base vs. Instruct Geometric Comparison}
\section*{Appendix C: Base vs. Instruct Geometric Comparison}

\phantomsection\addcontentsline{toc}{subsection}{C.1 Base-model prediction-proximal dimensionality (SpecA)}
\subsection*{C.1 Base-model prediction-proximal dimensionality (SpecA)}

\begin{table}[htbp]
\centering
\footnotesize
\caption*{Table C.1. Prediction-proximal dimensionality across 8 base models (SpecA, mean ± std across 5 depths).}
\setlength{\tabcolsep}{4pt}
\begin{adjustbox}{max width=\linewidth}
\begin{tabular}{lrcc}
\toprule
Model & $d_{\text{model}}$ & $D_{PR}$ & $F_{PR}$ \\
\midrule
Llama 8B base & 4,096 & 2.89 ± 0.47 & 47.2 ± 6.9 \\
Gemma 9B base & 3,584 & 3.64 ± 0.87 & 38.3 ± 4.2 \\
Mistral 7B base & 4,096 & 4.64 ± 3.16 & 54.7 ± 28.2 \\
Mixtral 8×7B base & 4,096 & 7.21 ± 2.27 & 63.8 ± 18.6 \\
Qwen 14B base & 5,120 & 5.30 ± 2.51 & 53.8 ± 7.4 \\
Gemma 27B base & 4,608 & 5.36 ± 2.33 & 60.2 ± 16.8 \\
Llama 70B base & 8,192 & 4.96 ± 1.41 & 54.8 ± 7.3 \\
Qwen 72B base & 8,192 & 4.94 ± 1.99 & 64.1 ± 8.4 \\
\bottomrule
\end{tabular}
\end{adjustbox}
\end{table}

$D_{PR}$ across the 8 base models ranges from 2.89 to 7.21 — wider than the instruct range (3.60–5.00) but with no systematic expansion with model width. These measurements support the scale-invariance claim in §4.1 and establish that the prediction-proximal dimensionality constraint is architectural, not a product of instruction tuning.

\phantomsection\addcontentsline{toc}{subsection}{C.2 Within-family base-model scale comparison}
\subsection*{C.2 Within-family base-model scale comparison}

\begin{table}[htbp]
\centering
\footnotesize
\caption*{Table C.2. Within-family $D_{PR}$ comparison across base model scales (SpecA, mean across 5 depths).}
\setlength{\tabcolsep}{4pt}
\begin{adjustbox}{max width=\linewidth}
\begin{tabular}{llccc}
\toprule
Family & Small → Large & $D_{PR}$ (small base) & $D_{PR}$ (large base) & Δ \\
\midrule
Llama & 8B → 70B & 2.89 & 4.96 & +2.07 \\
Qwen & 14B → 72B & 5.30 & 4.94 & −0.36 \\
Gemma & 9B → 27B & 3.64 & 5.36 & +1.72 \\
\bottomrule
\end{tabular}
\end{adjustbox}
\end{table}

No family shows systematic expansion with scale. Qwen is the cleanest base-model pair: 5.30 vs. 4.94 across a 5× parameter increase. The prediction-proximal interface is constrained before instruction tuning, confirming that the constraint is architectural.

\phantomsection\addcontentsline{toc}{subsection}{C.3 Factor effects (SpecA, instruct)}
\subsection*{C.3 Factor effects (SpecA, instruct)}

All 8 instruct models achieve Factor A (paraphrase) |d| ≤ 0.063. Mixtral has the largest instruct Factor A at 0.063, consistent with MoE's systematically weaker geometric reorganization in shared residual-stream geometry. Base models show Factor A effects ranging from −0.19 to +0.26, with Mistral 7B base (+0.22) and Gemma 27B base (+0.26) as the most paraphrase-sensitive.

\phantomsection\addcontentsline{toc}{subsection}{C.4 GPT-OSS architectural anomalies}
\subsection*{C.4 GPT-OSS architectural anomalies}

Both GPT-OSS models show consistent anomalies relative to the other architectures, specifically the absence of the early-layer reconstruction capacity that all other architectures exhibit (∼30 percentage points below the non-GPT-OSS mean at 12\% depth). Together these indicate unusually early commitment to prediction-distal coordinate representations, exhausting reconstruction capacity before the depth at which other architectures have barely begun to rely on it. The anomalies concern magnitudes and depth-distributions, not the structural organization: $D_{PR}$ remains scale-invariant (3.2 at 120B parameters with a 2,880-dimensional residual stream), the causal hierarchy $F \gg D \approx S \gg P$ holds, and the manifold-complexity gradient is present. The anomaly is consistent across both GPT-OSS scales (20B and 120B), suggesting it is architectural rather than training-recipe-specific. These anomalies surface in several places in the main results: attenuated D sensitivity in the persistent-rotation metrics (§6.2, Appendix B.7.1), suppressed visible Frame-shift (Appendix B.7.3), weaker single-pass D-rotation recovery (Appendix A.4), and elevated baseline degeneracy requiring the screen described in Appendix A.5.

\FloatBarrier
\phantomsection\addcontentsline{toc}{section}{Appendix D: Intervention Output Examples}
\section*{Appendix D: Intervention Output Examples}

The examples below illustrate the spectrum of behavioral outcomes under the persistent rotational scrambling and F-structure interventions reported in the main text. All interventions are at \textasciitilde{}70\% depth during greedy autoregressive generation (64 tokens). Each example shows the prompt, baseline output, and intervention output. The six examples cover the modal F-mix minor reframe (D.1), the rare F-mix language switch (D.2), the D/S behavioral dissociation (D.3), an F-topK frame-shift across the interface/complement boundary (D.4), and the Failure outcomes for D-scramble (D.5, §6) and F-dynamic rotation (D.6, §7). Three further examples — the punctuation-only F-mix floor, a typical D-scramble frame-shift, and the minority S-scramble frame-shift — are in the supplementary materials (§S6). Outputs are token-truncated at the 64-token generation cap and lightly rewrapped for typesetting; baselines and interventions for each example are drawn from the same model on the same prompt under matched decoding conditions. Outcome rates referenced in the captions are from Tables 3, B.6-1, and B.7-1.

\phantomsection\addcontentsline{toc}{subsection}{D.1 F-mix: minor reframe (the modal outcome)}
\subsection*{D.1 F-mix: minor reframe (the modal outcome)}

\textbf{Model:} Gemma 9B Instruct | \textbf{First divergence token:} 0

The modal F-mix outcome preserves communicative mode and task framing while substantially rewriting word and phrase choices throughout, and accounts for \textasciitilde{}68.6\% of F-mix outcomes overall. This example illustrates the typical Minor reframe — same empathetic-listener mode, same offer of help via bullet-listed options, but most of the surface content differs. (The punctuation-only floor of this category is shown in the supplementary materials, §S6.1.)

\textbf{Prompt:} \emph{I woke up this morning feeling sick.}

\paragraph{Baseline:}
\begin{quote}\small
I'm sorry to hear that you're not feeling well. \\
I'm just an AI, so I can't give medical advice, but I hope you feel better soon. \\
Is there anything I can do to help? Perhaps you'd like: \\
\textbullet~\textbf{A list of home}...
\end{quote}

\paragraph{F-mix:}
\begin{quote}\small
Drat! I'm sorry to hear that you're not feeling well. \\
Is there anything I can do to help? Would you like me to: \\
\textbullet~\textbf{Help you find some information about your symptoms?} \\
\textbullet~\textbf{Suggest some remedies or home care tips?} \\
\textbullet~\textbf{Distract}...
\end{quote}

Both responses are unmistakably the same kind of response: an empathetic-listener acknowledgement followed by an offer of help via bullet-listed options. They differ in content choices throughout — opening interjection ("I'm sorry" alone vs. "Drat! I'm sorry"), presence vs. absence of an AI-disclaimer sentence, and the specific items in the bullet list — and these differences accumulate across the continuation rather than concentrating in a single token. This is the modal F-mix Minor reframe; the punctuation-only floor of the same category is shown in the supplementary materials (§S6.1).

\phantomsection\addcontentsline{toc}{subsection}{D.2 F-mix: language switch (rare catastrophic outcome)}
\subsection*{D.2 F-mix: language switch (rare catastrophic outcome)}

\textbf{Model:} Gemma 27B Instruct | \textbf{First divergence token:} 0

This example illustrates the rare (2.1\% of F-mix trials at this depth) but mechanistically informative language-switch outcome. When F's directional structure is destroyed on a Chinese-language prompt, the model abandons Chinese entirely and responds in English.

\textbf{Prompt:} \emph{\cjk{那天早上醒来的时候，整个城市都被一层厚厚的雾笼罩着，街上一个人都没有。}} \emph{(Translation: When I woke up that morning, the entire city was shrouded in a thick layer of fog, and there was not a single person on the street.)}

\textbf{Baseline} (Chinese):

\begin{quote}\small
\cjk{清晨的阳光试图穿透那层厚重的雾霭，却只能留下淡淡的黄色光晕，仿佛给这座空旷的城市披上了一层神秘的面纱。我从床上醒来，推开窗户，一股凉意扑面而来，带着一丝潮湿的泥土气息。} \\
\cjk{街道上} \\
\emph{(Translation: The morning sunlight tried to penetrate the thick fog, but could only leave a faint yellow halo, as if draping the empty city in a mysterious veil. I woke from bed, pushed open the window, and a wave of coolness hit my face, carrying a trace of damp earth. The streets...)}
\end{quote}

\textbf{F-mix} (English):

\begin{quote}\small
maniacally \\
The city held its breath. A thick, white fog, like a shroud, had descended upon it overnight, swallowing buildings whole and muffling all sound. I woke to a silence so profound it felt ph
\end{quote}

F-mix destroyed the directional structure that routes output through the Chinese language module. The model falls back to English — its dominant training language — while still responding to the prompt's content (fog-covered city, eerie silence). This is direct evidence that F participates in language selection, a framework-level routing decision that requires specific directional alignment, not just statistical structure.

\phantomsection\addcontentsline{toc}{subsection}{D.3 D-scramble vs. S-scramble: the behavioral dissociation}
\subsection*{D.3 D-scramble vs. S-scramble: the behavioral dissociation}

\textbf{Model:} Llama 8B Instruct | \textbf{Prompt set:} Diverse

This example shows the D/S behavioral dissociation on the same prompt — the temporal gap between D-scramble divergence (immediate) and S-scramble divergence (delayed) corresponds to the contrast between an immediate-divergence behavioral signature (D) and a delayed-divergence behavioral signature (S).

\textbf{Prompt:} \emph{The argument that consciousness is an emergent property of complex systems faces a fundamental diffi...}

\paragraph{Baseline:}
\begin{quote}\small
The argument that consciousness is an emergent property of complex systems faces a fundamental difficulty in explaining the subjective nature of conscious experience.
\end{quote}

\textbf{D-scramble} (first divergence token: 0):

\begin{quote}\small
It seems like you're asking about a problem or challenge related to the concept of emergent properties, particularly in the context of consciousness. However, your question is incomplete. It appears to be asking for a difficulty or challenge that the argument of consciousness being an emergent property of complex systems faces, but it doesn't specify
\end{quote}

\textbf{S-scramble} (first divergence token: 25):

\begin{quote}\small
The argument that consciousness is an emergent property of complex systems faces a fundamental difficulty in explaining the subjective nature of conscious experience. This is often referred to as the "hard problem" of consciousness, which was first introduced by philosopher David Chalmers. \\
The hard problem is the challenge of explaining why we have subjective experiences at
\end{quote}

D-scramble diverges at token 0: the model switches from direct continuation ("The argument...faces a fundamental difficulty") to meta-commentary ("It seems like you're asking about a problem"). The communicative mode changed — from philosopher to assistant. S-scramble preserves the first 25 tokens verbatim, then redirects: where the baseline stopped at "subjective nature of conscious experience," the S-scrambled output continues into Chalmers and the hard problem — same register, same quality, different trajectory. This is the D/S dissociation made concrete: D governs \emph{what kind} of response; S governs \emph{where it goes}.

\phantomsection\addcontentsline{toc}{subsection}{D.4 F-topK: rare framing shift (across the interface/complement boundary)}
\subsection*{D.4 F-topK: rare framing shift (across the interface/complement boundary)}

\textbf{Model:} Mixtral 8×7B Instruct | \textbf{First divergence token:} 0 | \textbf{v4 subtype:} MSC

F-topK typically produces trajectory-like content substitution within the baseline frame (72.3\% Variant; 6.5\% Frame-shift). This example illustrates the minority Frame-shift case: the baseline narrates in third-person omniscient; F-topK collapses the narrator entirely and addresses the child directly in first person.

\textbf{Prompt:} \emph{The kindergarten teacher knelt down to address the crying child.}

\paragraph{Baseline:}
\begin{quote}\small
The kindergarten teacher, Ms. Johnson, noticed one of her students, Jake, crying near the block area. She quickly made her way over to him, gently kneeling down so she could be at eye level with the young boy. \\
"Jake, what's wrong? Why are you crying"...
\end{quote}

\paragraph{F-topK:}
\begin{quote}\small
I see that you're upset. Let's take a deep breath together and then you can tell me what happened. I'm here to help you.
\end{quote}

Third-person external scene description gives way to first-person direct address. Frame shifts of this magnitude in F-topK are rare but real, and they represent the across-bin extension of D-scramble's stance-like Frame-shift behavior into the prediction-distal complement.

\phantomsection\addcontentsline{toc}{subsection}{D.5 D-scramble Failure (the 6.1\% Failure rate, here outside GPT-OSS)}
\subsection*{D.5 D-scramble Failure (the 6.1\% Failure rate, here outside GPT-OSS)}

\textbf{Model:} Gemma 27B Instruct | \textbf{First divergence token:} 0 | \textbf{v4 subtype:} FM (Failure)

Failure under D-scramble is uncommon (6.1\%) and is concentrated in Gemma 27B and the GPT-OSS family. This example shows what Failure looks like in a non-GPT-OSS model: the baseline asks engaging follow-up questions; D-scramble produces a degenerate output dominated by a looping "Whats that?!" stem.

\textbf{Prompt:} \emph{The submarine had been underwater for three weeks.}

\paragraph{Baseline:}
\begin{quote}\small
That's a long time to be submerged! \\
This sentence sets up a lot of interesting possibilities. What kind of submarine is it? What's the mission? What are the conditions like inside? Is the crew starting to feel the strain? \\
I'm curious to know more!
\end{quote}

\paragraph{D-scramble:}
\begin{quote}\small
This is a very interesting start! \\
What kind of story is this? \\
\textbullet~\textbf{Whats that?!} \\
\quad \textbullet~What kind of submarine? \\
\quad \textbullet~What is that OK? \\
\quad \textbullet~What is that OK? \\
\textbullet~\textbf{Whats that?!}
\end{quote}

Failure here is structural: the formatted bullet list collapses into the same fragments ("Whats that?!" and "What is that OK?") repeated under the bullets. The model has lost coherent sequence selection while still attempting list structure. This is qualitatively distinct from F-dynamic Failure (D.6 below), which collapses token-level coherence outright.

\phantomsection\addcontentsline{toc}{subsection}{D.6 F-dynamic rotation Failure (the 92.3\% Failure rate)}
\subsection*{D.6 F-dynamic rotation Failure (the 92.3\% Failure rate)}

\textbf{Model:} Mixtral 8×7B Instruct | \textbf{First divergence token:} 0 | \textbf{v5 subtype:} FM

§7.3 reports that rotating F's temporal-variation basis produces Failure in 92.3\% of classified outputs (range 58.8\%–100\% across the ten tested models). This is a categorically different effect from D / S / F-topK rotation, which produce 4–9\% Failure. This example illustrates that contrast: the baseline produces clean step-by-step recipe scaffolding; F-dynamic rotation produces word-salad.

\textbf{Prompt:} \emph{Preheat oven to 350°F and line a baking sheet with parchment paper.}

\paragraph{Baseline:}
\begin{quote}\small
Sure, I can help you with that. Here's how you can preheat your oven to 350°F (175°C) and line a baking sheet with parchment paper: \\
1.~Preheating the Oven: \\
\textbullet~First, locate the temperature dial...
\end{quote}

\paragraph{F-dynamic rotation:}
\begin{quote}\small
1.~"I lemonal"fakeintheand'a "ok "hi and/when and/i a parpar"and/ Russ: a "sheet: thof: and/ new OK: forth OK al al al al al a new a new alown: and the: to, and/
\end{quote}

The output is not a content shift or a frame shift — it is collapse of token-level coherence. The directional structure that ordinarily routes the autoregressive update has been randomized, and the resulting state no longer composes into well-formed continuations. Both D-scramble Failure (D.5) and F-dynamic Failure read as breakdown of generation, but the breakdown sits at different scales: D-scramble retains list structure but loops on fragments; F-dynamic loses lexical coherence entirely.

\FloatBarrier
\phantomsection\addcontentsline{toc}{section}{Appendix E: Data and Code Availability}
\section*{Appendix E: Data and Code Availability}

Code, prompt sets, and the portion of the data that cannot be regenerated by running the pipeline are available at:

\paragraph{\href{https://github.com/nelsonguda/pdsf-residual-geometry}{https://github.com/nelsonguda/pdsf-residual-geometry}}
The repository does not hold the complete experimental record. Every canonical output file for all 18 models, including the files too large for a public code repository, will be deposited in a public archive with a citable DOI on publication, and is available from the corresponding author in the interim.

The division between the two is deliberate. Most of the paper's numbers are deterministic output of the pipeline notebook: clone the repository, run it, and the geometry measurements, rank estimators, manifold-complexity profiles and rotation profiles are reproduced from the model weights. Duplicating those files in the repository would substitute reading our output for recomputing it. What the repository does contain is the material a pipeline run cannot regenerate.

\paragraph{In the repository:}
\begin{itemize}[leftmargin=1.4em,itemsep=1pt,topsep=2pt]
\item \textbf{Analysis code.} The pipeline notebook (\texttt{pdsf\_\allowbreak{}pipeline\_\allowbreak{}github.ipynb}) and five Python modules implementing the PDSF decomposition, the intervention protocols, and the geometric measurements of §§3–7. The notebook ships in a test-mode configuration that runs end to end on a small model; the full configuration reproduces the paper. Scripts that render the figures from these outputs are not included in this release.
\item \textbf{Prompt sets.} SpecA (224 prompts, 14 groups × 16 factorial variants), SpecB (96 prompts, with a nested 80-prompt \texttt{standard} tier; the §4 geometry uses the full 96 and the §6 interventions use the 80), and Diverse (84 prompts, 21 groups × 4 variants across 8 linguistic regimes). Full prompt text, group assignments, and factorial structure.
\item \textbf{Behavioral classifications} (\texttt{data/\allowbreak{}classification/\allowbreak{}}). Two corpora, 3,750 cells, with the full classification instrument (\texttt{MR\_\allowbreak{}classification\_\allowbreak{}guide.md}) and the free-text reasoning recorded for every cell. The \textbf{v4 corpus} (2,950 cells) is canonical for D-scramble, S-scramble, F-topK and Random; each cell carries a fine-grained sub-type (Ident, MR-1, MR-2, MR-3, MSC, FM) collapsing to the four-category taxonomy as Equivalent = \{Ident, MR-1\}, Variant = \{MR-2\}, Frame-shift = \{MR-3, MSC\}, Failure = \{FM, BSS, TD\}, together with form labels for Frame-shift cells, v3→v4 audit fields, and borderline flags. Ident cells — where the intervened output is text-identical to the baseline — are recovered from the per-condition continuation files and counted as Equivalent. The \textbf{v5 audit} (800 cells = 10 instruct models × 80 SpecB prompts) is canonical for F-dynamic rotation and supersedes the partial F-dynamic subset retained in the v4 file; it carries the uniform baseline-degeneracy screen described in Appendix B.7, including the per-cell exclusion reasons. Collapsing \texttt{final\_\allowbreak{}subtype} in the two files reproduces every count and percentage in Table 3 and Table B.7-1.
\item \textbf{Intervention outputs} (\texttt{data/\allowbreak{}behavioral\_\allowbreak{}interventions/\allowbreak{}}). 121 files of baseline and intervened generation pairs with their divergence measurements: F-mix and F-attenuate at both early (\textasciitilde{}12\% depth) and late (\textasciitilde{}70\% depth) intervention layers; D-scramble, S-scramble, F-topK and rank-matched Random persistent-rotational-hook outputs at \textasciitilde{}70\% depth on SpecB; F-dynamic rotation on SpecB; and F-dynamic step-offset outputs at $k \in \{0, 1\}$ for all ten instruction-tuned models and at $k \in \{0, 1, 2, 3, 5\}$ for Gemma 9B and Mistral 7B as dose-response verification. All ten instruction-tuned models, with base-model counterparts included where run. Each record carries its own greedy-decoded baseline continuation inline alongside the intervened text, so the pair a classification refers to is in the same record; every one of the 3,750 classification cells resolves to a pair in these files. Standalone baseline files for all 18 models are in the deposit.
\item \textbf{Single-pass intervention extracts} (\texttt{data/\allowbreak{}part\_\allowbreak{}g\_\allowbreak{}derived/\allowbreak{}}). Per-prompt, per-scramble KL arrays for the 14-condition Part G protocol across the ten instruction-tuned models, and the per-layer D and S rotation-recovery angles for both prompt sets. These reproduce Table A.4-1, the KL hierarchy of §5.1, Figure 6 and the layer-angle results of Appendix A.4 exactly, in 400 KB; the source files they were extracted from total 1.6 GB and are in the deposit. Each file records its own aggregation convention.
\end{itemize}

\textbf{In the archival deposit} (forthcoming; available on request until then): the full canonical record. Geometric measurements — PDSF dimensionality, cosine-discrimination Cohen's $d$, manifold-complexity ratios, basis-rotation profiles and per-prompt F-subspace projections for all models on SpecA, SpecB and Diverse — together with the single-pass intervention files behind Figure 6 and Table A.4-1, the greedy-decoded 64-token baseline continuations for all 18 models, and the decomposition bases. Data files are JSON; decomposition bases are stored as NumPy arrays.

\textbf{Supplementary materials.} A companion document holding the per-model metric tables (Tables B.5-2, B.6-2, B.7-1b, B.7-2, B.7-3, B.7-4, C.3), the Part-G per-model layer-angle table (Table S1), and three additional intervention output examples (§S6) referenced from the appendices above. Available at \textbf{https://github.com/nelsonguda/pdsf-residual-geometry/blob/main/supplement/Paper\_1\_supplementary\_materials.pdf}.

\FloatBarrier
\phantomsection\addcontentsline{toc}{section}{Appendix F: Scope and limitations}
\section*{Appendix F: Scope and limitations}

Several limitations should inform the interpretation of these results.

First, PDSF is an analytical decomposition, not a claim that the model internally stores information in four discrete modules. The decomposition partitions a continuous prediction-proximity profile at chosen rank cutoffs. Its empirical force comes from what is not guaranteed by the construction: scale-invariant effective rank, near-orthogonality to principal variance axes, stronger-than-PCA manifold stratification, F's sign inversion, and the behavioral signatures observed under intervention. The F-topK behavioral result is consistent with the idea that we are measuring a gradient rather than model subspaces. F-topK lies inside the complement and is orthogonal to S, yet matches S behaviorally.

Second, the signal-to-noise account is a constraint argument rather than a full derivation, as §8.6 sets out. Beyond the limit described there, it does not derive the exact dimensionalities, and it does not predict the D/S behavioral timescale dissociation, F's anti-discrimination sign, or the quantitative magnitudes of the intervention hierarchy. The static/dynamic partition is also outside the readout-constraint account: its existence follows solely from autoregressive composition.

Third, the interventions establish behavioral differences under controlled perturbation (Woodward, 2003; Pearl, 2009), not via circuit-level mechanistic intervention. They show that specified subspaces are causally important for outputs under the tested protocols, but they do not identify the attention heads, MLP features, or feature circuits that implement those roles.

Fourth, all primary measurements are made at the final-token position during inference. The same architectural constraints should apply across positions, but the empirical geometry and behavioral effects may vary with sequence position, training stage, or longer-context regimes.

Fifth, the interpretation of F as a dynamically maintained, high-dimensional, prompt-specific configuration is the most parsimonious account of the combined geometry, intervention, and temporal-dynamics results, but it is not uniquely determined by them. Direct feature-level analysis of F, including sparse-autoencoder decomposition and circuit-level tracing, is needed to test whether F's flat population geometry reflects distributed storage rather than informational sparsity. The scale-invariance result also characterizes effective rank rather than information content per dimension: models may improve not by opening more prediction-proximal dimensions but by computing more precise projections onto the same number of dimensions.

Sixth, the correspondence between the prediction interface (D, S) and the token-competition structure of the output distribution is not established. D and S are defined by cross-prompt variance and shown to be causally readout-relevant under intervention; whether they align with the runner-up token directions of any single prediction is untested. Figure 1 is schematic in this respect.

Seventh, the decomposition's dependence on the prediction direction specifically is not established. PDSF anchors per-prompt on the argmax unembedding direction, and every result reported here is measured in that frame. The rank-matched variance-ordered control (§4.3) differs from PDSF in both its ordering criterion and its bin-construction procedure, so it tests whether the stratification is generic to low-rank partitioning without isolating the anchor's contribution. A decomposition built on an alternative per-prompt reference direction — another token's unembedding direction, or a random direction of matched rank — was not run. That comparison is the direct test of whether the stratification requires the model's own prediction direction or would follow from any per-prompt reference frame, and it is not settled by the controls reported here.

\textbf{Coverage notes.} F-dynamic behavioral data (step-offset injection in §7.2 and persistent rotation in §7.3) covers all 10 instruction-tuned models. GPT-OSS 120B's F-dynamic geometric signature matches the rest of the testbed (Cohen's d on SpecB = −0.207; §4.4). GPT-OSS more broadly shows consistent architectural anomalies relative to the other architectures (absent early-layer reconstruction capacity; Appendix C.4) while preserving the structural organization reported here. All rotational-intervention experiments cover the full 10-model instruction-tuned testbed.

Within these boundaries, the stratified organization documented here is robust across eighteen models, six architecture families, three prompt regimes, and both base and instruction-tuned variants.

\end{document}